\documentclass{article}
\usepackage[main, final]{neurips_2026}

\usepackage[utf8]{inputenc}
\usepackage[T1]{fontenc}
\usepackage{xcolor}   
\usepackage{hyperref}
\hypersetup{
    colorlinks = true,
    linkcolor = blue!60!black,
    citecolor = blue!60!black,
    urlcolor = blue!60!black
}
\usepackage{url}
\usepackage{booktabs}
\usepackage{amssymb}
\usepackage{amsfonts}
\usepackage{nicefrac}
\usepackage{microtype}
\usepackage{mathtools}
\usepackage{mathrsfs}
\usepackage{graphicx}
\usepackage{subcaption}
\usepackage{makecell}
\usepackage{multirow}
\usepackage[capitalize,noabbrev]{cleveref}
\usepackage{thmtools}
\usepackage{thm-restate}
\usepackage{tabularx}
\usepackage{booktabs}
\usepackage{algorithm}
\usepackage{algpseudocode}
\usepackage{adjustbox}

\usepackage{tikz}
\usetikzlibrary{automata}
\usetikzlibrary{positioning}
\usetikzlibrary{arrows.meta}

\usepackage{enumitem}
\setlist[itemize]{leftmargin=0.5cm,topsep=0pt,itemsep=0pt}
\setlist[enumerate]{leftmargin=0.5cm,topsep=0pt,itemsep=0pt}

\usepackage{pifont}

\usepackage{wrapfig}
\usepackage[normalem]{ulem}

\definecolor{commentgray}{gray}{0.45}
\algrenewcommand\algorithmiccomment[1]{%
  \hfill{\color{commentgray}\small\textit{// #1}}}

\colorlet{rebuttalcolor}{blue}

\makeatletter
\newcommand{\algrule}[1][.4pt]{%
  \par\vskip.3\baselineskip\hrule height #1\par\vskip.3\baselineskip}
\makeatother

\title{Utility-Constrained Policy Optimization}

\author{%
  Mehrdad Moghimi \\
  York University \\
  Toronto, Canada \\
  \texttt{moghimi@yorku.ca} \\
  \And
  Bernardo \'Avila Pires \\
  Google DeepMind \\
  London, United Kingdom \\
  \texttt{bavilapires@google.com} \\
}

\begin{document}

\maketitle

\begin{abstract}
Constrained MDPs (CMDPs) are a widely adopted framework for incorporating safety into RL agents; however, the framework does not support risk-sensitive constraints.
This can be problematic: For example, CMDPs allow for optimal solutions that, in order to satisfy the risk-neutral constraints, mix infrequent catastrophic behaviors and frequent, overly conservative ones.
Moreover, prior empirical results suggest that enforcing stricter, risk-sensitive constraints can improve performance even under risk-neutral evaluation.
The natural framework to incorporate risk-sensitive constraints is utility-constrained MDPs (UCMDPs), but no practical solutions for this problem existed.
In this work, we introduce a simple yet powerful methodology for UCMDPs and constrained RL.
Besides allowing for risk-sensitive constraints, our framework does not require us to fix constraint limits in advance of training the agent, provided that a sensible range is known.
This increases policy flexibility and, in practice, allows for adjustments to these limits at no extra training cost.
Besides benefiting from the generality of the framework, our agent shows strong performance in practice, consistently matching or outperforming existing baselines in several Safety Gymnasium benchmark tasks.
\end{abstract}

\section{Introduction}
\label{sec:intro}

An important part of building goal-oriented intelligent agents is to express and enforce constraints on their behavior, because behavioral requirements are present in many domains, in the form of operational constraints, safety requirements, regulations, user preferences, and so forth.

Reinforcement learning (RL)~\citep{sutton2018reinforcement} is a popular framework for designing goal-oriented agents, and the standard way to incorporate constraints into RL is through constrained Markov decision processes (CMDPs)~\citep{altman2021constrained}.
A common RL problem is to find a policy that maximizes the expected (possibly discounted) sum of rewards, and the CMDP framework incorporates constraints on the expected (possibly discounted) sums of other reward-like signals observed along with the standard rewards.

Despite being widely adopted as the basis for many constrained RL methods (for example,~\citep{ji2023safety,ji2024omnisafe}), a substantial limitation of the CMDP framework is that constraints are risk-neutral, that is, they are only enforced in expectation.
Because of risk neutrality, catastrophic constraint violations are acceptable as long as they are infrequent enough, however this can lead to unreliable (yet optimal) policies---for example, policies that satisfy constraints some of the time~\citep{ray2019benchmarking}, or overly-conservative policies~\citep{achiam2017constrained}.

We are not the first ones to consider this limitation.
\citet{kadota2006discounted} extended CMDPs to include utility-based objectives and constraints, enabling risk-sensitive formulations, but without a clear practical solution, as optimal policies are history-dependent and often intractable to learn.
\citet{achiam2017constrained} proposed to satisfy a conservative upper-bound on the constraint; 
\citet{sun2021safe} proposed to terminate episodes whenever a constraint is violated; \citet{huang2023safedreamer} ensured that constraints were satisfied frequently before optimizing the objective;
\citet{sootla2022saute} added a penalty for constraint violation to the RL rewards, with increasing weights throughout training (in a barrier-like fashion); 
\citet{sootla2022effects} also added a similar penalty and used a curriculum on the constraint limits for learning constraint-satisfying policies (instead of the barrier-like violation penalty);
\citet{jiang2024reward} added a penalty relative to the constraint violation, and \citet{sikchi2022learning} used model-based planning with a worst-case cost constraint during action selection.
Somewhat surprisingly, these efforts to address risk-neutral CMDP constraints have led to improved performance in constrained RL benchmarks, \emph{where constraints are risk-neutral}.
This suggests that there are further performance gains to be made if we can improve how risk-sensitive constraints are satisfied, and we argue that we can accomplish this via a suitable problem formulation and a solution method for it.

\begin{figure}[htbp]
  \centering
  \begin{subfigure}[T]{0.49\linewidth}
    \includegraphics[width=\linewidth]{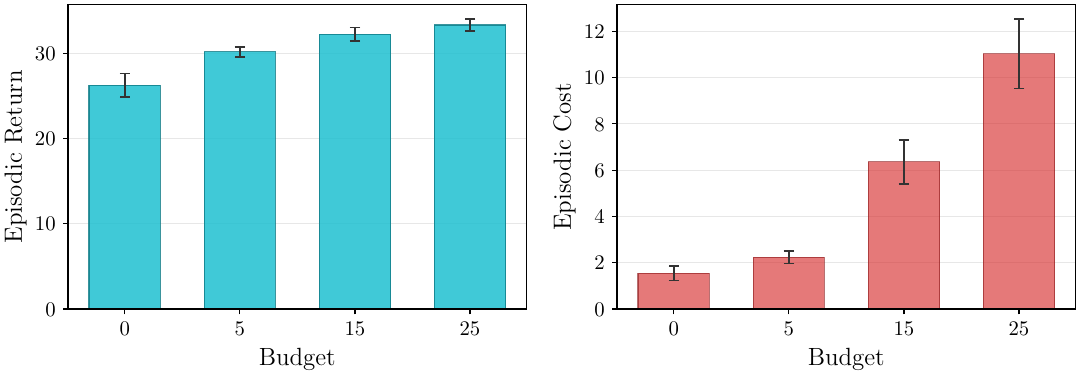}
    \caption{Episodic return (left) and cost (right) for different test-time cost budgets. For each seed, results are averaged over 1000 episodes; we then report the mean and 95\% Student’s $t$ confidence intervals across five seeds.}
    \label{fig:custom_side_by_side_1:return-and-cost-per-budget}
  \end{subfigure}
  \hfill
  \begin{subfigure}[T]{0.49\linewidth}
    \includegraphics[width=\linewidth]{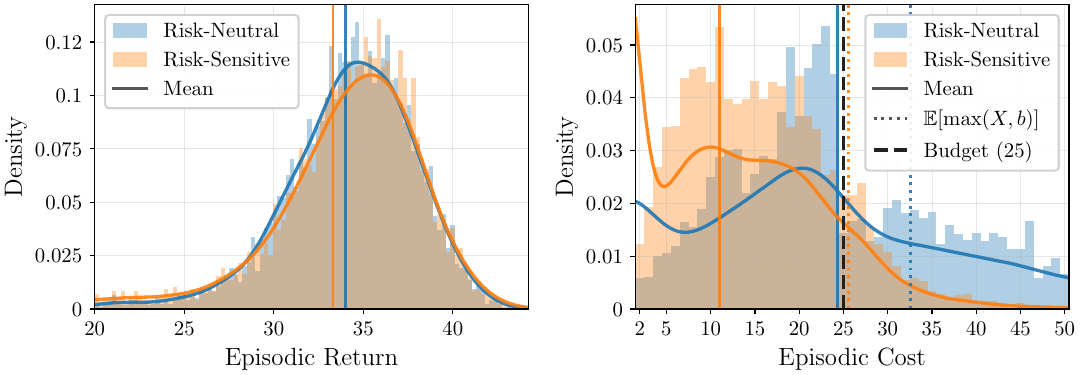}
    \caption{Empirical densities of episodic returns (left) and costs (right) for UCP with risk-sensitive vs.~risk-neutral constraints. Solid lines show means; dotted lines show expected excess cost; dashed line indicates the budget.}
    \label{fig:custom_side_by_side_1:distributions}
  \end{subfigure}
  \caption{Illustration of UCP in the \emph{Safety Navigation} CarGoal1 task.
  \label{fig:custom_side_by_side_1}}
\end{figure}

In this work, we modify the utility-constrained MDPs introduced by \citet{kadota2006discounted} to allow for practical solutions, and introduce a practical method for UCMDPs.
Our proposal builds on a framework introduced by \citet{pires2025optimizing} that allows for solving unconstrained MDPs with expected-utility objectives via dynamic programming.
A convenient side-effect of our modification is that constraint limits are inputs to policies along with the initial state of the MDP, so they can be set \emph{after} training agents, similar to what Saut\'e RL affords~\citep{sootla2022saute}. 
The solution method we introduce is called \emph{utility-constrained policies} (UCP), a Lagrangian deep RL agent for UCMDPs, and we demonstrate its effectiveness on the Safety Gymnasium benchmark~\citep{ji2023safety}.
\Cref{fig:custom_side_by_side_1} illustrates some of our agent's capabilities in the \emph{Safety Navigation} \emph{CarGoal} (level 1) task.

\Cref{fig:custom_side_by_side_1:return-and-cost-per-budget} shows the episodic return and the cost incurred by UCP, as a function of different cost budgets.
We use the same policy with different input values for the cost budgets (that is, we do not retrain the agent).
We see that the performance of the agent increases with the cost budget, as one would expect from an active constraint. 
More importantly, we observe the marginal gains in return as the budget increases, which can guide the choice of an appropriate budget.
Not only does the agent exhibit strong performance in the risk-neutral sense\footnote{To our knowledge, the best prior methods satisfying the expected cost budget of $25$ are the model-based RCEPETS~\citep{liu2020constrained,chua2018deep}, achieving an average return of $29.08$~\citep[Table 3(e)]{ji2024omnisafe}, and the model-free early-terminated TRPO~\citep{sun2021safe}, achieving $22.09$~\citep[Table 3(b)]{ji2024omnisafe}. See \cref{sec:performance} for additional results.
}, but it also reduces the tail of the distribution above the budget.
\Cref{fig:custom_side_by_side_1:distributions} illustrates the benefits of risk-sensitive constraints.
The plot shows the empirical distribution of the episodic return and cost for UCP (which has risk-sensitive constraints) and a variant of UCP with risk-neutral constraints (similar to CMDPs).
The risk-neutral UCP closely satisfies the constraint (the mean cost is slightly below the budget), often exceeding the cost budget, and sometimes incurring more than twice the cost, whereas the risk-sensitive UCP, which only penalizes the cost above the budget, has a smaller tail of exceeding the budget (the expected excess cost is almost zero, i.e. $\mathbb{E}[\max(X,b)]\approx b$), and a relatively small reduction in episodic return.

\textbf{Contributions.}
\begin{itemize}
    \item[-] We introduce a simple modification to UCMDPs to support optimal Markov policies and practical solutions. This modification allows constraint limits to be set \emph{after} computing a solution---they are inputs to policies along with the initial state of the MDP.
    \item[-] We introduce \emph{utility-constrained policies} (UCP), a Lagrangian algorithm for solving UCMDPs.
    \item[-] We conduct an extensive evaluation in Safety Gymnasium, demonstrating UCP's strong performance: UCP matches (overlapping 95\% confidence intervals) or outperforms the baselines in all six \emph{Safety Velocity} tasks considered, and outperforms, by a large margin, all the baselines in all sixteen \emph{Safety Navigation} tasks considered.
    \item[-] We empirically demonstrate the effects of addressing the two CMDP limitations: UCP reduces the expectation of costs above the budget relative to its risk-neutral variant, and it can perform subject to different cost budgets without the need to retrain.
\end{itemize}

\section{Background}
\label{sec:background}

\paragraph{Basic notation.}
We denote the real numbers by $\mathbb{R}$, natural numbers by $\mathbb{N}$, and $\mathbb{N}_0 \doteq \{0\} \cup \mathbb{N}$.
$\mathbb{I}\{\cdot\}$ denotes the indicator function and $(x)_+ \doteq \max\{x, 0\}$.
For a set $\mathcal{X}$, $\Delta(\mathcal{X})$ denotes the set of probability distributions over $\mathcal{X}$. 
For a random variable $X \sim \nu$, we write $\mathrm{df}(X) = \nu$.

\paragraph{Markov decision processes.}
In the standard RL problem, the environment is formalized as a Markov decision process (MDP)~\citep{puterman2014markov}, defined by a state space $\mathcal{S}$, an action space $\mathcal{A}$, a transition kernel $\mathscr{P} : \mathcal{S} \times \mathcal{A} \rightarrow \Delta(\mathcal{S})$, and a reward kernel $\mathscr{R} : \mathcal{S} \times \mathcal{A} \rightarrow \Delta(\mathbb{R})$.
In this work, we consider a more general MDP formulation, where we replace the reward signal with vector-valued \emph{cumulants}~\citep{sutton2011horde,sutton2018reinforcement}.
In this formulation, we have $\mathcal{Z} \doteq \mathbb{R}^m$ (for some $m \in \mathbb{N}$) and we use, instead of the reward kernel, a \emph{cumulant kernel} $\mathscr{R} : \mathcal{S} \times \mathcal{A} \rightarrow \Delta(\mathcal{Z})$.
We assume that cumulants have uniformly bounded first moment, and we denote the set of such distributions by $\mathcal{P}^1(\mathcal{Z})$.
We also assume that $\mathcal{A}$ is compact.

At each time step $t$, the agent observes state $S_t$, selects action $A_t$, and observes a reward vector $R_{t+1} \in \mathcal{Z}$ and next state $S_{t+1}$.
The history up to timestep $t$ is the sequence $S_0, A_0, S_1, R_1, \ldots, S_t, R_t$, and a policy is a mapping from histories to distributions over actions. We denote the set of history-based policies by $\Pi_{\mathrm{H}}$.
A \emph{Markov policy} is a mapping $\mathcal{S} \times \mathcal{A} \rightarrow \Delta(\mathcal{A})$.

Given a discount factor $\gamma \in (0, 1)$,
the \emph{discounted return} of a policy $\pi$, is the sum of discounted rewards observed by following $\pi$ starting from $S_t$: $G^\pi_t \doteq \sum_{k=0}^\infty \gamma^k R_{t+k+1}$.
With $\gamma < 1$ and the assumption on bounded first moments, the return distributions are also in $\mathcal{P}^1(\mathcal{Z})$.
We will occasionally use double-indexing for the returns, writing $G_t = (G_{t, 1}, \ldots, G_{t,m})$.

In standard RL, we have $\mathcal{Z} = \mathbb{R}$ ($m = 1$) and the cumulant has the semantics of a reward; the problem is to find a policy that maximizes expected return, 
\(
    \sup_{\pi \in \Pi_{\mathrm{H}}} \mathbb{E}[G^\pi_0],
\)
with $\mathcal{Z} = \mathbb{R}$. Under our assumptions, it can be shown that this problem can be solved with Markov policies (the supremum over Markov policies is equal to the supremum over history-based policies).

\paragraph{Distributional RL.}
Distributional RL~\citep{morimura2010nonparametric,bellemare2023distributional} extends value-related classic RL results to return distributions, and provides methods to evaluate policies in terms of their (per-state, per-action) return distributions (whereas classic policy evaluation concerns scalar value functions).
On the practical side, a number of methods have been proposed for evaluating policies in a distributional sense, using function approximators, notably, categorical methods~\citep{bellemare2017distributional}, quantile regression~\citep{dabney2018distributional}, and particle-based methods~\citep{freirich2019distributional,wiltzer2024foundations}.

\paragraph{Return distribution optimization.}
The problem of return distribution optimization (RDO) was studied by \citet{marthe2023beyond,pires2025optimizing} as a family of RL-like objectives that we wish to optimize with dynamic programming.
These objectives are expressed in terms of functionals of distributions (besides the expectation), and a special case of RDO is to optimize the \emph{expected utility} objective:
\(
    \sup_{\pi \in \Pi_{\mathrm{H}}} \mathbb{E}[u(G^\pi_0)],
\)
where $u : \mathcal{Z} \rightarrow \mathbb{R}$ is a \emph{utility}.
This problem cannot be solved with Markov policies except for a small class of utilities (identity and exponential functions)~\citep{marthe2023beyond}.
Informally, the state does not contain enough information for making correct decisions, and some kind of memory is required.

\citet{pires2025optimizing} proposed to address this shortcoming by leveraging a summary of the history that they called \emph{stock}, here denoted $Z_t \in \mathcal{Z}$ (and defined below), and which is incorporated into the objective being optimized and into the states, yielding \emph{stock-augmented} RDO.
Henceforth, we only consider all RDO problems to be stock-augmented.

We call the special case of RDO with expected utilities as \emph{utility optimization} (UO):
\begin{equation}
    \sup_{\pi \in \Pi_{\mathrm{H}}} \mathbb{E}[u(Z_0 + G^\pi_0)],
    \label{eq:uo-objective}
\end{equation}
where $Z_0$ is an \emph{initial stock} from the initial stock-augmented state $(S_0, Z_0)$.
The stock $Z_t$ is defined recursively for $t \in \mathbb{N}_0$ as:
\begin{equation}
    Z_{t+1} \doteq \gamma^{-1}\left(Z_t + R_{t+1}\right).
    \label{eq:discounted-stock}
\end{equation}
\citet{pires2025optimizing} showed that \cref{eq:uo-objective} can be solved with policies that are Markov with respect to stock-augmented states when $x \mapsto u(x) - u(0)$ is positively homogeneous\footnote{We say a function $f : \mathcal{Z} \rightarrow \mathbb{R}$ is \emph{positively homogeneous} if there exists $t \in \mathbb{R}$ such that for all $\alpha > 0$ and $x \in \mathcal{Z}$, we have $f(\alpha x) = \alpha^t \, f(x)$.} and Lipschitz---a broader class than in the case without stock augmentation.

\paragraph{Constrained Markov decision processes.}
Informally, the constrained Markov decision process (CMDP)~\citep{altman2021constrained} problem is to optimize expected return subject to constraints on expected costs.
Normally, these quantities are denoted separately, but it is convenient for us to combine the reward and costs into a vector of cumulants where, in the case of CMDPs (but not in general), the first coordinate corresponds to the reward, and the others to costs.
Formulated this way, the CMDP problem is:
\[
    \sup_{\pi \in \Pi_{\mathrm{H}}} \mathbb{E}[G^\pi_{0, 1}] \quad \mbox{ subject to } \quad \mathbb{E}[G^\pi_{0, i}] \leq v_i \quad (2 \leq i \leq m),
\]
where $\mathcal{Z} = \mathbb{R}^m$ and $v \in \mathcal{Z}$ is a vector of \emph{constraint limits} ($v_1$ is irrelevant). 
From \citet[Theorem 2.1,][p.~26]{altman2021constrained}, it can be shown that an optimal Markov policy exists for the constrained RL problem.

There are multiple solution methods for CMDPs; in this work, we focus on the \emph{Lagrangian approach},
which consists of solving the \emph{dual problem}
\begin{equation}
    \inf_{\lambda \geq 0}\sup_{\pi \in \Pi_{\mathrm{H}}} \left(\mathbb{E}[G^\pi_{0, 1}] - \sum_{i=2}^m \lambda_i \left(\mathbb{E}[G^\pi_{0,i}] - v_i \right)\right).
    \label{eq:cmdp-dual}
\end{equation}
For fixed $\lambda$, the inner optimization can be cast as an RL problem with a \emph{scalarized reward} $R_{t, 1} - \sum_{i=2}^m \lambda_i R_{t,i}$ and solved with standard methods. 

\paragraph{Utility-constrained Markov decision processes.}
\citet{kadota2006discounted} generalized the CMDP formulation so that the objective and constraints can be utilities of the return distribution, namely, the UCMDP problem:
\begin{equation}
    \sup_{\pi \in \Pi_{\mathrm{H}}} \mathbb{E}[f(G^\pi_0)] \quad \mbox{ subject to } \quad \mathbb{E}[g(G^\pi_0)] \leq 0,
    \label{eq:ucmdp-problem-no-stock}
\end{equation}
where $f: \mathcal{Z} \rightarrow \mathbb{R}$ is objective function, $g: \mathcal{Z} \rightarrow \mathbb{R}^n$ is the constraint function (for some $n \in \mathbb{N}$, with constraint limits folded into $g$).
\citet{kadota2006discounted} studied the existence of a constrained optimal history-based policy, and the result of \citet{pires2025optimizing} for the unconstrained case suggests that, without stock augmentation, we may not be able to tackle the UCMDP problem using Markov policies, except for a limited family of objective and constraint utilities.

\section{Related Work}
\label{sec:related-work}

\paragraph{CMDPs.}
The challenge of learning optimal policies while adhering to safety or operational limits has been typically addressed using CMDPs, with different solution strategies. 

Primal-dual methods are the most common approach to solve CMDPs, converting the constrained problem into an unconstrained one by penalizing constraint violations using adaptive Lagrange multipliers. 
Simple Lagrangian-based algorithms have been observed to perform well empirically, matching or exceeding more sophisticated approaches in earlier benchmark evaluations~\citep{ray2019benchmarking}. 
Several works have improved on the basic primal-dual methodology, in the context of constrained RL: RCPO uses a multi-timescale approach to guarantee convergence to a feasible fixed point~\citep{tessler2018reward}, while PID Lagrangian employs control-theoretic methods for more stable multiplier updates with less Lagrange multiplier oscillation~\citep{stooke2020responsive}.

The so-called primal approaches, in contrast, do not introduce Lagrange multipliers and instead operate directly on the constrained optimization problem. These methods improve the reward while using constraint functions to remain within the feasible region. The seminal work in this category is CPO, which extends trust-region methods by ensuring each policy update satisfies constraint upper-bounds~\citep{achiam2017constrained}. Subsequent primal methods have explored different update mechanisms: PCPO alternates between reward-maximizing and projection steps~\citep{yang2020projection}, FOCOPS simplifies the update by first solving in a non-parametric space~\citep{zhang2020first}, and CRPO alternates between reward-maximizing and constraint-satisfying updates based on constraint status, all without using dual variables~\citep{xu2021crpo}.

Beyond these optimization-based approaches, other methods have been explored. Lyapunov-based methods, inspired by control theory, define a function estimating the potential for future constraint violation and guarantee cumulative constraint satisfaction by ensuring the policy does not increase its expected value~\citep{chow2018lyapunov}. Another line of work reformulates constraints, for instance by converting cumulative constraints into instantaneous, state-based constraints using Backward Value Functions~\citep{satija2020constrained}, or by casting safe RL as a probabilistic inference problem and extending MPO \citep{abdolmaleki2018maximum} to the constrained setting~\citep{liu2022constrained}. 

\paragraph{Beyond CMDPs.}
As mentioned, CMDPs suffer from the limitation of risk-neutral constraints.
\citet{kadota2006discounted} introduced UCMDPs and studied the existence of optimal policies.
These were history-based policies, which can be intractable to learn and are, we argue, impractical. 
A number of works have attempted to remedy the risk-neutral constraints in CMDPs, without explicitly adopting the UCMDP formulation.
However, in these works, state augmentation has proved a crucial component for performance.
Early work by \citet{xu2011probabilistic} demonstrated that chance-constrained MDPs could be solved by augmenting the state with accumulated reward, 
and \citet{chow2018risk} extended this idea to static Conditional Value-at-Risk (CVaR) constraints by tracking cumulative constraint cost.
Building on these ideas, Saut\'e RL~\citep{sootla2022saute} and Simmer~\citep{sootla2022effects} augmented the state with discounted stock to guide decision-making, leading to improved constraint satisfaction and safer policies, while \citet{sun2021safe} used the undiscounted stock to implement an early termination mechanism that enforces constraints. Budgeted MDPs~\citep{boutilier2016budget,carrara2019budgeted} also condition the policy on a budget, but the budget limits expected cost and is assigned by a dynamic program or selected by the policy, whereas our stock adds only realized cost. \cref{app:comparison-state-augmented} contains a more detailed review of related work, comparing the design of existing methods and ours.

Distributional RL methods, which learn the full distribution of returns rather than just expectations~\citep{morimura2010nonparametric,bellemare2017distributional,dabney2018distributional}, offer another avenue to go beyond risk-neutral constraints.  These techniques enable a range of risk-aware constraints: constraining quantiles~\citep{jung2022quantile} or spectral risk measures~\citep{kim2024spectral} of the cost distribution, constraining the cost-to-go CVaR~\citep{yang2021wcsac,yang2023safety}, and leveraging Extreme Value Theory to target high-impact tail events that expectation-based methods often miss~\citep{gao2025extreme}.
In this work, we leverage distributional estimates to build the policy improvement criterion for our agent.

\section{Methodology}
\label{sec:methodology}
Given the practical effectiveness of stock augmentation for constrained problems and its effectiveness for unconstrained UO problems, we adopt the stock-augmented formulation of the UCMDP problem (\emph{cf.}~\cref{eq:ucmdp-problem-no-stock}):
\begin{equation}
    \sup_{\pi \in \Pi_{\mathrm{H}}} \mathbb{E}[f(Z_0 + G^\pi_0)] \quad \mbox{ subject to } \quad \mathbb{E}[g(Z_0 + G^\pi_0)] \leq 0,
    \label{eq:ucmdp-problem}
\end{equation}
where $f, g : \mathcal{Z} \rightarrow \mathbb{R}$ are objective and constraint utilities, respectively\footnote{In general, $g$ will have a vector-valued co-domain, but we restrict our presentation to the scalar case for simplicity (and because the extension to the vector-valued case is straightforward).}.
We propose to solve this problem with a primal-dual actor-critic method that we call \emph{utility-constrained policies} (UCP).
UCP optimizes the following objective (\emph{cf.}~\cref{eq:cmdp-dual}) for $z \in \mathcal{Z}$:
\begin{equation}
    \inf_{\lambda_{z} \geq 0}\sup_{\pi \in \Pi_{\mathrm{H}}} \left( \vphantom{\sum} \mathbb{E}[f(Z_0 + G^\pi_0)] - \lambda_{z} \, \mathbb{E}[g(Z_0 + G^\pi_0)] \right) \quad (Z_0 = z).
    \label{eq:cmdp-dual-objective}
\end{equation}
For a fixed $\lambda_{z}$, the dual problem in \cref{eq:cmdp-dual-objective} reduces to UO with $u(x) = f(x) - \lambda_{z} g(x)$. 
We repurpose SAC~\citep{haarnoja2018soft} for solving UO following the rationale presented by \citet[Section 6]{pires2025optimizing}, which amounts to modifying SAC so that policy improvement maximizes an estimate of the UO objective, rather than an estimate of the RL objective (Q-values).

The actor in SAC optimizes the RL objective via the reparameterization trick:
\[
    L^{\mathrm{RL}}_{\mathrm{actor}}(s, \theta) = Q_\psi(s, A_\theta) - \alpha \log \pi_\theta(A_\theta|s) \tag{$A_{\theta} \doteq \mathrm{tanh}(\mu_\theta(s) + \sigma_\theta(s) \, \varepsilon)$}
\]
where $Q_\psi$ is the critic's action-value function estimator with parameters $\psi$,
$\theta$ are the actor parameters,
and $\varepsilon$ is sampled from a standard Normal distribution.
The critic parameters in SAC are learned through least squares regression, similar to fitted Q iteration~\citep{ernst2005tree}.

For UCP, we replace $Q_\psi$ with our estimate of the UO objective:
\[
    L^{\mathrm{UO}}_{\mathrm{actor}}(s, z, \theta) = \mathbb{E}[u(z + G_\psi(s, z, A_\theta))] - \alpha \log \pi_\theta(A_\theta|s, z) \tag{$A_{\theta} \doteq \mathrm{tanh}(\mu_\theta(s, z) + \sigma_\theta(s, z) \, \varepsilon)$}
\]
where the $\mathrm{df}(G_\psi(s, z, a))$ is given by a distributional critic estimating return distributions through quantile regression~\citep{dabney2018distributional}.

We also need to optimize $\lambda_{z}$ in order to solve the dual problem in \cref{eq:cmdp-dual-objective}.
To do so, we generate episodes periodically throughout training starting from $(S_0, z)$, observe the $n$ independent samples of the $G^\pi_0$, denoted by $(G^{(1)}_0, \ldots, G^{(n)}_0)$, and update the Lagrange multiplier via
 $\lambda_z \leftarrow \left( \lambda_z + \eta_\lambda \left( \frac{1}{n} \sum_{i=1}^n  g\left(z + G^{(i)}_0\right) - \epsilon \right) \right)_+$
where $\eta_\lambda$ is a learning rate and $\epsilon > 0$ is a small tolerance\footnote{The constraint $\mathbb{E}[ g(z + G_0) ] \leq 0$ may not be strictly feasible~\citep[Section 5.2.3, p.~226]{boyd2004convex} (for example, when $g(z) = (z_1)_+$, as we consider for our experiments), making the optimization more challenging.
In addition, imperfect return distribution estimates further complicate the problem. Introducing a tolerance mitigates both issues and has the practical benefit that $\lambda_z$ decreases when the constraint is satisfied (unlike the case $\epsilon = 0$).
In \cref{app:sensitivity_to_eps} we present a brief empirical study of the sensitivity of UCP to the choice of $\epsilon$.}.
In practice, we employ a function approximator to maintain $\lambda_z$ for $z \in \mathcal{Z}$ (see \cref{app:implementation}).

In our formulation, $-Z_{0,\geq2}$ acts as the effective constraint limits, and a desirable property is to be able to change this value to obtain different constrained behavior from the same optimal UCMDP policy, without needing to retrain it (similar to the generalization across safety budgets discussed by \citet[p.~7]{sootla2022saute}). Therefore, during training, we sample the initial stock $Z_0$ uniformly at random from a range of stock values that we wish the agent to learn to cope with. Note that the dependency of $\lambda_{z}$ on $z$ in \cref{eq:cmdp-dual-objective} is important to retain this flexibility to constraint limits. In some settings, this randomization may also introduce additional, helpful diversity to the training data~\citep{pires2025optimizing}---it may increase coverage of the augmented-state space. 

Besides sampling $Z_0$ at the beginning of each episode, we diverge from \citet{pires2025optimizing} and decouple the discounting used in the agent's (future-looking) returns from the one used in the agent's (past-looking) stock updates. 
More specifically, we use an \emph{undiscounted stock}: $\bar{Z}_{t+1} = \bar{Z}_t + R_{t+1}$.
(We write $\bar{Z}_t$ to distinguish it from discounted stock and more clearly indicate where this stock update is used, but we do not modify the notation for $\bar{Z}_0$ since $\bar{Z}_0 = Z_0$.)
We have found out that, empirically, this leads to better performance than using the discounted stock from \cref{eq:discounted-stock} (see the comparison in \cref{sec:main_text_ablation}). 
We attribute this improvement to two factors. First, in standard constrained RL, constraint limits are defined over the undiscounted cumulative cost of an episode. Using an undiscounted stock directly aligns the agent's state with this actual evaluation metric, allowing the stock to act as a tracker of the remaining budget regardless of how much time has passed. Second, the reverse-discounted update exponentially inflates early costs over time. This creates a larger range of stock values that the agent must learn to control for, which can make early training too hard. 
By keeping the stock undiscounted, we prevent this expansion of the augmented state space, leading to an easier learning problem.

\section{Experiments}
\label{sec:experiments}

In this section, we show that our practical approach to UCMDPs benefits from the expressivity of utility constraints, while attaining strong benchmark performance, in some cases with substantial performance gains. We evaluate UCP in the Safety Gymnasium benchmark~\citep{ji2023safety}, a suite of tasks where the agent must maximize expected return, while satisfying constraints on a scalar \emph{cost} signal (\citet{ji2023safety} stipulate a cost limit of $25$ in their evaluations).
In this paper, we consider only the single-agent tasks of the benchmark.
With the cumulants in ``(reward, cost)'' form (thus, $\mathcal{Z} = \mathbb{R}^2$), we frame the constrained RL problem in the benchmark as a UCMDP (see \cref{eq:ucmdp-problem}) with $f(x) = x_1$ (maximize expected return), $g(x) = (x_2)_+$ (subject to not exceeding the cost limit), and $Z_{0, 2} = -b$ for evaluation, where $b$ is some budget that we can select at evaluation time to maximize performance, as long as the expected cost does not exceed the benchmark's limit of $25$.
In this case, the constraint in \cref{eq:ucmdp-problem} becomes $\mathbb{E}[(G_{0,c}-b)_+] \leq  0$. 
It targets cost realizations above $b$, weighted by their excess, while leaving already-safe behavior below $b$ unrestricted.
In terms of theoretical guarantees,
we know from the results by \citet{pires2025optimizing} that, with the above choices of $f$ and $g$, the utility in the inner optimization of the dual problem (\cref{eq:cmdp-dual-objective}) can be optimized via dynamic programming, regardless of $\lambda$, since it is Lipschitz and positively homogeneous.

In a constrained RL setting, there are two important criteria for comparing policy performance: Whether they satisfy the constraint, and, among those that do, which has the best performance in terms of the objective.
In Safety Gymnasium, we first assess whether the cost is less than $25$ in expectation, and then compare expected returns.

We compare UCP to existing baselines in \cref{sec:performance}, where it matches or outperforms them on nearly all benchmark tasks. We also empirically demonstrate two key capabilities of stock-augmented UCMDPs: satisfying different constraint limits without retraining and better controlling the tails of cost distributions (\cref{sec:main_text_ablation}). Implementation and hyperparameter tuning details are provided in \cref{app:implementation}, and the code is available at \url{https://github.com/MehrdadMoghimi/UCP}.

\subsection{Benchmark Performance}
\label{sec:performance}

Our comparison considers several baselines from previous work (reproduced using OmniSafe~\citep{ji2024omnisafe}), in the \emph{Safety Velocity} and \emph{Safety Navigation} task groups.\footnote{For ease of comparison, the results in this section can be found in table format (\cref{tab:comparison_combined}) in \cref{app:tables}. All intervals are 95\% Student's $t$ confidence intervals over 5 training runs. For each seed, we average the performance of $1000$ episodes. 
Following common practice, we report \emph{undiscounted} episodic returns and costs.}
These baselines combine ``base'' RL methods---PPO~\citep{schulman2017proximal}, 
SAC~\citep{haarnoja2018soft} and 
TD3~\citep{fujimoto2018addressing}---with techniques for converting RL methods into constrained RL methods---Lagrangian optimization~\citep[Section 5.2]{ray2019benchmarking} (abbreviated Lag),
PID-Lagrangian~\citep{stooke2020responsive},
Saut\'e RL~\citep{sootla2022saute} and 
early terminated MDPs~\citep{sun2021safe} (abbreviated ET).

In UCP, the initial cost stock ($Z_{0, 2}$, the negative of the cost budget) is sampled uniformly at random from $[-30, 0]$, and the cost budget for evaluation is set after training.
We explore setting different cost budgets in \cref{sec:main_text_ablation}, but for the plots in this section, we have selected the best-performing $z_0$ per task from $\{0, -5, -15, -25\}$, provided that the resulting policy has expected cost below $25$ (with 95\% confidence over seeds).
We allowed ourselves this choice because we expect that, in practice, UCP constraint limits will be adjusted after training.

\begin{figure}[h]
  \centering
  \includegraphics[width=\linewidth]{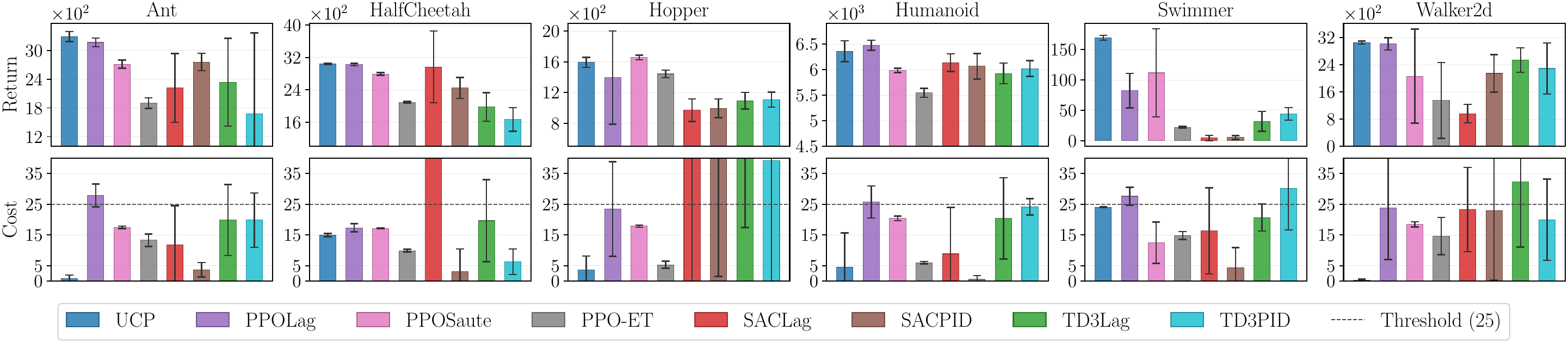}
  \caption{Comparison of UCP and OmniSafe baselines in the \emph{Safety Velocity} tasks.
}
  \label{fig:comparison_velocity}
\end{figure}

\begin{figure}[h]
  \centering
  \begin{subfigure}{\linewidth}
    \centering
    \includegraphics[width=\linewidth]{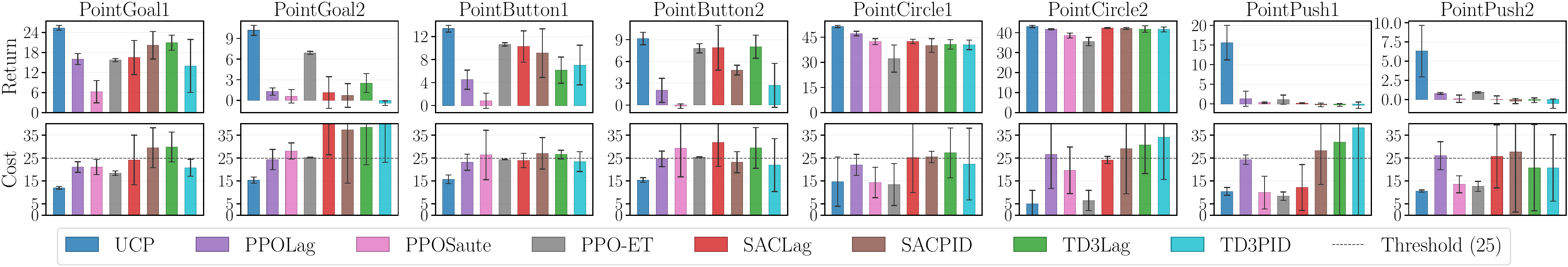}
    \label{fig:comparison_navigation_point}
  \end{subfigure}
  \hfill
  \begin{subfigure}{\linewidth}
    \centering
    \includegraphics[width=\linewidth]{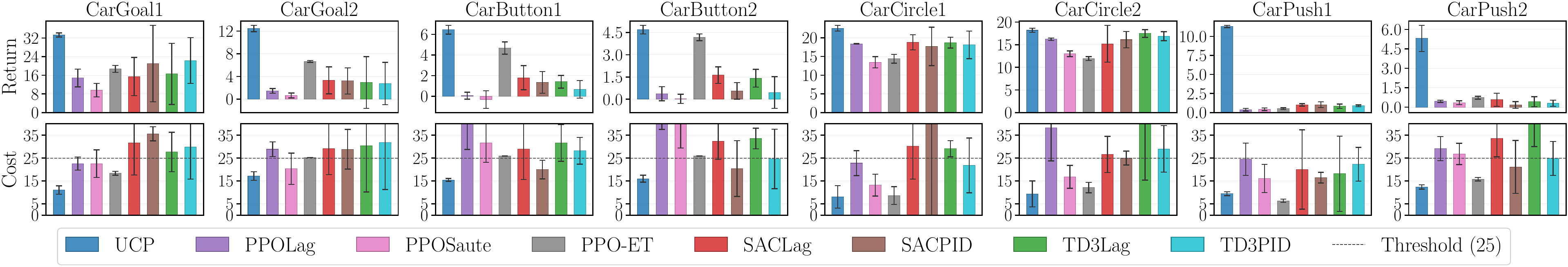}
    \label{fig:comparison_navigation_car}
  \end{subfigure}
  \caption{Comparison of UCP and OmniSafe baselines in the \emph{Safety Navigation} tasks. 
}
  \label{fig:comparison_navigation}
\end{figure}

We report the estimates of expected (undiscounted) return and expected (undiscounted) cost for UCP and the baselines, in \emph{Safety Velocity} (\cref{fig:comparison_velocity}) and \emph{Safety Navigation} (\cref{fig:comparison_navigation}) tasks. 
UCP matches or outperforms the other methods in terms of expected returns, while satisfying the cost constraint throughout. 
Interestingly, in most of the \emph{Safety Velocity} tasks, the cost has little bearing on the performance of UCP---for example, in Ant and Walker2d, UCP outperforms the other methods without incurring almost any cost. 
While there are relatively minor performance gains compared to the best baseline in these tasks, by evaluating UCP with different budgets, we find that modifying the cost budget has little \emph{marginal utility} for most of the safety velocity tasks, that is, modifying the cost budget does not significantly change the objective (we explore this in \cref{sec:main_text_ablation}).

In the \emph{Safety Navigation} tasks, we observe substantially larger gains.
Moreover, in these tasks, we observe that selecting the cost budget clearly has marginal utility (see \cref{tab:ctx_orig_combined} in \cref{app:tables}).
Accordingly, we report results at the commonly used budget of $25$. 
While our extensive hyperparameter tuning improved the OmniSafe baselines in several cases (such as the Button tasks), UCP still inherently benefits from certain architectural choices (e.g., distributional critics). To transparently isolate the performance gains of our core proposed components from these standard RL techniques, we conduct a detailed ablation study in \cref{sec:main_text_ablation}.

\subsection{Multiple Constraints}
\label{sec:multiple_constraints}

\Cref{sec:performance} follows the standard Safety Gymnasium evaluation in which the environments return a single scalar cost per step~\citep{ji2023safety}.
Underneath, this scalar aggregates multiple cost components, such as hazards, vase contact, and vase velocity in CarGoal2.
Since UCMDPs permit vector-valued costs, we also evaluate MC-UCP (multi-constraint UCP), with one stock coordinate, one distributional cost critic, and one Lagrange multiplier per cost component.
We train MC-UCP for the same number of steps as UCP and evaluate it at different initial cost stocks, which reveals the reward--cost trade-off of a task more clearly.
For example, as \cref{tab:multiconstraint_cargoal2_main} shows for CarGoal2, relaxing only hazards raises hazard cost from $2.66$ to $5.89$, and relaxing only vase velocity raises that cost from $1.67$ to $5.16$, whereas relaxing only vase contact, which is rare, barely changes its cost.
Relaxing several coordinates jointly also yields a large gain in return.
With $z_{0, (2, 3, 4)} = (-15, -15, -15)$, MC-UCP attains return $14.15$ at a total cost of $12.45$, whereas UCP attains $12.5$ at a total cost of $17.2$ (\cref{tab:comparison_combined}).
\Cref{app:multiconstraint} presents the full results, including CarButton2 and CarPush2.
\begin{table}[htb]
\centering
\caption{MC-UCP on CarGoal2 at selected initial cost stocks. Total cost is the benchmark's scalar cost, which is not directly constrained. Intervals are 95\% Student's $t$ intervals across five seeds.}
\label{tab:multiconstraint_cargoal2_main}
\begin{adjustbox}{max width=\linewidth,center}
\begin{tabular}{lccccc}
\toprule
$(z_{0, 2},z_{0, 3},z_{0, 4})$ & Return & Total cost & Hazards & Vases Contact & Vases Velocity \\
\midrule
$(0, 0, 0)$ & $-0.33\pm0.57$ & \textcolor{red}{$4.30\pm2.18$} & \textcolor{red}{$2.66\pm1.74$} & \textcolor{red}{$0.34\pm0.16$} & \textcolor{red}{$1.67\pm0.47$} \\
\midrule
$(-25, 0, 0)$ & $2.36\pm1.30$ & \textcolor{red}{$7.40\pm0.99$} & \textcolor{red}{$5.89\pm0.82$} & \textcolor{red}{$0.29\pm0.13$} & \textcolor{red}{$1.52\pm0.23$} \\
$(0, -25, 0)$ & $2.17\pm1.65$ & \textcolor{red}{$3.38\pm1.27$} & \textcolor{red}{$1.38\pm0.39$} & \textcolor{red}{$0.38\pm0.22$} & \textcolor{red}{$1.99\pm0.89$} \\
$(0, 0, -25)$ & $6.58\pm2.05$ & \textcolor{red}{$7.53\pm0.88$} & \textcolor{red}{$2.36\pm0.69$} & \textcolor{red}{$0.87\pm0.17$} & \textcolor{red}{$5.16\pm0.28$} \\
\midrule
$(-25, 0, -25)$ & $13.30\pm1.60$ & \textcolor{red}{$19.25\pm2.01$} & \textcolor{red}{$12.57\pm0.60$} & \textcolor{red}{$0.98\pm0.39$} & \textcolor{red}{$6.76\pm1.67$} \\
\midrule
$(-5, -5, -5)$ & $5.24\pm1.16$ & \textcolor{red}{$2.82\pm0.32$} & \textcolor{red}{$1.42\pm0.43$} & \textcolor{red}{$0.22\pm0.03$} & \textcolor{red}{$1.41\pm0.13$} \\
$(-15, -15, -15)$ & $14.15\pm0.42$ & \textcolor{red}{$12.45\pm1.69$} & \textcolor{red}{$6.55\pm0.58$} & \textcolor{red}{$0.82\pm0.21$} & \textcolor{red}{$5.92\pm1.41$} \\
$(-25, -25, -25)$ & $16.36\pm0.68$ & \textcolor{red}{$27.85\pm2.39$} & \textcolor{red}{$13.72\pm0.83$} & \textcolor{red}{$2.19\pm0.46$} & \textcolor{red}{$14.25\pm2.45$} \\
\bottomrule
\end{tabular}
\end{adjustbox}
\end{table}

\subsection{Ablation Study}
\label{sec:main_text_ablation}

To better understand how each component of our method contributes to its empirical performance, we conduct a comprehensive ablation study with the components summarized in \cref{tab:ablation_components}. 

\begin{table}[H]
  \centering
  \caption{Ablation components for all evaluated variants.}
  \label{tab:ablation_components}
  \begin{tabular}{lcccc}
    \toprule
    Method & Utility $g(x)$ & \makecell{Stock \\ Augmentation} & \makecell{Stock \\ Randomization} & \makecell{Undiscounted Stock} \\
    \midrule
    UCP & $(x_2)_+$ & $\checkmark$ & $\checkmark$ & $\checkmark$ \\
    UCP-DS & $(x_2)_+$ & $\checkmark$ & $\checkmark$ & $\times$ \\
    UCP-NR & $(x_2)_+$ & $\checkmark$ & $\times$ & $\checkmark$ \\
    UCP-Mean & $x_2$ & $\checkmark$ & $\checkmark$ & $\checkmark$ \\
    UCP-NR-Mean & $x_2$ & $\checkmark$ & $\times$ & $\checkmark$ \\
    UCP-NA & $x_2$ & $\times$ & N/A & N/A \\
    \bottomrule
  \end{tabular}
\end{table}

We evaluate the following variants: UCP-NA (No Augmentation), a Lagrangian RL baseline for CMDPs; UCP-NR-Mean (No Randomization, Mean Utility), which adds stock augmentation to study the effect of the larger state space; UCP-Mean, which further adds initial stock randomization while retaining a risk-neutral constraint; UCP-NR (No Randomization), which uses a risk-sensitive constraint but omits stock randomization, highlighting the role of randomization in generalization; UCP-DS (Discounted Stock), which combines risk sensitivity and randomization but uses discounted instead of undiscounted stock to assess this design choice; and UCP, with all components.

We report the results for all methods and tasks across four budget levels, as a way to understand the impact of the budget selection in task performance. UCP-NR, UCP-NR-Mean, and UCP-NA are trained separately for each budget level, whereas UCP-Mean and UCP are trained once with the same number of timesteps and evaluated across all four budgets. UCP-DS performed substantially worse than all other variants; therefore, its results are reported separately in \cref{app:tables}, \cref{tab:ctx_cgamma_combined}. For brevity, we discuss CarGoal1 results here, with full results across all environments presented in \cref{app:ablations}.

\begin{figure}[h]
  \centering
  \begin{subfigure}{0.49\textwidth}
    \centering
    \includegraphics[width=\textwidth]{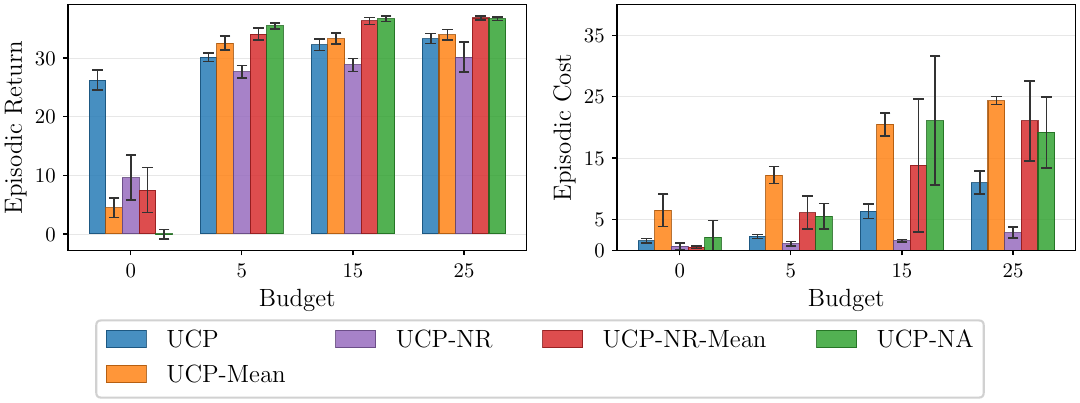}
    \caption{Mean episodic return and cost across four cost budget levels ($0$, $5$, $15$, $25$).}
    \label{fig:cargoal1_barplot}
  \end{subfigure}
  \hfill
  \begin{subfigure}{0.49\textwidth}
    \centering
    \includegraphics[width=\linewidth]{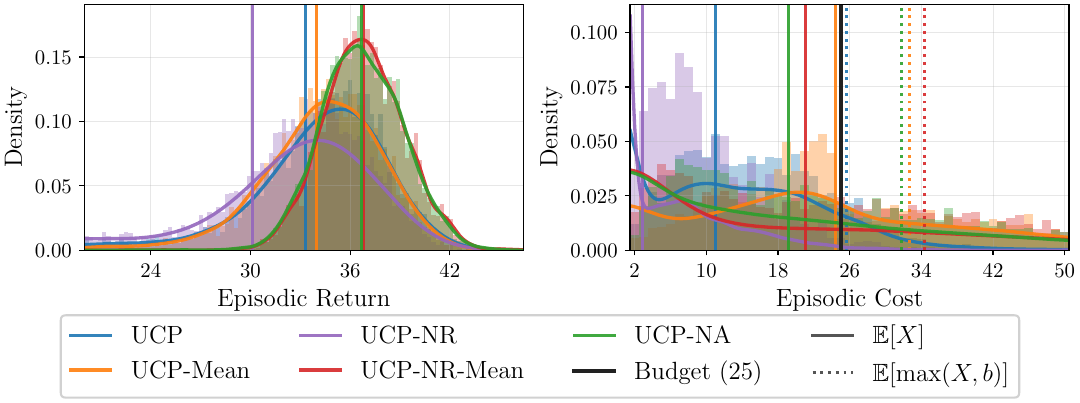}
    \caption{Empirical distributions of episodic return and cost at cost budget of $25$ ($5$ seeds, $1000$ episodes).}
    \label{fig:cargoal1_histogram}
  \end{subfigure}
  \caption{Ablation study on CarGoal1: risk-sensitive utility reduces budget violations, and initial stock randomization enables cross-budget generalization without retraining.}
  \label{fig:both}
\end{figure}

By comparing UCP-NA and UCP-NR-Mean in \cref{fig:cargoal1_barplot}, we observe that stock augmentation does not degrade performance and can even improve it, particularly under the tightest budget of 0. 
Next, comparing UCP and UCP-Mean to their non-randomized counterparts (UCP-NR and UCP-NR-Mean) shows that randomizing the initial stock enables generalization across budget levels without retraining, albeit with different effects. Relative to UCP-NR, UCP achieves a higher return at the cost of slightly increased constraint violations. This is expected, as UCP-NR is trained separately for each budget level and can learn more precise value estimates to satisfy the constraint. However, since the cost under UCP remains below the allocated budget within the 95\% confidence interval, UCP is the preferred approach overall. In contrast, UCP-Mean does not provide a similar benefit: It does not improve return and, in fact, incurs a higher cost than UCP-NR-Mean.
The worse performance of UCP-NR-Mean and UCP-Mean relative to UCP in some cases may be surprising, considering that UCP is optimizing for stricter constraints than the benchmark's. However, this is in line with findings from previous work, where stricter constraints improved performance in the risk-neutral sense (see \cref{sec:intro}).
Taken together, these results indicate that UCP achieves the best balance between return and cost among the considered variants. Importantly, this behavior is not specific to CarGoal1 and, as discussed in \cref{app:ablations}, is consistently observed across other environments. Finally, at the tightest budget of 0, both UCP and UCP-NR achieve high return with near-zero cost, highlighting the effectiveness of the risk-sensitive utility. Notably, UCP attains significantly higher return without a proportional increase in cost, which suggests that the combination of initial stock randomization and risk-sensitive utility can provide an effective setup for strong performance under tight constraints.

To further investigate the effect of these components, \cref{fig:cargoal1_histogram} shows the distribution of return and cost across all evaluation trajectories (5 seeds, 1000 episodes each) at budget $25$. The distributions clearly show that the risk-sensitive utility reduces the probability of exceeding the budget. 
\begin{table}[h]
  \centering
  \caption{Evaluation statistics on CarGoal1 with cost budget of $25$. 
  The third column shows the estimated conditional expectation of the undiscounted cost given that it (strictly) exceeds the budget, and the fourth to the estimated probability of the undiscounted cost exceeding the budget.}
  \label{tab:hist_nav_c25_main}
  \begin{tabular}{lccccc}
    \toprule
    Method & Return & Cost & Cond.~Expected Cost & Freq. of exceeding Budget \\
    \midrule
    UCP & $33.3\pm0.2$ & \textcolor{red}{$11.0\pm0.3$} & $31.4$ & $0.094$ \\
    UCP-Mean & $34.0\pm0.1$ & \textcolor{red}{$24.4\pm0.6$} & $46.0$ & $0.364$ \\
    UCP-NR & $30.1\pm0.2$ & \textcolor{red}{$2.9\pm0.2$} & $33.3$ & $0.007$ \\
    UCP-NR-Mean & $36.8\pm0.1$ & \textcolor{red}{$21.1\pm0.8$} & $53.6$ & $0.326$ \\
    UCP-NA & $36.7\pm0.1$ & \textcolor{red}{$19.2\pm0.6$} & $48.1$ & $0.292$ \\
    \bottomrule
  \end{tabular}
\end{table}
More precisely, \cref{tab:hist_nav_c25_main} shows that UCP-NR achieves a near-zero probability of budget violation, albeit with a reduction in return. UCP also reduces the probability of exceeding the budget, but we believe that the exploration from initial stock randomization mitigates the associated performance drop.

\section{Conclusion}
\label{sec:conclusion}

We introduced stock augmentation to unlock practical solutions for UCMDPs and to overcome limitations of risk-neutral constraints in CMDPs. As a bonus, constraint limits are included in the initial augmented state, allowing policies to be evaluated with different limits without retraining. 
To solve stock-augmented UCMDPs, we proposed UCP, a Lagrangian actor-critic method with an outer optimization similar to TRPO-Lagrangian and PPO-Lagrangian~\citep{ray2019benchmarking}, and an inner optimization consisting of three simple modifications to SAC~\citep{haarnoja2018soft,haarnoja2017reinforcement}: i) input undiscounted stock ($\bar{z}_t$) to actor and critic; ii) use a distributional critic (here, quantile regression~\citep{dabney2018distributional}) to estimate the cumulant distribution; and iii) train the actor to maximize the critic's expected-utility estimate instead of a Q-value estimate.

Despite strong empirical results, a key limitation is the lack of theoretical analysis. For instance, is there a principled reason to use undiscounted rather than discounted stock?  What optimality guarantees can be established for Lagrangian dynamic programming in UCMDPs, and can they inform improvements to UCP? Beyond theory, further improvements to UCP may reduce approximation and statistical error, yielding more precise constraint satisfaction. Our implementation is just one instance of a broader class of utility-constrained agents; the SAC-based changes can extend to other RL methods (e.g., PPO~\citep{schulman2017proximal}, MPO~\citep{abdolmaleki2018maximum}) and other distributional methods for the critic~\citep{bellemare2017distributional,wiltzer2024foundations,zhang2021distributional}. The outer optimization could also be improved, for example, with PID-Lagrangian~\citep{stooke2020responsive} updates.

\begin{ack}
We thank Mark Rowland for his review and feedback on this work. 
We thank the Digital Research Alliance of Canada (\href{https://alliancecan.ca/en}{alliancecan.ca}) for providing the computational resources used in this work.
We thank the NeurIPS reviewers, area chair and program committee for their review and feedback on this work.
\end{ack}

\bibliography{neurips_2026}
\bibliographystyle{plainnat}

\clearpage
\appendix
\crefalias{section}{appendix}

\section{Comparison with State-Augmented Constrained RL}
\label{app:comparison-state-augmented}

Augmenting the state space with a variable that tracks the accumulated safety cost has been explored in recent constrained RL literature. 
In this section, we review two prominent prior works, Saut\'e RL~\citep{sootla2022saute} and the reward penalty method by \citet{jiang2024reward}, to highlight the theoretical and practical distinctions of our stock-augmented distributional approach. 

Recall our notation for scalar reward and cost: The agent observes a task reward $R_{t, 1}$ and a safety cost $R_{t, 2}$. We aim to keep the cumulative cost below a threshold $b$. To achieve this, we augment the state with an undiscounted stock variable $\bar{z}_t \in \mathbb{R}$ related to the cost, initialized as $z_0 = -b$. The boundary for constraint violation is thus reached when $\bar{z}_t > 0$.

\subsection{Saut\'e RL}

Saut\'e RL~\citep{sootla2022saute} proposes to address the constrained RL problem by focusing on \emph{almost sure} (probability $1$) safety constraints. The core mechanism of Saut\'e RL relies on reshaping the immediate task reward based on the state of the (discounted) stock variable $z_t$.

Saut\'e RL defines a reshaped immediate reward $\tilde{R}_t$ as follows:
\begin{equation}
    \tilde{R}_t(s_t, z_t, a_t) = 
    \begin{cases}
        R_{t, 1} & \text{if } z_{t+1} \leq 0, \\
        -n & \text{if } z_{t+1} > 0,
    \end{cases}
\end{equation}
where $n$ is a large constant, with the goal of approximating the case with infinite penalty.

A fundamental limitation of this approach is that the almost-sure safety constraint can be problematic for dynamic programming (and likely for established deep RL methods).
\citet[Appendix C.2]{pires2025optimizing} show that dynamic programming cannot optimize indicator utilities in the infinite-horizon discounted case.
On the practical side, as noted by \citet{jiang2024reward}, the reward drops to a flat, large ($n$) penalty upon violation, regardless of the severity of the breach.
That is, all violations are equally bad, and the agent may receive a limited policy-improvement signal from violations.

It is also worth noting that \citet{sootla2022saute} use separate discount factors for reward and cost. Because they merge these signals into a single modified reward, this effectively introduces one discount factor for the stock and another for accumulating the modified reward. Our view differs: while our formulation could also allow separate discount factors per signal, we instead use a single discount factor for all signals. As discussed in \cref{sec:methodology}, our focus is on how, for a single signal, the agent should discount the backward-looking stock versus the forward-looking return estimate.

\subsection{Reward Penalties on Augmented States}

To overcome the limitations of Saut\'e RL, \citet{jiang2024reward} proposed a formulation that also uses state augmentation but employs a more structured reward penalty mechanism for ``soft'' constraints. 

Translated to our notation, assuming the stock variable tracks the undiscounted cost, \citet{jiang2024reward} apply a scaled Lagrangian penalty directly to the immediate reward, but \emph{only} when the trajectory becomes infeasible. Their modified reward is:
\begin{equation}
    \tilde{R}_t(s_t, z_t, a_t) = 
    \begin{cases}
        R_{t, 1} & \text{if } z_{t} \leq 0 \text{ and } z_{t+1} \leq 0, \\
        R_{t, 1} - \frac{\lambda}{\gamma^t} (z_{t+1}-z_{0}) & \text{if } z_{t} \leq 0 \text{ and } z_{t+1} > 0, \\
        R_{t, 1} - \frac{\lambda}{\gamma^t} R_{t, 2} & \text{if } z_{t} > 0,
    \end{cases}
\end{equation}
where $\lambda > 0$ is a penalty parameter. 

It is important to note that the term $(z_{t+1}-z_{0})$ in the middle condition corresponds to the total accumulated cost from the beginning of the episode. Consequently, the entirety of the historical accumulated cost is applied as a single-step penalty exactly at the transition where the safety threshold is breached. Because this formulation adds the trajectory's cumulative cost history into the reward of a single time step, the immediate reward function exhibits a sharp discontinuity at the boundary $z_t = 0$, which the standard value function must subsequently approximate.

Despite this discontinuity, this formulation provides a more principled penalty structure than Saut\'e RL. By scaling the penalty linearly with the actual cost incurred (rather than applying an arbitrary constant scalar $-n$), \citet{jiang2024reward} ensure that the objective function preserves the relative ordering of trajectories based on the magnitude of their constraint violations. This scaling, they claim, enables the optimization of ``soft'' constraints, such as expected cost and CVaR, without restricting the agent to strictly zero-violation policies like Saut\'e RL.

\subsection{UCP}

All stock-augmented methods, including Saut\'e RL and our own UCP, share the structural advantage of being able to generalize across different safety budgets by sampling $z_0$ during training. However, prior methods rely on manually designing heuristics to reshape the \emph{immediate scalar reward} based on the current stock $z_t$. Our methodology offers a more principled alternative using distributional RL.

Instead of using a modified step-by-step task reward, we maintain separate, unmodified signals for reward and cost. Our penalty is applied over the \emph{distribution} of the entire future cost return using the expected utility $\mathbb{E} \left[ g(Z_0 + G_0^{\pi}) \right]$. This distinction provides two key algorithmic advantages:
\begin{enumerate}
    \item In the frameworks of \citet{sootla2022saute} and \citet{jiang2024reward}, the non-linear safety penalty is injected directly into the environment's immediate reward. In contrast, our approach keeps the original step-by-step reward and cost signals. We use a distributional critic to model the true distribution of cumulative costs, and apply the utility function over the full distribution during the policy update. 
    \item Because prior methods combine the task reward and cost penalty into a single scalar $\tilde{R}_t$, the standard critic entangles two different signals. Thus, changing the penalty weight invalidates the learned value function.
     By separating the reward and cost critics, and handling the trade-off dynamically via the Lagrange multiplier $\lambda$ strictly at the policy optimization step, our method decouples the task gradient from the safety gradient and eliminates the need for piecewise reward engineering.
\end{enumerate} 

We emphasize that we augment the state with the \emph{undiscounted stock}, which we have found, empirically, to be a better choice than augmenting with the discounted stock (see \cref{sec:methodology}).

\section{Risk-neutral versus risk-sensitive constraints}
\label{app:illustration}

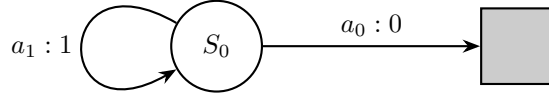
\begin{figure}[htb]
    \begin{center}
        \begin{tikzpicture}[
            >=Stealth,
            node distance=4cm,
            thick,
            auto
        ]

            \node[state, minimum size=1.2cm] (S0) {$S_0$};

            \node[draw, rectangle, fill=gray!40, minimum size=1cm, right of=S0] (box) {};

            \path[->] 
                (S0) edge [loop left, looseness=8, out=150, in=210] node {$a_1: 1$} (S0)
                (S0) edge node {$a_0: 0$} (box);
        
        \end{tikzpicture}
        \caption{Simple example MDP for illustrating differences between the CMDP and UCMDP frameworks.\label{fig:simple-mdp}}
    \end{center}
\end{figure}

In this section, we provide an illustrative example to motivate why we may want to frame a problem as a stock-augmented UCMDP, instead of a CMDP.
Consider the MDP in \cref{fig:simple-mdp}, where action $a_0$ transitions to the terminal state with reward $0$, and action $a_1$ stays in state $s_0$, with a reward of $1$.
Suppose we wish to maximize return, subject to it being at most $1$.

In CMDPs, we would maximize expected return subject to $\mathbb{E}[G^\pi_0] \leq 1$.
The solution to this problem is a stochastic policy taking action $a_1$ with probability $(1 + \gamma)^{-1}$.
This policy mixes overshooting and undershooting the return, but achieves an expected return of exactly one. 

This risk-neutral behavior, though optimal, can be undesirable: Undershooting the constraint comes at a cost in terms of the return, and overshooting it can be catastrophic.
Yet the policy is optimal, so we cannot do better unless we change the problem formulation in some way.

In the example MDP in \cref{fig:simple-mdp}, a history-based policy can deliver returns of exactly one with probability one.
A stock-augmented Markov policy can also do so.
Such a policy would be optimal in the stock-augmented CMDP problem, but so would the risk-neutral optimal policy of the original CMDP problem.
We can use the UCMDP framework to remedy that.

Consider the (stock-augmented) UCMDP problem with $f(z) = z$ and $g(z) = (z)_+$: The objective is to maximize expected return, subject to $\mathbb{E}[(Z_0 + G^\pi_0)_+] \leq 0$.
In this case, in $(s_0, z)$ it is optimal to choose $a_1$ for $z \leq -1$ and $a_0$ otherwise.
To satisfy the constraint on the return being at most $1$ with an optimal policy, we can set $Z_0 = -1$, as the optimal policy at $(s_0, -1)$ is to choose $a_1$, transition to $(s_0, 0)$, and then choose $a_0$.
This policy obtains a return of exactly one, with probability one, when starting from $(s_0, -1)$.
Moreover, the optimal stochastic CMDP policy described above is no longer optimal for this problem.

\section{Implementation Details}
\label{app:implementation}

Our implementation is extends the implementation of the SAC algorithm in LeanRL~\citep{huang2022cleanrl} with our proposed components.

\subsection{Pseudocode}

\Cref{alg:ucp} contains the pseudocode for UCP.

In the general UCMDP formulation (\cref{eq:ucmdp-problem}), the undiscounted stock $\bar{Z}_t\in\mathcal{Z}$ with $\mathcal{Z}\subseteq\mathbb{R}^m$ tracks all cumulants simultaneously. 
In the case of Safety Gymnasium, where $m = 2$ and $f$ is the \emph{first-coordinate projection function} (i.e.\ $f(z) = z_1$, so the objective is the expected return), the reward stock $z_{0, 1}$ enters the objective only as an additive constant $z_{0, 1} + G_{0, 1}$ whose expectation shifts the objective by $z_{0, 1}$; this does not affect the optimal policy. 
Consequently, we only augment the state with the \emph{cost} stock $\bar{Z}_{t, 2}$, since only $g(z_{0, 2}+G_{0, 2})$ depends non-trivially on the initial stock (having noted that we use the undiscounted stock $\bar{Z}_t$ instead of $Z_t$ for empirical reasons).
We implement stock augmentation by expanding the agent's observation space as described in \citet{pires2025optimizing}.

The cumulant distributions are estimated via quantile regression~\citep{dabney2018distributional}, which estimates the marginal distributions of the total discounted cumulants (reward and cost in the case of Safety Gymnasium), and estimating the marginals is enough for estimating the expected utility objective in our case. 
Quantile regression uses $N$ evenly-spaced cumulative probabilities $\tau_i = \frac{2i - 1}{2N}$ for $i \in \{1, \dots, N\}$, and the critic network is trained by minimizing the quantile regression loss across all pairwise combinations of quantiles and $n$-step cumulant estimates. For an $n$-step trajectory formed as $S_t$, $R_{t+1}, \ldots, R_{t+n}$, $(S_{t+n}, \bar{Z}_{t+n})$, we build $n$-step TD-error estimates as $\delta_{ij} \doteq \left(\sum_{k=0}^{n-1} \gamma^k R_{t+k+1} + \gamma^n Y'_j - Y_i\right)_{i=1,j=1}^{N, N}$, where $(Y_i)_{i=1}^N$ are the critic network outputs for the augmented state-action triple $(S_t, \bar{Z}_t, A_t)$ and $(Y'_j)_{j=1}^N$ are the target critic network outputs for the augmented state-action triple $(S_{t+n}, \bar{Z}_{t+n}, A_{t+n})$. We then minimize the quantile regression loss
\[
    \sum_{i=1}^N\sum_{j=1}^N\rho_{\tau_i}(\delta_{ij})=\sum_{i=1}^N\sum_{j=1}^N | \delta_{ij} | \, |\tau_i - \mathbf{1}_{\{\delta_{ij} < 0\}}|
\]
with respect to the critic network weights.
Note that the target bootstrap values $(Y'_j)_{j=1}^N$ for the returns come from a double-critic, to mitigate overestimation bias: Inspired by Soft Actor-Critic~\citep{haarnoja2018soft}, whenever selecting the target quantiles, we use all the quantiles from the critic network for which the mean over quantiles is smaller.
For the cost quantiles, we use a single critic.
We use target networks for the actor and the critics, and we update them via Polyak averaging~\citep{schwarzer2023bigger}.

Safety constraints are enforced by jointly optimizing the policy and a Lagrangian multiplier. The Lagrange multiplier, $\lambda(z_0)$, dynamically adjusts the penalty based on the initial cost stock (or budget) sampled at the beginning of each training episode.
We use function approximation to maintain the estimates of the Lagrange multipliers $\lambda_z$ for different stock values, via a $K$-bin discretization over $[z_{\mathrm{min}},z_{\mathrm{max}}]$.
In \cref{app:sensitivity_to_K} we present a brief empirical study of the sensitivity of UCP to the choice of $K$.

\begin{algorithm}[htbp]
\scriptsize 
\caption{Utility-Constrained Policies (UCP)}
\label{alg:ucp}
\begin{algorithmic}[1]

\Require
  Utilities $f,g:\mathbb{R}^2 \to \mathbb{R}$;
  Quantile levels $\tau_i=\tfrac{2i-1}{2N}$, $i\in[N]$;
  Learning rates $\eta_\theta,\eta_\psi,\eta_\phi,\eta_\lambda$;
  Regularization weight $\beta$;
  $n$-step horizon $n$; Discount factor $\gamma$; Batch size $B$;
  Polyak coefficient $\tau_{\mathrm{ema}}$;
  $K$ discretization levels for $\lambda$ over $[z_{\mathrm{min}},z_{\mathrm{max}}]$;
  Tolerance $\varepsilon \geq 0$.

\State \textbf{Initialize}
  Actor $\pi_\theta$;
  Twin reward critics $G^{(1)}_\psi,G^{(2)}_\psi$;
  Cost critic $G^c_\phi$;
  Target parameters $\bar{\psi} \leftarrow \psi$ and $\bar{\phi} \leftarrow \phi$

\State \textbf{Initialize}
  Discretized Lagrange Multipliers $\{\lambda_k\}_{k=1}^{K}$, each $\lambda_k \leftarrow \lambda_0 > 0$
  \Comment{%
    index $k(z_0) = \bigl\lfloor
    \tfrac{z_0-z_{\mathrm{min}}}{z_{\mathrm{max}}-z_{\mathrm{min}}}\, (K-1)
    \bigr\rfloor+1$}

\State \textbf{Initialize}
  Replay buffer $\mathcal{B} \leftarrow \emptyset$;
  Cost trajectory buffer $\mathcal{B}_c \leftarrow \emptyset$
\Comment{Fixed-size FIFO}

\algrule

\For{each environment step $t=0, 1, 2,\ldots$}

  \State At episode start: sample
    $z_0 \sim \mathrm{Uniform} \left(\left\{
      z_{\mathrm{min}},z_{\mathrm{min}} + \tfrac{z_{\mathrm{max}}-z_{\mathrm{min}}}{K-1},\ldots,z_{\mathrm{max}}
    \right\}\right)$
  \Comment{Initial stock randomization}

  \State $A_t \sim \pi_\theta(\,\cdot\mid S_t,\bar{Z}_t)$;\quad
         $(S_{t+1}, R_{t+1}, C_{t+1}) \sim  P(\,\cdot\mid S_t,A_t)$

  \State $\bar{Z}_{t+1}\leftarrow \bar{Z}_t + (R_{t+1},\,C_{t+1})$
  \Comment{undiscounted stock update}

  \State $\mathcal{B}\leftarrow\mathcal{B}\cup
    \{(S_t,\bar{Z}_t,A_t,R_{t+1},C_{t+1},S_{t+1},\bar{Z}_{t+1},z_{0, 2},\mathrm{done}_t)\}$
  \Comment{$z_{0, 2}$: episode-start cost stock, stored for multiplier lookup}

  \If{episode done}
    \State $\mathcal{B}_c\leftarrow\mathcal{B}_c\cup\{(z_{0, 2},G^c_0)\}$
    \Comment{$G^c_0$: episodic cost return}
  \EndIf

  \algrule

    \If{$|\mathcal{B}|\geq B$}

    \State Sample
      $\bigl\{(s_i,\bar{z}_i,a_i,r^{(n)}_i,c^{(n)}_i,s'_i,\bar{z}'_i,z_{0,i},\mathrm{done}_i)\bigr\}_{i=1}^{B}$
      from $\mathcal{B}$
    \Comment{$r^{(n)}_i,c^{(n)}_i = \sum_{l=0}^{n-1}\gamma^l(R_{t+l+1},C_{t+l+1})$}

    \algrule[.2pt]
    \Statex \hspace{22pt} \textit{\small\color{commentgray}Distributional Critic Updates}
    
    \State $\tilde{a}'_i \sim \bar{\pi}_\theta(\,\cdot\mid s'_i,\bar{z}'_i) \quad \forall i \in \{1,\dots,B\}$
    \Comment{sample next actions for all transitions in batch}

    \State $j^*_i = \displaystyle\arg\min_{j\in\{1, 2\}}
             \mathbb{E}_\tau \left[f \left(\bar{z}'_{i, 1}+ G^{(j)}_{\bar{\psi}}(s'_i,\bar{z}'_i,\tilde{a}'_i)\right)\right]$
    \Comment{select target critic by lower expected utility}

    \State $Y_i \leftarrow  r^{(n)}_i
             +\bigl(1 - \mathrm{done}_i\bigr)\gamma^n\,
             \bigl( G^{(j^*_i)}_{\bar{\psi}}(s'_i,\bar{z}'_i,\tilde{a}'_i) - \alpha \log \bar\pi_\theta(\tilde{a}'_i\mid s'_i, \bar{z}'_i)\bigr)$
    \Comment{distributional Bellman target}

    \For{$j\in\{1, 2\}$}
      \Comment{twin reward critics loss}
      \State $\mathcal{L}^{(j)}_Q(\psi) \leftarrow 
        \dfrac{1}{BN^2} \sum_{i=1}^{B}\sum_{j=1}^{N}\sum_{m=1}^{N}
        \rho_{\tau_j} \left(Y_{i,m}-G^{(j)}_{\bar{\psi}}(s_i,\bar{z}_i,a_i)_{\tau_j}\right)$
      \Comment{all-pairs quantile loss}
    \EndFor

    \State $\psi\leftarrow\psi-\eta_\psi\,\nabla_\psi \left(\mathcal{L}^{(1)}_Q+\mathcal{L}^{(2)}_Q\right)$

    \Statex

    \State $Y^c_i \leftarrow  c^{(n)}_i
           +\bigl(1 - \mathrm{done}_i\bigr)\gamma^n\, G^c_{\bar{\phi}}(s'_i,\bar{z}'_i,\tilde{a}'_i)$
    \Comment{cost target uses same action $\tilde{a}'_i$}

    \State $\mathcal{L}_C(\phi) \leftarrow 
      \dfrac{1}{BN^2} \sum_{i=1}^{B}\sum_{j=1}^{N}\sum_{m=1}^{N}
      \rho_{\tau_j} \left(Y^c_{i,m}-G^c_{\bar{\phi}}(s_i,z_i,a_i)_{\tau_j}\right)$

    \State $\phi\leftarrow\phi-\eta_\phi\,\nabla_\phi\,\mathcal{L}_C(\phi)$
    
    \algrule[.2pt]

    \Statex \hspace{22pt} \textit{\small\color{commentgray}Actor Update (SAC + Lagrangian with UO objective)}

    \State $\tilde{a}_i \sim \pi_\theta(\,\cdot\mid s_i,\bar{z}_i)$
    \Comment{reparameterization:
      $\tilde{a}_i = \tanh \left(\mu_\theta(s_i,\bar{z}_i)+\sigma_\theta(s_i,\bar{z}_i)\odot\varepsilon\right)$,
      $\varepsilon \sim \mathcal{N}(0,I)$}

    \State $\lambda_i \leftarrow \lambda_{k(z_{0,i})}$

    \State $U_i \leftarrow
           \mathbb{E}_\tau \left[f \left(\bar{z}_{i, 1}+G^{(j^*_i)}_{\bar{\psi}}(s_i,\bar{z}_i,\tilde{a}_i)\right)\right]$
    \Comment{$j^*_i$ from current-state critic selection}

    \State $U^c_i \leftarrow
           \mathbb{E}_\tau \left[g \left(\bar{z}_{i, 2}+G^c_{\bar{\phi}}(s_i,\bar{z},\tilde{a}_i)\right)\right]$
    \Comment{$\bar{z}_{i, 2}$ added to each quantile}

    \State $\mathcal{L}_{\mathrm{reg},i}\leftarrow
           \beta \left(\|\mu_\theta(s_i,\bar{z}_i)\|^2+\|\log\sigma_\theta(s_i,\bar{z}_i)\|^2\right)$

    \State $\mathcal{L}_\pi(\theta)\leftarrow
      \dfrac{1}{B}\sum_{i=1}^{B}
      \dfrac{-U_i +\mathcal{L}_{\mathrm{reg},i}
             +\lambda_i\, U^c_i}
            {1+\lambda_i}$

    \State $\theta\leftarrow\theta-\eta_\theta\,\nabla_\theta\,\mathcal{L}_\pi(\theta)$

    \algrule[.2pt]
    \Statex \hspace{22pt} \textit{\small\color{commentgray}Discretized Lagrange Multiplier Update}

    \If{$|\mathcal{B}_c|>0$}
      \For{$(z^{(j)}_{0, 2},\,G^{c,(j)}_0)\in\mathcal{B}_c$}
        \State $\nu^{(j)}\leftarrow
               g \left(z^{(j)}_{0, 2}+G^{c,(j)}_0\right)-\varepsilon$
        \Comment{constraint violation; $\mathbb{E}[g(z_{0, 2}+G^c_0)]\leq 0$ is the constraint}
      \EndFor
      \State $\mathcal{L}_\lambda\leftarrow
             -\dfrac{1}{|\mathcal{B}_c|}\sum_j\lambda_{k(z^{(j)}_{0, 2})}\,\nu^{(j)}$
      \State $\{\lambda_k\}\leftarrow\{\lambda_k\}-\eta_\lambda\,\nabla_{\{\lambda_k\}}\mathcal{L}_\lambda$;\quad
             $\lambda_k\leftarrow\operatorname{clip}(\lambda_k,\lambda_{\min},\lambda_{\max})\;\forall\,k$
    \EndIf

    \algrule[.2pt]
    \Statex \hspace{22pt} \textit{\small\color{commentgray}Polyak Target Updates}

    \State $\bar{\psi} \leftarrow (1 - \tau_{\mathrm{ema}})\bar{\psi} + \tau_{\mathrm{ema}}\psi$
    \State $\bar{\phi} \leftarrow (1 - \tau_{\mathrm{ema}})\bar{\phi} + \tau_{\mathrm{ema}}\phi$
    
  \EndIf
\EndFor

\Statex

\end{algorithmic}
\end{algorithm}

\subsection{Hyperparameters}
\label{app:hyperparameters}

The experiments use a shared set of core hyperparameters across all tasks, with specific adjustments made to the discount factor, $n$-step return, and critic learning rates depending on the task group (\emph{Safety Velocity} or \emph{Safety Navigation}). 
These values are summarized in \cref{tab:hyperparameters}. 
Similar to the OmniSafe benchmark, \emph{Safety Navigation} and Humanoid (from \emph{Safety Velocity}) are trained for $3$ million steps, while the remaining \emph{Safety Velocity} tasks use $1$ million steps. Also, Swimmer (from \emph{Safety Velocity}) uses a discount factor of $0.995$, while other tasks from \emph{Safety Velocity} use $0.99$. For both our method and the OmniSafe baselines, we performed a hyperparameter search over the discount factor ${0.99, 0.995, 0.999}$ and critic learning rates ${3\times10^{-4}, 1\times10^{-4}, 3\times10^{-5}}$, and reported results with the best hyperparameters for each baseline, for each task group. Despite this search, we found the default hyperparameters in OmniSafe to generally yield the best results. The notable exception was the cost normalization setting for off-policy methods; we found that enabling it (\texttt{cost\_normalize=True}) improved the results for the Push and Button tasks with both Car and Point agents. Consequently, the baseline results for these specific tasks are reported with cost normalization enabled, while all other baseline configurations remain at their defaults.

Finally, our implementation supports $n$-step returns, for which we observed that larger values (e.g., $n=10$) improved performance on \emph{Safety Navigation} tasks.
OmniSafe does not have $n$-step returns implemented, so their baselines do not use it.
This may be contributing to the baseline performances observed in \emph{Safety Navigation}, though, for reference and comparison, one can use UCP-NA as an RL baseline \emph{with} $n$-step returns (see the discussion in \cref{app:tables}).

\begin{table}[htb!]
\centering
\caption{Algorithm Hyperparameters.}
\label{tab:hyperparameters}
\begin{tabular}{lcc}
\toprule
\textbf{Hyperparameter} & \textbf{\emph{Safety Navigation}} & \textbf{\emph{Safety Velocity}} \\
\midrule
Policy Learning Rate & \multicolumn{2}{c}{$3 \times 10^{-4}$} \\
Batch Size & \multicolumn{2}{c}{$256$} \\
Replay Buffer Size & \multicolumn{2}{c}{$10^6$} \\
Target Smoothing Coef. Actor ($\tau_p$) & \multicolumn{2}{c}{$0.05$} \\
Target Smoothing Coef. Critic ($\tau_q, \tau_c$) & \multicolumn{2}{c}{$0.005$} \\
Number of Quantiles ($N$) & \multicolumn{2}{c}{$100$} \\
Network Hidden Sizes & \multicolumn{2}{c}{$[256, 256]$} \\
Constraint Violation Tolerance ($\epsilon$) & \multicolumn{2}{c}{$1 \times 10^{-6}$} \\
\midrule
Discount Factor$^*$ ($\gamma$) & $0.999$& $0.99$\\
$N$-Step Return ($n$) & $10$& $1$\\
Critic Learning Rate & $3 \times 10^{-5}$& $1 \times 10^{-4}$\\
Training steps$^*$ & $3 \times 10^{6}$ & $1 \times 10^{6}$ \\
\bottomrule
\end{tabular}
\end{table}

It is worth noting that some prior works (e.g., \citet{wu2024off, liu2022constrained, mccarthy2025optimistic, li2026off}) modify the simulation time step in Safety Gymnasium agents or disable random layout generation in the environments. In contrast, we keep all Safety Gymnasium settings unchanged. Since such modifications can substantially affect the achievable return and cost, our results are not directly comparable to those works.

\subsection{Compute Resources}
\label{app:compute}
All experiments were conducted on the Digital Research Alliance of Canada servers. OmniSafe experiments were performed on CPU nodes with 4 CPU cores and 4 GB RAM; each run took up to 2 days, particularly for Safety Navigation tasks with 3M steps. Experiments for UCP and its variants were run on GPU nodes, each using 1/8 of an A100 GPU (5 GB VRAM), along with 2 CPU cores and 15 GB RAM. Depending on the number of training steps, each run took approximately 12--36 hours.

\section{Component Ablations}
\label{app:ablations}

Despite our efforts to ensure a fair comparison between UCP and baseline algorithms, by using the same hyperparameters where possible and performing comparable tuning across methods, certain differences (e.g., the use of a distributional value function or $n$-step returns) can still significantly influence the results. To better isolate the contribution of each proposed component, we therefore conduct an ablation study. In \cref{sec:main_text_ablation}, we introduced the different variants of our method; here, we present their performance across all environments. We observe that several patterns discussed in \cref{sec:main_text_ablation} consistently reappear in other environments.

Before comparing algorithms, it is helpful to highlight the differences between \emph{Safety Velocity} and \emph{Safety Navigation} tasks. In \emph{Safety Velocity} tasks, returns across different budget levels are generally similar. In some cases (e.g., Ant, Hopper, and Walker2d), higher budgets even lead to lower returns. By visualizing the learned policies, we observe that agents tend to quickly expend their budget by operating at unsafe velocities early on, and then switch to safer behavior. This is reflected by costs closely matching the budget levels in most tasks (except Humanoid). We hypothesize that the combination of higher initial velocities and the subsequent transition to safe behavior introduces instabilities that can reduce overall return.

\begin{figure}[htb!]
  \centering
  \includegraphics[width=\linewidth]{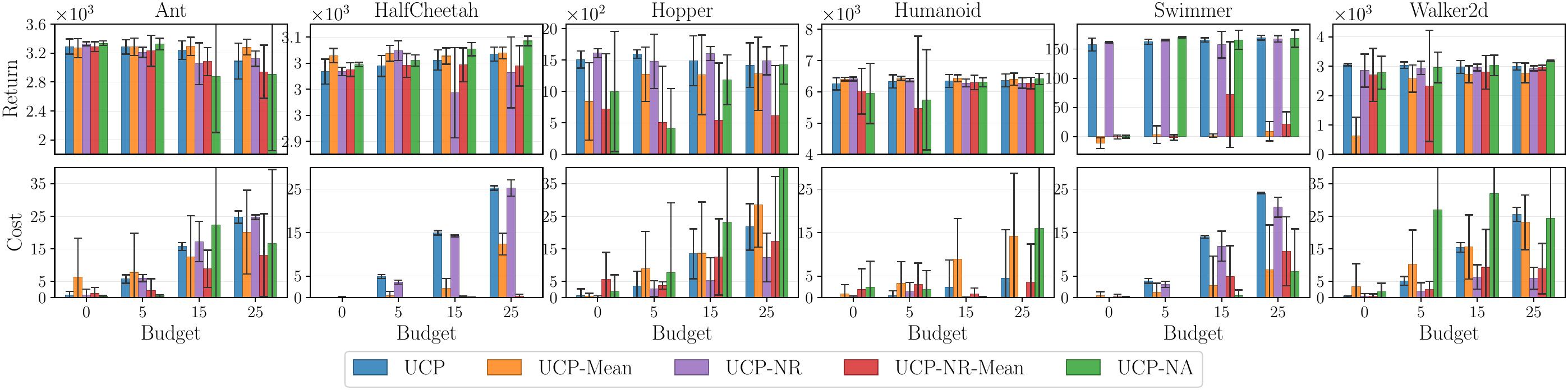}
  \caption{Ablation results on the \emph{Safety Velocity} tasks for UCP, UCP-Mean, UCP-NR, UCP-NR-Mean, and UCP-NA}
  \label{fig:barplots_final_ablation_velocity}
\end{figure}

In contrast, \emph{Safety Navigation} tasks exhibit a different pattern. Here, we observe a strong correlation between budget and return (except in Circle tasks). Agents incur cost while exploring the environment and interacting with costly regions or objects; higher budgets make the agent less sensitive to these penalties, allowing more direct paths to the goal and resulting in higher returns.

Comparing UCP and UCP-Mean with UCP-NR and UCP-NR-Mean, respectively, we find that randomizing the initial stock enables effective generalization across tasks and budget levels without retraining. There are, however, some nuanced differences. In \emph{Safety Navigation} tasks, UCP matches or exceeds the return of UCP-NR across all environments and budget levels, although UCP-NR tends to be more conservative in terms of cost. Importantly, UCP still satisfies the budget constraint at the 95\% confidence level, which makes it a preferable approach overall.

Examining UCP-Mean and UCP-NR-Mean across environments and budget levels, we observe that UCP-NR-Mean achieves average costs closer to the allocated budget, whereas UCP-Mean exceeds the budget in several cases. This suggests that stock randomization in UCP-Mean can negatively affect its performance in this setting.

Comparing all four main variants, UCP achieves the best balance between return and cost. For the remaining variants, we observe similar performance between UCP-NA and UCP-NR-Mean across environments and budget levels, indicating that state augmentation does not harm performance and can even provide improvements in some cases. In contrast, UCP-DS generally exhibits higher costs or lower returns compared to UCP across environments and budget levels, empirically supporting our discussion in \cref{sec:methodology}.

Finally, the results across tasks and budget levels highlight the advantages of the UCMDP framework. Stock randomization enables a more complete understanding of the reward–cost relationship and allows practitioners to select budgets accordingly. It also enables learning policies for multiple budget levels without retraining, removes the need for manually designing modified reward signals (since each signal is learned with its own critic), and allows flexible application of different utility functions to the cost signals.

\begin{figure}[htb!]
  \centering
  \includegraphics[width=\linewidth]{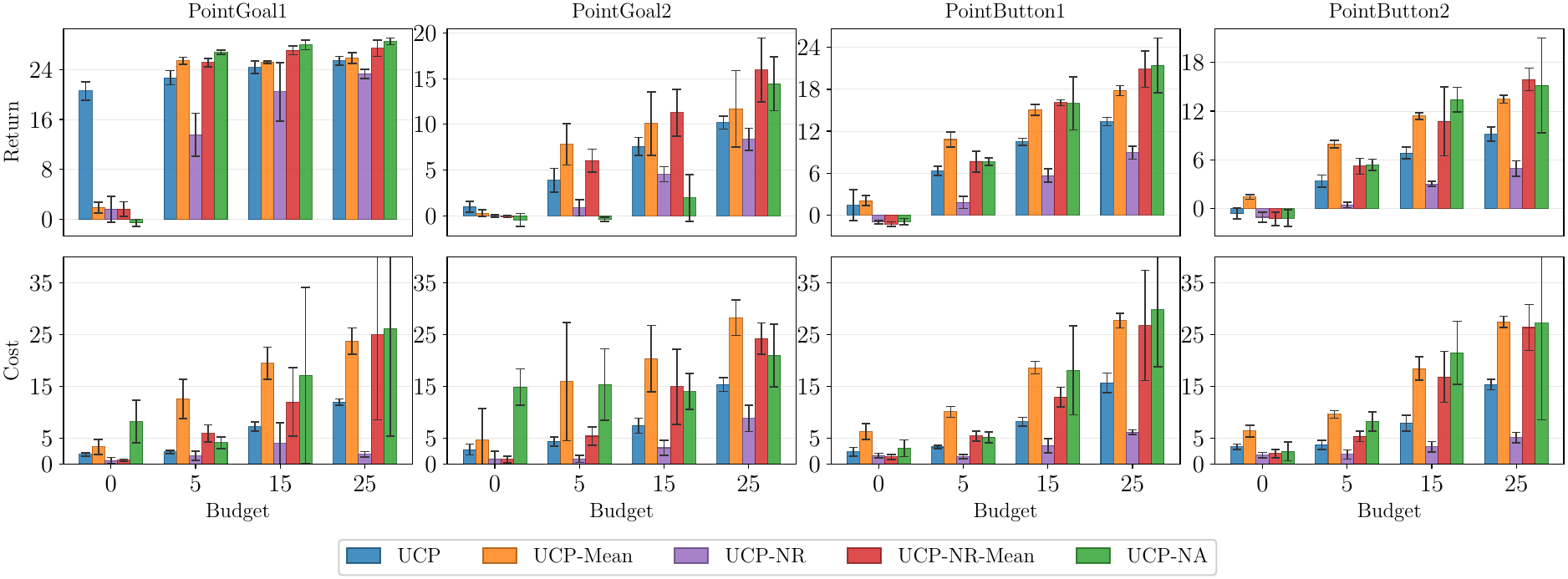}
  \caption{Ablation results on the \emph{Safety Navigation} tasks for UCP, UCP-Mean, UCP-NR, UCP-NR-Mean, and UCP-NA.}
  \label{fig:barplots_final_ablation_navigation_point_a}
\end{figure}

\begin{figure}[htb!]
  \centering
  \includegraphics[width=\linewidth]{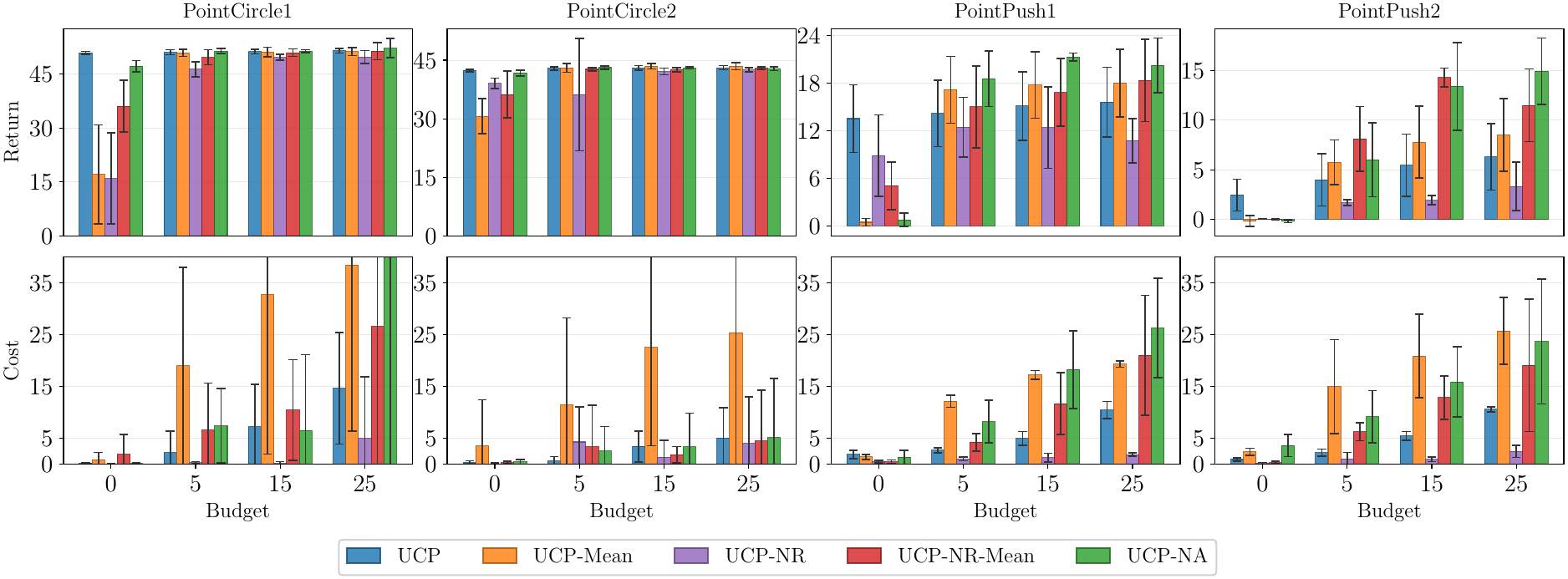}
  \caption{Ablation results on the \emph{Safety Navigation} tasks for UCP, UCP-Mean, UCP-NR, UCP-NR-Mean, and UCP-NA.}
  \label{fig:barplots_final_ablation_navigation_point_b}
\end{figure}

\begin{figure}[htb!]
  \centering
  \includegraphics[width=\linewidth]{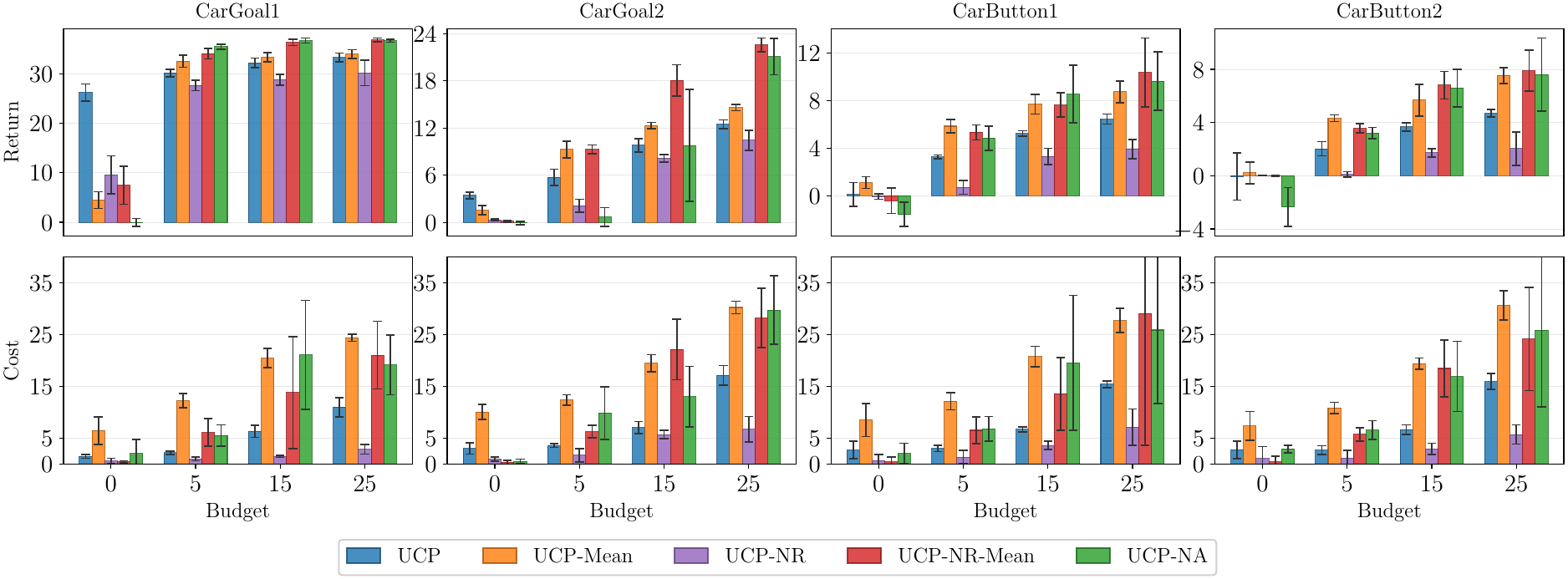}
  \caption{Ablation results on the \emph{Safety Navigation} tasks for UCP, UCP-Mean, UCP-NR, UCP-NR-Mean, and UCP-NA.}
  \label{fig:barplots_final_ablation_navigation_car_a}
\end{figure}

\begin{figure}[htb!]
  \centering
  \includegraphics[width=\linewidth]{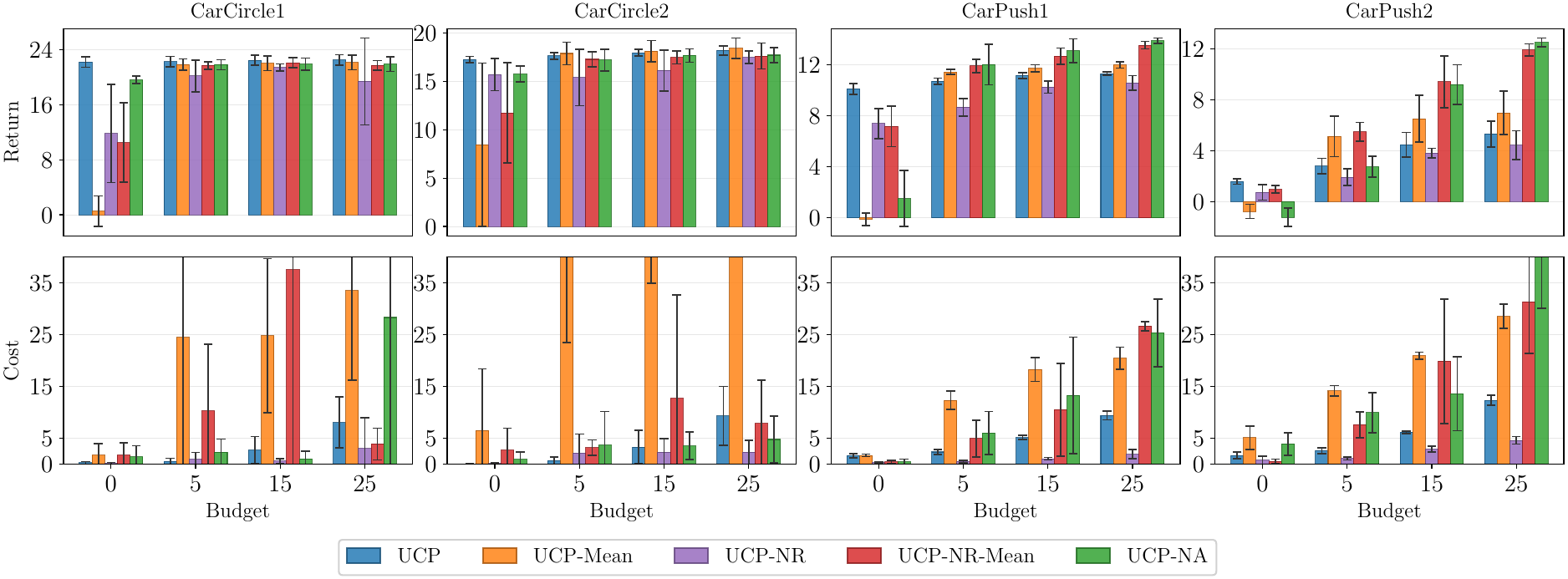}
  \caption{Ablation results on the \emph{Safety Navigation} tasks for UCP, UCP-Mean, UCP-NR, UCP-NR-Mean, and UCP-NA.}
  \label{fig:barplots_final_ablation_navigation_car_b}
\end{figure}

\clearpage

\section{Sensitivity to the Lagrange Multiplier Function Approximation}
\label{app:sensitivity_to_K}

As discussed in \cref{sec:methodology,app:implementation}, we use function approximation to maintain the estimates of the Lagrange multipliers $\lambda_z$, via a $K$-bin discretization over $[z_{\mathrm{min}},z_{\mathrm{max}}]$.
In this section, we conduct a brief study of the impact of the discretization granularity $K$ on the performance of UCP, and we see that, as expected, return improves with the resolution up to the $K = 31$ that we have used in the main text.
Note that a denser grid does not require additional environment interaction or network evaluation, but only additional scalar Lagrange multipliers, so its computational cost is negligible.

We evaluated UCP on the CarGoal1 task with different values of $K$ ($4$, $13$, $31$), and the results are given in \cref{tab:ablation_K}.
\begin{table}[htbp!]
\centering
\caption{UCP on CarGoal1 with different discretization levels $K$, evaluated at different initial stocks $z_0$. Each cell shows return (top) and cost (bottom, red), with 95\% Student's $t$ intervals across five seeds.}
\label{tab:ablation_K}
\begin{adjustbox}{max width=1.2\linewidth,center}
\begin{tabular}{lcccc}
\toprule
Discretization & $z_0=0$ & $z_0=-5$ & $z_0=-15$ & $z_0=-25$ \\
\midrule
\multirow{2}{*}{$K=4$} & $1.5\pm1.7$ & $0.1\pm0.8$ & $23.9\pm2.4$ & $28.9\pm1.5$ \\
 & \textcolor{red}{$0.4\pm0.2$} & \textcolor{red}{$0.5\pm0.3$} & \textcolor{red}{$4.2\pm1.1$} & \textcolor{red}{$7.1\pm1.3$} \\
\midrule
\multirow{2}{*}{$K=13$} & $17.9\pm7.8$ & $25.4\pm6.5$ & $30.3\pm2.2$ & $31.7\pm2.4$ \\ 
 & \textcolor{red}{$1.2\pm0.2$} & \textcolor{red}{$1.4\pm0.4$} & \textcolor{red}{$6.1\pm0.6$} & \textcolor{red}{$11.4\pm0.9$} \\
\midrule
\multirow{2}{*}{$K=31$} & $26.2\pm1.7$ & $30.1\pm0.8$ & $32.2\pm1.0$ & $33.3\pm0.9$ \\
 & \textcolor{red}{$1.5\pm0.4$} & \textcolor{red}{$2.2\pm0.3$} & \textcolor{red}{$6.4\pm1.2$} & \textcolor{red}{$11.0\pm1.9$} \\
\bottomrule
\end{tabular}
\end{adjustbox}
\end{table}
Over the training range $[-30, 0]$, $K=4$, $13$, and $31$ give stock spacings of $10$, $2.5$, and $1$.
Return improves at every budget, for example from $0.1$ to $25.4$ to $30.1$ at $z_0=-5$, and from $28.9$ to $31.7$ to $33.3$ at $z_0=-25$.
A denser grid exposes the agent to more initial stocks and assigns distinct multiplier values to nearby budgets, which may improve both interpolation to evaluation stocks and overall learning.
For every $z_0<0$, all three settings satisfy the benchmark's risk-neutral constraint, so the improvement does not sacrifice mean-budget satisfaction.

\section{Sensitivity to the Constraint Violation Tolerance \texorpdfstring{$\epsilon$}{eps}}
\label{app:sensitivity_to_eps}

As discussed in \cref{sec:methodology}, the constraint $\mathbb{E}[ (z + G_0)_+ ] \leq 0$ may not be strictly feasible~\citep[Section 5.2.3, p.~226]{boyd2004convex}.
This and the fact that we use imperfect return distribution estimators for the Lagrangian multiplier updates can make the optimization challenging.
Therefore, we use a small constraint violation tolerance $\epsilon$ which mitigates both issues and has the practical benefit that $\lambda_z$ decreases when the constraint is approximately satisfied, unlike the case where there is no tolerance.

We can see the practical impact of the tolerance in \cref{tab:ablation_epsilon}, where we report the performance on CarGoal1 of UCP with different values of $\epsilon$.
\begin{table}[htbp!]
\centering
\caption{UCP on CarGoal1 with different constraint violation tolerances $\epsilon$, evaluated at different initial stocks $z_0$. Each cell shows return (top) and cost (bottom, red), with 95\% Student's $t$ intervals across five seeds.}
\label{tab:ablation_epsilon}
\begin{adjustbox}{max width=1.2\linewidth,center}
\begin{tabular}{lcccc}
\toprule
Tolerance & $z_0=0$ & $z_0=-5$ & $z_0=-15$ & $z_0=-25$ \\
\midrule
\multirow{2}{*}{$\epsilon=0$} & $13.3\pm4.4$ & $26.8\pm1.3$ & $31.7\pm0.8$ & $32.8\pm0.5$ \\
 & \textcolor{red}{$0.8\pm0.3$} & \textcolor{red}{$1.1\pm0.5$} & \textcolor{red}{$4.7\pm0.7$} & \textcolor{red}{$10.0\pm1.2$} \\
\midrule
\multirow{2}{*}{$\epsilon=10^{-6}$} & $26.2\pm1.7$ & $30.1\pm0.8$ & $32.2\pm1.0$ & $33.3\pm0.9$ \\
 & \textcolor{red}{$1.5\pm0.4$} & \textcolor{red}{$2.2\pm0.3$} & \textcolor{red}{$6.4\pm1.2$} & \textcolor{red}{$11.0\pm1.9$} \\
\midrule
\multirow{2}{*}{$\epsilon=10^{-4}$} & $28.4\pm1.5$ & $31.2\pm1.3$ & $33.0\pm0.8$ & $33.9\pm0.6$ \\
 & \textcolor{red}{$1.8\pm0.4$} & \textcolor{red}{$2.5\pm0.3$} & \textcolor{red}{$7.1\pm0.9$} & \textcolor{red}{$11.5\pm1.4$} \\
\bottomrule
\end{tabular}
\end{adjustbox}
\end{table}
Setting $\epsilon = 0$ prevents the multiplier from decreasing, because $g \geq 0$, so accumulated early increases cannot be undone.
This results in an overly conservative learned policy. At budget $0$, $\epsilon = 0$ obtains return $13.3$ at cost $0.8$, compared with return $28.4$ at cost $1.8$ for $\epsilon = 10^{-4}$. The return and mean cost increase as $\epsilon$ and $-z_0$ increase (though $\epsilon$ remains relatively small), and at budgets $5$, $15$, and $25$, all tested settings remain feasible in mean.
A positive tolerance lets $\lambda_z$ decrease whenever the estimated excess is below $\epsilon$, so the multiplier can recover from early over-penalization.
Accordingly, $\epsilon=10^{-6}$ gives an intermediate result, while $\epsilon=10^{-4}$ is the least conservative and obtains the highest return at every budget.

\section{Empirical Distributions}
\label{app:ablations_distributions}

In this section, we present the empirical distributions of return and cost for all environments at a budget level of 25. The tables accompanying each figure report the probability that trajectories from each policy exceed the allocated budget, as well as the magnitude of the excess cost.

Across all \emph{Safety Navigation} tasks (except Circle), we observe that the probability of violating the constraint under risk-neutral variants such as UCP-NA, UCP-NR-Mean, and UCP-Mean exceeds 30\%. We also find that UCP-NR achieves the lowest probability of budget violations, albeit at a noticeable drop in terms of return. In contrast, UCP achieves a more favorable balance between return and constraint satisfaction.

Another notable observation is that, despite the lower violation probability of UCP-NR, the expected excess cost under UCP is often smaller. This suggests that the increased diversity in the training data in UCP leads to more controlled violations when they occur.

\begin{figure}[htb!]
  \centering
  \includegraphics[width=\linewidth]{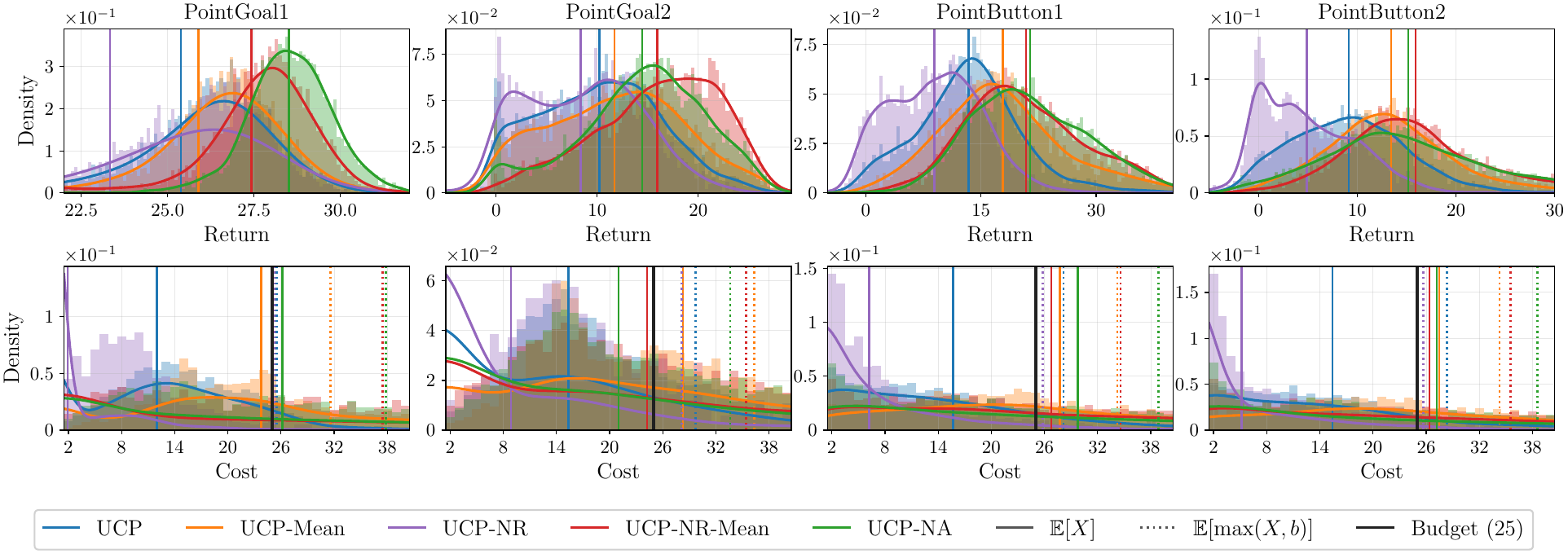}
  \caption{Empirical distributions of episodic return and cost at budget 25 on the \emph{Safety Navigation} task with the \emph{Point} agent.}
  \label{fig:figures_final_histogram_navigation_point_a_budget_25}
\end{figure}

\begin{table}[htbp]
\centering
\caption{Evaluation statistics on \emph{Safety Navigation} (\emph{Point}) tasks with cost budget of $25$. 
  The third column corresponds to the estimated conditional expectation of the undiscounted cost given that it (strictly) exceeds the budget, and fourth to the estimated probability of the undiscounted cost being above the budget.}
\label{tab:hist_set_nav_point_a_c25}
  \begin{subtable}{0.49\linewidth}
    \centering
    \resizebox{\linewidth}{!}{
    \begin{tabular}{lccccc}
    \toprule
    Method & Return & Cost & \makecell{Conditional \\ Expected Cost} & \makecell{Freq. of Cost \\ above Budget} \\
    \midrule
    UCP & $25.4\pm0.1$ & \textcolor{red}{$12.0\pm0.3$} & $31.7$ & $0.077$ \\
    UCP-Mean & $25.9\pm0.1$ & \textcolor{red}{$23.7\pm0.5$} & $43.7$ & $0.352$ \\
    UCP-NR & $23.3\pm0.2$ & \textcolor{red}{$1.9\pm0.2$} & $53.0$ & $0.009$ \\
    UCP-NR-Mean & $27.4\pm0.1$ & \textcolor{red}{$24.9\pm0.9$} & $58.4$ & $0.373$ \\
    UCP-NA & $28.5\pm0.0$ & \textcolor{red}{$26.2\pm0.9$} & $58.3$ & $0.385$ \\
    \bottomrule
    \end{tabular}
    }
    \caption{PointGoal1}
    \label{tab:hist_pointgoal1_c25_sub}
  \end{subtable}
\hfill
  \begin{subtable}{0.49\linewidth}
    \centering
    \resizebox{\linewidth}{!}{
    \begin{tabular}{lccccc}
    \toprule
    Method & Return & Cost & \makecell{Conditional \\ Expected Cost} & \makecell{Freq. of Cost \\ above Budget} \\
    \midrule
    UCP & $10.2\pm0.2$ & \textcolor{red}{$15.4\pm0.6$} & $49.5$ & $0.192$ \\
    UCP-Mean & $11.7\pm0.2$ & \textcolor{red}{$28.3\pm0.7$} & $50.7$ & $0.441$ \\
    UCP-NR & $8.4\pm0.2$ & \textcolor{red}{$8.9\pm0.7$} & $57.4$ & $0.097$ \\
    UCP-NR-Mean & $15.9\pm0.2$ & \textcolor{red}{$24.2\pm0.8$} & $53.3$ & $0.367$ \\
    UCP-NA & $14.4\pm0.2$ & \textcolor{red}{$21.0\pm0.9$} & $53.2$ & $0.305$ \\
    \bottomrule
    \end{tabular}
    }
    \caption{PointGoal2}
    \label{tab:hist_pointgoal2_c25_sub}
  \end{subtable}
\par\medskip
  \begin{subtable}{0.49\linewidth}
    \centering
    \resizebox{\linewidth}{!}{
    \begin{tabular}{lccccc}
    \toprule
    Method & Return & Cost & \makecell{Conditional \\ Expected Cost} & \makecell{Freq. of Cost \\ above Budget} \\
    \midrule
    UCP & $13.4\pm0.2$ & \textcolor{red}{$15.7\pm0.5$} & $42.5$ & $0.181$ \\
    UCP-Mean & $17.8\pm0.2$ & \textcolor{red}{$27.7\pm0.6$} & $45.8$ & $0.444$ \\
    UCP-NR & $8.9\pm0.2$ & \textcolor{red}{$6.2\pm0.4$} & $41.5$ & $0.048$ \\
    UCP-NR-Mean & $20.9\pm0.2$ & \textcolor{red}{$26.8\pm0.6$} & $46.8$ & $0.441$ \\
    UCP-NA & $21.4\pm0.2$ & \textcolor{red}{$29.8\pm0.9$} & $57.3$ & $0.429$ \\
    \bottomrule
    \end{tabular}
    }
    \caption{PointButton1}
    \label{tab:hist_pointbutton1_c25_sub}
  \end{subtable}
\hfill
  \begin{subtable}{0.49\linewidth}
    \centering
    \resizebox{\linewidth}{!}{
    \begin{tabular}{lccccc}
    \toprule
    Method & Return & Cost & \makecell{Conditional \\ Expected Cost} & \makecell{Freq. of Cost \\ above Budget} \\
    \midrule
    UCP & $9.2\pm0.2$ & \textcolor{red}{$15.4\pm0.5$} & $44.1$ & $0.174$ \\
    UCP-Mean & $13.5\pm0.2$ & \textcolor{red}{$27.5\pm0.6$} & $46.0$ & $0.441$ \\
    UCP-NR & $4.9\pm0.1$ & \textcolor{red}{$5.1\pm0.3$} & $41.8$ & $0.041$ \\
    UCP-NR-Mean & $15.9\pm0.2$ & \textcolor{red}{$26.4\pm0.7$} & $50.2$ & $0.417$ \\
    UCP-NA & $15.2\pm0.2$ & \textcolor{red}{$27.2\pm1.0$} & $61.6$ & $0.371$ \\
    \bottomrule
    \end{tabular}
    }
    \caption{PointButton2}
    \label{tab:hist_pointbutton2_c25_sub}
  \end{subtable}
\end{table}

\begin{figure}[htb!]
  \centering
  \includegraphics[width=\linewidth]{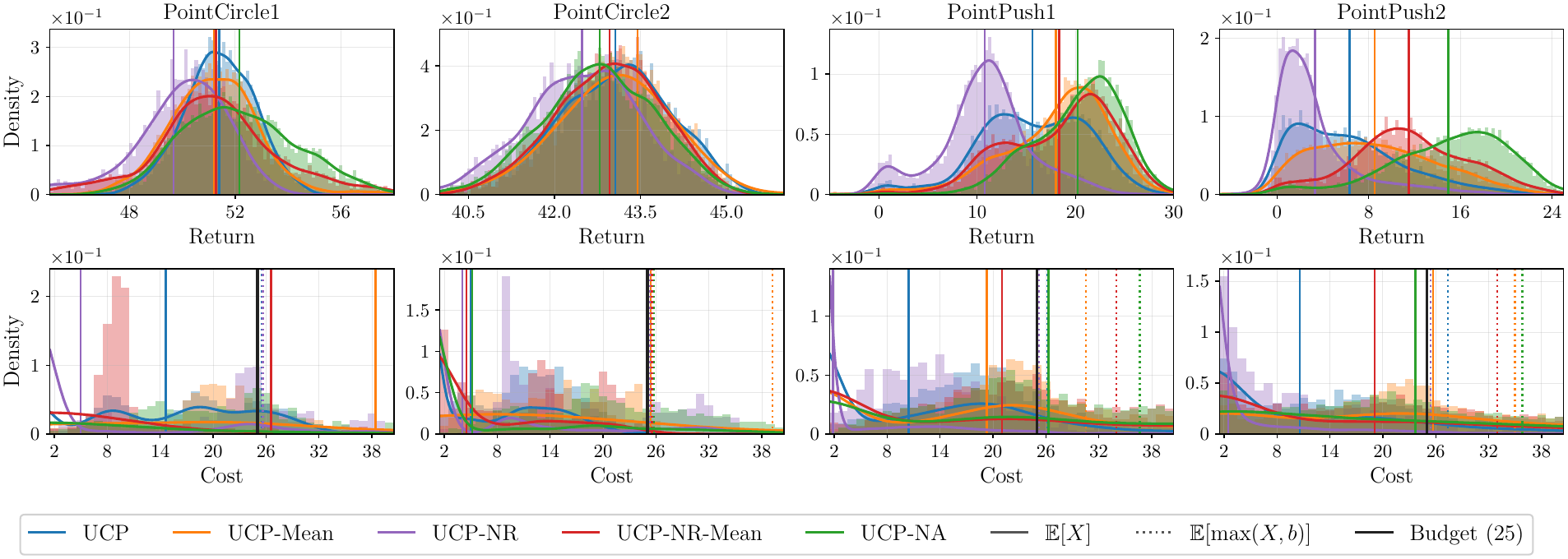}
  \caption{Empirical distributions of episodic return and cost at budget 25 on the \emph{Safety Navigation} task with the \emph{Point} agent.}
  \label{fig:figures_final_histogram_navigation_point_b_budget_25}
\end{figure}


\begin{table}[htbp]
\centering
\caption{Evaluation statistics on \emph{Safety Navigation} (\emph{Point}) tasks with cost budget of $25$. 
  The third column corresponds to the estimated conditional expectation of the undiscounted cost given that it (strictly) exceeds the budget, and fourth to the estimated probability of the undiscounted cost being above the budget.}
\label{tab:hist_set_nav_point_b_c25}
  \begin{subtable}{0.49\linewidth}
    \centering
    \resizebox{\linewidth}{!}{
    \begin{tabular}{lccccc}
    \toprule
    Method & Return & Cost & \makecell{Conditional \\ Expected Cost} & \makecell{Freq. of Cost \\ above Budget} \\
    \midrule
    UCP & $51.4\pm0.0$ & \textcolor{red}{$14.6\pm0.3$} & $28.4$ & $0.164$ \\
    UCP-Mean & $51.2\pm0.1$ & \textcolor{red}{$38.4\pm1.4$} & $77.0$ & $0.415$ \\
    UCP-NR & $49.7\pm0.1$ & \textcolor{red}{$5.0\pm0.3$} & $33.1$ & $0.078$ \\
    UCP-NR-Mean & $51.3\pm0.1$ & \textcolor{red}{$26.5\pm1.4$} & $91.3$ & $0.273$ \\
    UCP-NA & $52.2\pm0.1$ & \textcolor{red}{$54.9\pm1.6$} & $99.9$ & $0.537$ \\
    \bottomrule
    \end{tabular}
    }
    \caption{PointCircle1}
    \label{tab:hist_pointcircle1_c25_sub}
  \end{subtable}
\hfill
  \begin{subtable}{0.49\linewidth}
    \centering
    \resizebox{\linewidth}{!}{
    \begin{tabular}{lccccc}
    \toprule
    Method & Return & Cost & \makecell{Conditional \\ Expected Cost} & \makecell{Freq. of Cost \\ above Budget} \\
    \midrule
    UCP & $43.1\pm0.0$ & \textcolor{red}{$5.0\pm0.2$} & $26.3$ & $0.002$ \\
    UCP-Mean & $43.4\pm0.1$ & \textcolor{red}{$25.4\pm1.8$} & $126.1$ & $0.141$ \\
    UCP-NR & $42.5\pm0.0$ & \textcolor{red}{$4.1\pm0.2$} & $30.2$ & $0.067$ \\
    UCP-NR-Mean & $43.0\pm0.0$ & \textcolor{red}{$4.5\pm0.5$} & $300.1$ & $0.002$ \\
    UCP-NA & $42.8\pm0.0$ & \textcolor{red}{$5.1\pm0.3$} & $35.1$ & $0.079$ \\
    \bottomrule
    \end{tabular}
    }
    \caption{PointCircle2}
    \label{tab:hist_pointcircle2_c25_sub}
  \end{subtable}
\par\medskip
  \begin{subtable}{0.49\linewidth}
    \centering
    \resizebox{\linewidth}{!}{
    \begin{tabular}{lccccc}
    \toprule
    Method & Return & Cost & \makecell{Conditional \\ Expected Cost} & \makecell{Freq. of Cost \\ above Budget} \\
    \midrule
    UCP & $15.6\pm0.2$ & \textcolor{red}{$10.4\pm0.4$} & $35.6$ & $0.108$ \\
    UCP-Mean & $18.0\pm0.2$ & \textcolor{red}{$19.3\pm0.6$} & $44.0$ & $0.293$ \\
    UCP-NR & $10.8\pm0.1$ & \textcolor{red}{$1.9\pm0.2$} & $38.8$ & $0.014$ \\
    UCP-NR-Mean & $18.4\pm0.2$ & \textcolor{red}{$21.0\pm0.7$} & $52.2$ & $0.330$ \\
    UCP-NA & $20.2\pm0.1$ & \textcolor{red}{$26.3\pm0.8$} & $53.2$ & $0.413$ \\
    \bottomrule
    \end{tabular}
    }
    \caption{PointPush1}
    \label{tab:hist_pointpush1_c25_sub}
  \end{subtable}
\hfill
  \begin{subtable}{0.49\linewidth}
    \centering
    \resizebox{\linewidth}{!}{
    \begin{tabular}{lccccc}
    \toprule
    Method & Return & Cost & \makecell{Conditional \\ Expected Cost} & \makecell{Freq. of Cost \\ above Budget} \\
    \midrule
    UCP & $6.3\pm0.1$ & \textcolor{red}{$10.6\pm0.5$} & $43.5$ & $0.129$ \\
    UCP-Mean & $8.5\pm0.2$ & \textcolor{red}{$25.7\pm0.9$} & $51.8$ & $0.372$ \\
    UCP-NR & $3.3\pm0.1$ & \textcolor{red}{$2.5\pm0.2$} & $42.8$ & $0.023$ \\
    UCP-NR-Mean & $11.5\pm0.1$ & \textcolor{red}{$19.1\pm0.8$} & $53.2$ & $0.283$ \\
    UCP-NA & $14.9\pm0.1$ & \textcolor{red}{$23.7\pm1.3$} & $58.5$ & $0.323$ \\
    \bottomrule
    \end{tabular}
    }
    \caption{PointPush2}
    \label{tab:hist_pointpush2_c25_sub}
  \end{subtable}
\end{table}

\begin{figure}[htb!]
  \centering
  \includegraphics[width=\linewidth]{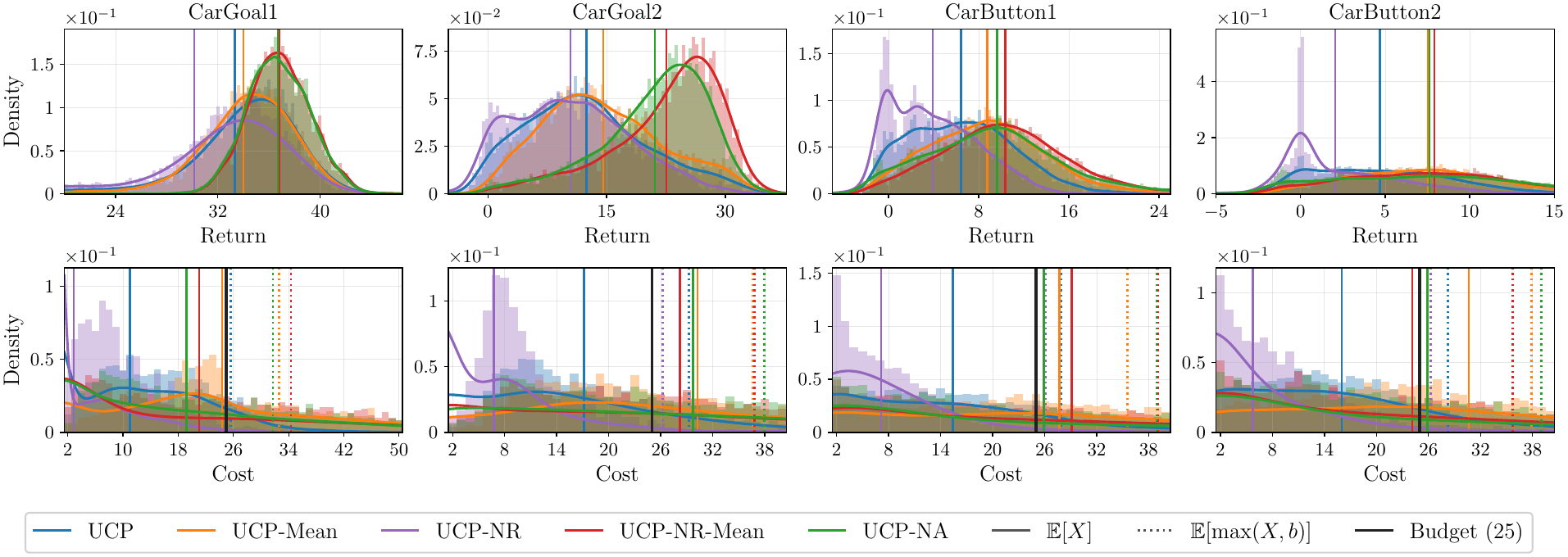}
  \caption{Empirical distributions of episodic return and cost at budget 25 on the \emph{Safety Navigation} task with the \emph{Car} agent.}
  \label{fig:figures_final_histogram_navigation_car_a_budget_25}
\end{figure}

\begin{table}[htbp]
\centering
\caption{Evaluation statistics on \emph{Safety Navigation} (\emph{Car}) tasks with cost budget of $25$. 
  The third column corresponds to the estimated conditional expectation of the undiscounted cost given that it (strictly) exceeds the budget, and fourth to the estimated probability of the undiscounted cost being above the budget.}
\label{tab:hist_set_nav_car_a_c25}
  \begin{subtable}{0.49\linewidth}
    \centering
    \resizebox{\linewidth}{!}{
    \begin{tabular}{lccccc}
    \toprule
    Method & Return & Cost & \makecell{Conditional \\ Expected Cost} & \makecell{Freq. of Cost \\ above Budget} \\
    \midrule
    UCP & $33.3\pm0.2$ & \textcolor{red}{$11.0\pm0.3$} & $31.4$ & $0.094$ \\
    UCP-Mean & $34.0\pm0.1$ & \textcolor{red}{$24.4\pm0.6$} & $46.0$ & $0.364$ \\
    UCP-NR & $30.1\pm0.2$ & \textcolor{red}{$2.9\pm0.2$} & $33.3$ & $0.007$ \\
    UCP-NR-Mean & $36.8\pm0.1$ & \textcolor{red}{$21.1\pm0.8$} & $53.6$ & $0.326$ \\
    UCP-NA & $36.7\pm0.1$ & \textcolor{red}{$19.2\pm0.6$} & $48.1$ & $0.292$ \\
    \bottomrule
    \end{tabular}
    }
    \caption{CarGoal1}
    \label{tab:hist_cargoal1_c25_sub}
  \end{subtable}
\hfill
  \begin{subtable}{0.49\linewidth}
    \centering
    \resizebox{\linewidth}{!}{
    \begin{tabular}{lccccc}
    \toprule
    Method & Return & Cost & \makecell{Conditional \\ Expected Cost} & \makecell{Freq. of Cost \\ above Budget} \\
    \midrule
    UCP & $12.5\pm0.2$ & \textcolor{red}{$17.2\pm0.6$} & $45.8$ & $0.205$ \\
    UCP-Mean & $14.6\pm0.2$ & \textcolor{red}{$30.2\pm0.8$} & $50.0$ & $0.465$ \\
    UCP-NR & $10.4\pm0.2$ & \textcolor{red}{$6.7\pm0.4$} & $56.2$ & $0.039$ \\
    UCP-NR-Mean & $22.5\pm0.2$ & \textcolor{red}{$28.2\pm0.7$} & $50.8$ & $0.459$ \\
    UCP-NA & $21.1\pm0.2$ & \textcolor{red}{$29.7\pm0.8$} & $53.2$ & $0.459$ \\
    \bottomrule
    \end{tabular}
    }
    \caption{CarGoal2}
    \label{tab:hist_cargoal2_c25_sub}
  \end{subtable}
\par\medskip
  \begin{subtable}{0.49\linewidth}
    \centering
    \resizebox{\linewidth}{!}{
    \begin{tabular}{lccccc}
    \toprule
    Method & Return & Cost & \makecell{Conditional \\ Expected Cost} & \makecell{Freq. of Cost \\ above Budget} \\
    \midrule
    UCP & $6.4\pm0.1$ & \textcolor{red}{$15.4\pm0.5$} & $41.7$ & $0.174$ \\
    UCP-Mean & $8.7\pm0.1$ & \textcolor{red}{$27.7\pm0.8$} & $48.2$ & $0.456$ \\
    UCP-NR & $3.9\pm0.1$ & \textcolor{red}{$7.2\pm0.7$} & $71.9$ & $0.024$ \\
    UCP-NR-Mean & $10.4\pm0.2$ & \textcolor{red}{$29.1\pm0.9$} & $58.7$ & $0.417$ \\
    UCP-NA & $9.6\pm0.2$ & \textcolor{red}{$25.9\pm1.2$} & $67.1$ & $0.329$ \\
    \bottomrule
    \end{tabular}
    }
    \caption{CarButton1}
    \label{tab:hist_carbutton1_c25_sub}
  \end{subtable}
\hfill
  \begin{subtable}{0.49\linewidth}
    \centering
    \resizebox{\linewidth}{!}{
    \begin{tabular}{lccccc}
    \toprule
    Method & Return & Cost & \makecell{Conditional \\ Expected Cost} & \makecell{Freq. of Cost \\ above Budget} \\
    \midrule
    UCP & $4.7\pm0.1$ & \textcolor{red}{$16.0\pm0.6$} & $43.2$ & $0.178$ \\
    UCP-Mean & $7.5\pm0.1$ & \textcolor{red}{$30.6\pm1.0$} & $52.2$ & $0.473$ \\
    UCP-NR & $2.0\pm0.1$ & \textcolor{red}{$5.7\pm0.6$} & $59.4$ & $0.036$ \\
    UCP-NR-Mean & $7.9\pm0.1$ & \textcolor{red}{$24.1\pm0.9$} & $55.3$ & $0.353$ \\
    UCP-NA & $7.6\pm0.2$ & \textcolor{red}{$25.8\pm1.2$} & $67.9$ & $0.327$ \\
    \bottomrule
    \end{tabular}
    }
    \caption{CarButton2}
    \label{tab:hist_carbutton2_c25_sub}
  \end{subtable}
\end{table}

\begin{figure}[htb!]
  \centering
  \includegraphics[width=\linewidth]{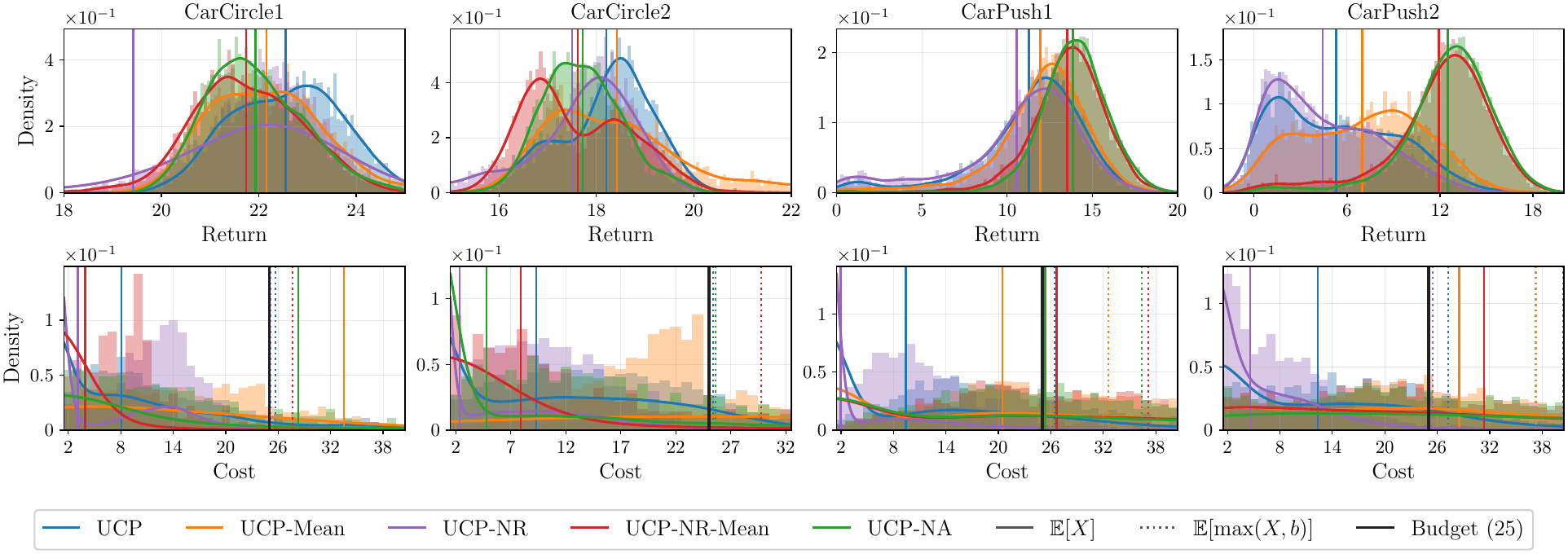}
  \caption{Empirical distributions of episodic return and cost at budget 25 on the \emph{Safety Navigation} task with the \emph{Car} agent.}
  \label{fig:figures_final_histogram_navigation_car_b_budget_25}
\end{figure}

\begin{table}[htbp]
\centering
\caption{Evaluation statistics on \emph{Safety Navigation} (\emph{Car}) tasks with cost budget of $25$. 
  The third column corresponds to the estimated conditional expectation of the undiscounted cost given that it (strictly) exceeds the budget, and fourth to the estimated probability of the undiscounted cost being above the budget.}
\label{tab:hist_set_nav_car_b_c25}
  \begin{subtable}{0.49\linewidth}
    \centering
    \resizebox{\linewidth}{!}{
    \begin{tabular}{lccccc}
    \toprule
    Method & Return & Cost & \makecell{Conditional \\ Expected Cost} & \makecell{Freq. of Cost \\ above Budget} \\
    \midrule
    UCP & $22.6\pm0.0$ & \textcolor{red}{$8.1\pm0.3$} & $34.2$ & $0.075$ \\
    UCP-Mean & $22.2\pm0.0$ & \textcolor{red}{$33.5\pm1.5$} & $92.9$ & $0.303$ \\
    UCP-NR & $19.4\pm0.2$ & \textcolor{red}{$3.1\pm0.2$} & $45.8$ & $0.004$ \\
    UCP-NR-Mean & $21.7\pm0.0$ & \textcolor{red}{$3.9\pm0.6$} & $111.6$ & $0.031$ \\
    UCP-NA & $21.9\pm0.0$ & \textcolor{red}{$28.3\pm1.1$} & $75.5$ & $0.352$ \\
    \bottomrule
    \end{tabular}
    }
    \caption{CarCircle1}
    \label{tab:hist_carcircle1_c25_sub}
  \end{subtable}
\hfill
  \begin{subtable}{0.49\linewidth}
    \centering
    \resizebox{\linewidth}{!}{
    \begin{tabular}{lccccc}
    \toprule
    Method & Return & Cost & \makecell{Conditional \\ Expected Cost} & \makecell{Freq. of Cost \\ above Budget} \\
    \midrule
    UCP & $18.2\pm0.0$ & \textcolor{red}{$9.3\pm0.3$} & $30.1$ & $0.073$ \\
    UCP-Mean & $18.4\pm0.0$ & \textcolor{red}{$83.9\pm3.1$} & $128.5$ & $0.600$ \\
    UCP-NR & $17.5\pm0.0$ & \textcolor{red}{$2.3\pm0.1$} & $35.0$ & $0.003$ \\
    UCP-NR-Mean & $17.6\pm0.0$ & \textcolor{red}{$7.9\pm1.0$} & $88.9$ & $0.075$ \\
    UCP-NA & $17.7\pm0.0$ & \textcolor{red}{$4.8\pm0.3$} & $35.1$ & $0.063$ \\
    \bottomrule
    \end{tabular}
    }
    \caption{CarCircle2}
    \label{tab:hist_carcircle2_c25_sub}
  \end{subtable}
\par\medskip
  \begin{subtable}{0.49\linewidth}
    \centering
    \resizebox{\linewidth}{!}{
    \begin{tabular}{lccccc}
    \toprule
    Method & Return & Cost & \makecell{Conditional \\ Expected Cost} & \makecell{Freq. of Cost \\ above Budget} \\
    \midrule
    UCP & $11.3\pm0.1$ & \textcolor{red}{$9.4\pm0.4$} & $35.9$ & $0.130$ \\
    UCP-Mean & $12.0\pm0.1$ & \textcolor{red}{$20.5\pm0.7$} & $46.9$ & $0.347$ \\
    UCP-NR & $10.6\pm0.1$ & \textcolor{red}{$2.0\pm0.2$} & $42.5$ & $0.011$ \\
    UCP-NR-Mean & $13.5\pm0.1$ & \textcolor{red}{$26.6\pm0.8$} & $53.0$ & $0.435$ \\
    UCP-NA & $13.9\pm0.1$ & \textcolor{red}{$25.3\pm0.8$} & $53.0$ & $0.408$ \\
    \bottomrule
    \end{tabular}
    }
    \caption{CarPush1}
    \label{tab:hist_carpush1_c25_sub}
  \end{subtable}
\hfill
  \begin{subtable}{0.49\linewidth}
    \centering
    \resizebox{\linewidth}{!}{
    \begin{tabular}{lccccc}
    \toprule
    Method & Return & Cost & \makecell{Conditional \\ Expected Cost} & \makecell{Freq. of Cost \\ above Budget} \\
    \midrule
    UCP & $5.3\pm0.1$ & \textcolor{red}{$12.4\pm0.5$} & $40.7$ & $0.144$ \\
    UCP-Mean & $7.0\pm0.1$ & \textcolor{red}{$28.5\pm1.1$} & $54.0$ & $0.424$ \\
    UCP-NR & $4.4\pm0.1$ & \textcolor{red}{$4.6\pm0.3$} & $51.8$ & $0.017$ \\
    UCP-NR-Mean & $11.9\pm0.1$ & \textcolor{red}{$31.4\pm1.2$} & $60.4$ & $0.437$ \\
    UCP-NA & $12.5\pm0.1$ & \textcolor{red}{$43.9\pm2.0$} & $76.5$ & $0.518$ \\
    \bottomrule
    \end{tabular}
    }
    \caption{CarPush2}
    \label{tab:hist_carpush2_c25_sub}
  \end{subtable}
\end{table}

\begin{figure}[htb!]
  \centering
  \includegraphics[width=\linewidth]{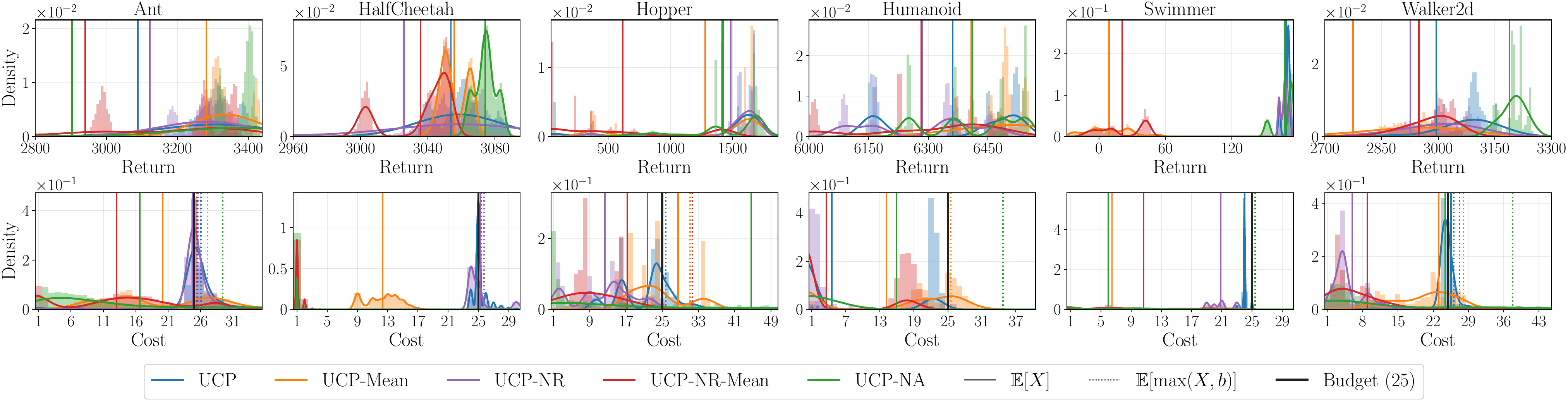}
  \caption{Empirical distributions of episodic return and cost at budget 25 on the \emph{Safety Velocity} tasks.}
  \label{fig:figures_final_histogram_velocity_budget_25}
\end{figure}


\begin{table}[htbp]
\centering
\caption{Evaluation statistics on \emph{Safety Velocity} tasks with cost budget of $25$. 
  The third column corresponds to the estimated conditional expectation of the undiscounted cost given that it (strictly) exceeds the budget, and fourth to the estimated probability of the undiscounted cost being above the budget.}
\label{tab:hist_set_vel_c25}
  \begin{subtable}{0.49\linewidth}
    \centering
    \resizebox{\linewidth}{!}{
    \begin{tabular}{lccccc}
    \toprule
    Method & Return & Cost & \makecell{Conditional \\ Expected Cost} & \makecell{Freq. of Cost \\ above Budget} \\
    \midrule
    UCP & $3088.7\pm24.8$ & \textcolor{red}{$24.8\pm0.2$} & $27.6$ & $0.395$ \\
    UCP-Mean & $3281.2\pm13.1$ & \textcolor{red}{$20.2\pm0.3$} & $30.6$ & $0.371$ \\
    UCP-NR & $3122.8\pm18.5$ & \textcolor{red}{$24.7\pm0.1$} & $27.0$ & $0.305$ \\
    UCP-NR-Mean & $2940.7\pm18.1$ & \textcolor{red}{$13.0\pm0.3$} & $28.4$ & $0.136$ \\
    UCP-NA & $2903.1\pm28.5$ & \textcolor{red}{$16.6\pm0.5$} & $42.3$ & $0.257$ \\
    \bottomrule
    \end{tabular}
    }
    \caption{Ant}
    \label{tab:hist_ant_c25_sub}
  \end{subtable}
\hfill
  \begin{subtable}{0.49\linewidth}
    \centering
    \resizebox{\linewidth}{!}{
    \begin{tabular}{lccccc}
    \toprule
    Method & Return & Cost & \makecell{Conditional \\ Expected Cost} & \makecell{Freq. of Cost \\ above Budget} \\
    \midrule
    UCP & $3053.8\pm3.0$ & \textcolor{red}{$25.3\pm0.0$} & $27.0$ & $0.192$ \\
    UCP-Mean & $3055.7\pm0.2$ & \textcolor{red}{$12.3\pm0.1$} & -- & $0.000$ \\
    UCP-NR & $3025.8\pm5.5$ & \textcolor{red}{$25.3\pm0.1$} & $29.5$ & $0.170$ \\
    UCP-NR-Mean & $3035.8\pm0.6$ & \textcolor{red}{$0.3\pm0.0$} & -- & $0.000$ \\
    UCP-NA & $3074.2\pm0.2$ & \textcolor{red}{$0.0\pm0.0$} & -- & $0.000$ \\
    \bottomrule
    \end{tabular}
    }
    \caption{HalfCheetah}
    \label{tab:hist_halfcheetah_c25_sub}
  \end{subtable}
\par\medskip
  \begin{subtable}{0.49\linewidth}
    \centering
    \resizebox{\linewidth}{!}{
    \begin{tabular}{lccccc}
    \toprule
    Method & Return & Cost & \makecell{Conditional \\ Expected Cost} & \makecell{Freq. of Cost \\ above Budget} \\
    \midrule
    UCP & $1419.1\pm14.9$ & \textcolor{red}{$21.7\pm0.2$} & $28.1$ & $0.258$ \\
    UCP-Mean & $1281.7\pm14.6$ & \textcolor{red}{$28.5\pm0.4$} & $44.3$ & $0.320$ \\
    UCP-NR & $1486.5\pm10.2$ & \textcolor{red}{$12.4\pm0.2$} & $29.2$ & $0.020$ \\
    UCP-NR-Mean & $616.9\pm16.4$ & \textcolor{red}{$17.3\pm0.8$} & $89.1$ & $0.104$ \\
    UCP-NA & $1423.9\pm9.4$ & \textcolor{red}{$44.6\pm1.6$} & $96.5$ & $0.439$ \\
    \bottomrule
    \end{tabular}
    }
    \caption{Hopper}
    \label{tab:hist_hopper_c25_sub}
  \end{subtable}
\hfill
  \begin{subtable}{0.49\linewidth}
    \centering
    \resizebox{\linewidth}{!}{
    \begin{tabular}{lccccc}
    \toprule
    Method & Return & Cost & \makecell{Conditional \\ Expected Cost} & \makecell{Freq. of Cost \\ above Budget} \\
    \midrule
    UCP & $6361.7\pm4.6$ & \textcolor{red}{$4.5\pm0.2$} & -- & $0.000$ \\
    UCP-Mean & $6407.5\pm20.3$ & \textcolor{red}{$14.2\pm0.3$} & $27.3$ & $0.217$ \\
    UCP-NR & $6284.1\pm4.8$ & \textcolor{red}{$0.0\pm0.0$} & $26.0$ & $0.000$ \\
    UCP-NR-Mean & $6281.7\pm9.6$ & \textcolor{red}{$3.5\pm0.2$} & -- & $0.000$ \\
    UCP-NA & $6410.7\pm3.5$ & \textcolor{red}{$16.0\pm0.9$} & $63.9$ & $0.250$ \\
    \bottomrule
    \end{tabular}
    }
    \caption{Humanoid}
    \label{tab:hist_humanoid_c25_sub}
  \end{subtable}
\par\medskip
  \begin{subtable}{0.49\linewidth}
    \centering
    \resizebox{\linewidth}{!}{
    \begin{tabular}{lccccc}
    \toprule
    Method & Return & Cost & \makecell{Conditional \\ Expected Cost} & \makecell{Freq. of Cost \\ above Budget} \\
    \midrule
    UCP & $169.4\pm0.1$ & \textcolor{red}{$24.1\pm0.0$} & -- & $0.000$ \\
    UCP-Mean & $9.2\pm0.6$ & \textcolor{red}{$6.5\pm0.2$} & $28.1$ & $0.002$ \\
    UCP-NR & $167.5\pm0.1$ & \textcolor{red}{$20.8\pm0.0$} & -- & $0.000$ \\
    UCP-NR-Mean & $21.1\pm0.6$ & \textcolor{red}{$10.7\pm0.2$} & $43.5$ & $0.022$ \\
    UCP-NA & $167.9\pm0.3$ & \textcolor{red}{$6.0\pm0.3$} & $74.4$ & $0.007$ \\
    \bottomrule
    \end{tabular}
    }
    \caption{Swimmer}
    \label{tab:hist_swimmer_c25_sub}
  \end{subtable}
\hfill
  \begin{subtable}{0.49\linewidth}
    \centering
    \resizebox{\linewidth}{!}{
    \begin{tabular}{lccccc}
    \toprule
    Method & Return & Cost & \makecell{Conditional \\ Expected Cost} & \makecell{Freq. of Cost \\ above Budget} \\
    \midrule
    UCP & $2994.9\pm10.2$ & \textcolor{red}{$25.6\pm0.1$} & $29.8$ & $0.229$ \\
    UCP-Mean & $2775.0\pm15.7$ & \textcolor{red}{$23.1\pm0.4$} & $36.2$ & $0.267$ \\
    UCP-NR & $2926.2\pm11.1$ & \textcolor{red}{$6.0\pm0.1$} & $27.5$ & $0.012$ \\
    UCP-NR-Mean & $2948.7\pm6.8$ & \textcolor{red}{$8.9\pm0.5$} & $81.3$ & $0.040$ \\
    UCP-NA & $3189.3\pm3.4$ & \textcolor{red}{$24.4\pm1.0$} & $62.9$ & $0.338$ \\
    \bottomrule
    \end{tabular}
    }
    \caption{Walker2d}
    \label{tab:hist_walker2d_c25_sub}
  \end{subtable}
\end{table}

\clearpage

\section{Tables}
\label{app:tables}
In this section, we present our results in table format for ease of reference and comparison.

In the Button and Push tasks with both Point and Car agents, the OmniSafe baselines perform significantly worse than our results. Comparing with the results reported in \citet[Table 5]{ji2023safety}, we find that our outcomes, even after extensive hyperparameter tuning, are consistent with those reported. However, our UCP-NA baseline, which uses a risk-neutral objective with a distributional critic, shows that the performance of existing methods can be improved using standard techniques such as distributional critics and $n$-step returns. We report UCP-NA specifically to highlight how much of the performance gain comes from our proposed components versus these standard design choices.


\begin{table}[htbp!]
\centering
\caption{Comparison of UCP with OmniSafe baselines on Velocity and Navigation environments.}
\label{tab:comparison_combined}
\begin{adjustbox}{max width=1.2\linewidth,center}
\begin{tabular}{lcccccccc}
\toprule
Environment & \textbf{UCP} & \textbf{PPOLag} & \textbf{PPOSaute} & \textbf{PPO-ET} & \textbf{SACLag} & \textbf{SACPID} & \textbf{TD3Lag} & \textbf{TD3PID} \\
\midrule
\multicolumn{9}{l}{\textit{Velocity Environments}} \\
\midrule
\multirow{2}{*}{Ant} & $3289.6\pm104.5$ & $3166.8\pm88.6$ & $2711.2\pm84.6$ & $1899.5\pm111.6$ & $2215.5\pm717.3$ & $2756.7\pm182.1$ & $2335.4\pm915.5$ & $1677.5\pm1686.3$ \\
 & \textcolor{red}{$0.9\pm1.1$} & \textcolor{red}{$27.9\pm3.7$} & \textcolor{red}{$17.5\pm0.4$} & \textcolor{red}{$13.3\pm2.0$} & \textcolor{red}{$11.8\pm12.7$} & \textcolor{red}{$3.7\pm2.3$} & \textcolor{red}{$19.8\pm11.5$} & \textcolor{red}{$19.8\pm8.8$} \\
\midrule
\multirow{2}{*}{HalfCheetah} & $3044.9\pm15.6$ & $3024.1\pm31.5$ & $2792.7\pm31.7$ & $2091.8\pm21.5$ & $2967.9\pm886.0$ & $2447.0\pm258.0$ & $1977.4\pm353.3$ & $1664.6\pm294.2$ \\
 & \textcolor{red}{$14.9\pm0.5$} & \textcolor{red}{$17.3\pm1.3$} & \textcolor{red}{$17.1\pm0.2$} & \textcolor{red}{$9.9\pm0.4$} & \textcolor{red}{$146.0\pm89.6$} & \textcolor{red}{$3.0\pm7.5$} & \textcolor{red}{$19.6\pm13.4$} & \textcolor{red}{$6.3\pm4.1$} \\
\midrule
\multirow{2}{*}{Hopper} & $1593.3\pm65.9$ & $1394.4\pm607.2$ & $1656.4\pm29.0$ & $1442.7\pm50.2$ & $970.1\pm148.0$ & $994.1\pm121.9$ & $1093.8\pm106.5$ & $1107.0\pm97.5$ \\
 & \textcolor{red}{$3.6\pm4.5$} & \textcolor{red}{$23.5\pm15.4$} & \textcolor{red}{$17.9\pm0.4$} & \textcolor{red}{$5.3\pm1.1$} & \textcolor{red}{$63.5\pm72.7$} & \textcolor{red}{$63.5\pm62.0$} & \textcolor{red}{$53.1\pm35.7$} & \textcolor{red}{$39.3\pm43.4$} \\
\midrule
\multirow{2}{*}{Humanoid} & $6361.7\pm205.5$ & $6478.3\pm100.4$ & $5986.0\pm42.5$ & $5547.7\pm86.3$ & $6137.2\pm172.4$ & $6068.6\pm250.1$ & $5925.7\pm202.2$ & $6021.6\pm154.9$ \\
 & \textcolor{red}{$4.5\pm11.1$} & \textcolor{red}{$25.7\pm5.2$} & \textcolor{red}{$20.4\pm0.8$} & \textcolor{red}{$6.0\pm0.4$} & \textcolor{red}{$8.9\pm15.1$} & \textcolor{red}{$0.6\pm1.2$} & \textcolor{red}{$20.4\pm13.2$} & \textcolor{red}{$24.2\pm2.6$} \\
\midrule
\multirow{2}{*}{Swimmer} & $169.4\pm4.1$ & $82.2\pm28.3$ & $111.6\pm72.4$ & $22.0\pm1.3$ & $4.2\pm4.4$ & $5.1\pm3.1$ & $31.6\pm16.4$ & $43.8\pm10.3$ \\
 & \textcolor{red}{$24.1\pm0.2$} & \textcolor{red}{$27.6\pm2.9$} & \textcolor{red}{$12.5\pm6.7$} & \textcolor{red}{$14.8\pm1.3$} & \textcolor{red}{$16.3\pm14.0$} & \textcolor{red}{$4.3\pm6.6$} & \textcolor{red}{$20.7\pm4.4$} & \textcolor{red}{$30.2\pm13.6$} \\
\midrule
\multirow{2}{*}{Walker2d} & $3056.5\pm45.2$ & $3015.3\pm181.8$ & $2064.9\pm1381.5$ & $1349.7\pm1116.6$ & $960.2\pm271.1$ & $2146.1\pm553.5$ & $2538.3\pm362.4$ & $2290.1\pm752.9$ \\
 & \textcolor{red}{$0.3\pm0.3$} & \textcolor{red}{$23.7\pm16.8$} & \textcolor{red}{$18.5\pm0.8$} & \textcolor{red}{$14.6\pm6.0$} & \textcolor{red}{$23.3\pm13.7$} & \textcolor{red}{$23.0\pm22.6$} & \textcolor{red}{$32.3\pm21.2$} & \textcolor{red}{$19.9\pm13.2$} \\
\midrule
\multicolumn{9}{l}{\textit{Navigation Environments}} \\
\midrule
\multirow{2}{*}{PointGoal1} & $25.4\pm0.7$ & $16.0\pm1.6$ & $6.2\pm3.3$ & $15.7\pm0.5$ & $16.5\pm5.1$ & $20.2\pm4.1$ & $20.9\pm2.3$ & $13.9\pm7.9$ \\
 & \textcolor{red}{$12.0\pm0.6$} & \textcolor{red}{$21.1\pm2.4$} & \textcolor{red}{$21.1\pm3.3$} & \textcolor{red}{$18.4\pm0.9$} & \textcolor{red}{$24.2\pm10.8$} & \textcolor{red}{$29.6\pm8.7$} & \textcolor{red}{$29.8\pm6.5$} & \textcolor{red}{$20.8\pm3.7$} \\
\midrule
\multirow{2}{*}{PointGoal2} & $10.2\pm0.7$ & $1.3\pm0.5$ & $0.6\pm1.0$ & $6.9\pm0.2$ & $1.1\pm2.3$ & $0.7\pm1.7$ & $2.5\pm1.4$ & $-0.4\pm0.3$ \\
 & \textcolor{red}{$15.4\pm1.3$} & \textcolor{red}{$24.4\pm4.4$} & \textcolor{red}{$28.1\pm3.6$} & \textcolor{red}{$25.2\pm0.2$} & \textcolor{red}{$40.8\pm14.3$} & \textcolor{red}{$37.3\pm23.3$} & \textcolor{red}{$38.5\pm16.4$} & \textcolor{red}{$51.9\pm28.8$} \\
\midrule
\multirow{2}{*}{PointButton1} & $13.4\pm0.6$ & $4.5\pm1.7$ & $0.8\pm1.3$ & $10.6\pm0.3$ & $10.3\pm2.7$ & $9.1\pm4.2$ & $6.2\pm2.3$ & $7.0\pm3.5$ \\
 & \textcolor{red}{$15.7\pm1.9$} & \textcolor{red}{$23.2\pm3.5$} & \textcolor{red}{$26.4\pm10.8$} & \textcolor{red}{$24.4\pm0.1$} & \textcolor{red}{$23.9\pm3.1$} & \textcolor{red}{$27.1\pm6.9$} & \textcolor{red}{$26.6\pm1.9$} & \textcolor{red}{$23.5\pm4.3$} \\
\midrule
\multirow{2}{*}{PointButton2} & $9.2\pm0.9$ & $2.0\pm1.7$ & $-0.2\pm0.3$ & $7.8\pm0.6$ & $7.9\pm3.1$ & $4.8\pm0.7$ & $8.0\pm1.6$ & $2.7\pm3.0$ \\
 & \textcolor{red}{$15.4\pm1.0$} & \textcolor{red}{$24.7\pm3.4$} & \textcolor{red}{$29.4\pm12.6$} & \textcolor{red}{$25.4\pm0.1$} & \textcolor{red}{$31.8\pm10.5$} & \textcolor{red}{$23.2\pm4.6$} & \textcolor{red}{$29.4\pm8.9$} & \textcolor{red}{$21.9\pm11.6$} \\
\midrule
\multirow{2}{*}{PointCircle1} & $51.4\pm0.6$ & $47.2\pm1.4$ & $42.4\pm1.8$ & $32.3\pm8.2$ & $42.5\pm1.4$ & $40.0\pm4.1$ & $40.7\pm2.9$ & $40.5\pm2.9$ \\
 & \textcolor{red}{$14.6\pm10.8$} & \textcolor{red}{$22.0\pm4.6$} & \textcolor{red}{$14.3\pm6.7$} & \textcolor{red}{$13.4\pm9.1$} & \textcolor{red}{$25.2\pm15.3$} & \textcolor{red}{$25.5\pm2.5$} & \textcolor{red}{$27.3\pm10.9$} & \textcolor{red}{$22.4\pm15.7$} \\
\midrule
\multirow{2}{*}{PointCircle2} & $43.1\pm0.5$ & $41.8\pm0.3$ & $38.6\pm1.2$ & $35.5\pm2.1$ & $42.3\pm0.3$ & $42.2\pm0.4$ & $41.8\pm1.6$ & $41.6\pm1.2$ \\
 & \textcolor{red}{$5.0\pm5.9$} & \textcolor{red}{$26.7\pm14.9$} & \textcolor{red}{$19.7\pm10.2$} & \textcolor{red}{$6.5\pm4.4$} & \textcolor{red}{$24.2\pm1.6$} & \textcolor{red}{$29.1\pm19.6$} & \textcolor{red}{$30.8\pm12.6$} & \textcolor{red}{$34.1\pm18.5$} \\
\midrule
\multirow{2}{*}{PointPush1} & $15.6\pm4.4$ & $1.3\pm1.9$ & $0.3\pm0.2$ & $1.1\pm1.2$ & $0.2\pm0.2$ & $-0.3\pm0.5$ & $-0.3\pm0.3$ & $-0.3\pm0.8$ \\
 & \textcolor{red}{$10.4\pm1.7$} & \textcolor{red}{$24.3\pm2.1$} & \textcolor{red}{$9.9\pm7.2$} & \textcolor{red}{$8.4\pm1.9$} & \textcolor{red}{$12.2\pm10.0$} & \textcolor{red}{$28.3\pm14.8$} & \textcolor{red}{$31.9\pm36.9$} & \textcolor{red}{$38.3\pm65.4$} \\
\midrule
\multirow{2}{*}{PointPush2} & $6.3\pm3.4$ & $0.8\pm0.1$ & $0.1\pm0.5$ & $0.9\pm0.1$ & $-0.0\pm0.5$ & $-0.2\pm0.3$ & $-0.1\pm0.3$ & $-0.5\pm0.6$ \\
 & \textcolor{red}{$10.6\pm0.5$} & \textcolor{red}{$26.1\pm6.1$} & \textcolor{red}{$13.5\pm3.7$} & \textcolor{red}{$12.6\pm2.2$} & \textcolor{red}{$25.7\pm13.8$} & \textcolor{red}{$27.7\pm26.3$} & \textcolor{red}{$20.8\pm18.8$} & \textcolor{red}{$20.7\pm14.5$} \\
\midrule
\multirow{2}{*}{CarGoal1} & $33.3\pm0.9$ & $14.8\pm3.8$ & $9.6\pm2.9$ & $18.7\pm1.5$ & $15.4\pm8.2$ & $21.0\pm16.4$ & $16.6\pm13.1$ & $22.3\pm9.8$ \\
 & \textcolor{red}{$11.0\pm1.9$} & \textcolor{red}{$22.6\pm2.8$} & \textcolor{red}{$22.6\pm6.0$} & \textcolor{red}{$18.4\pm0.8$} & \textcolor{red}{$31.7\pm14.0$} & \textcolor{red}{$35.6\pm3.1$} & \textcolor{red}{$27.6\pm8.6$} & \textcolor{red}{$29.9\pm14.1$} \\
\midrule
\multirow{2}{*}{CarGoal2} & $12.5\pm0.6$ & $1.5\pm0.4$ & $0.6\pm0.4$ & $6.6\pm0.2$ & $3.3\pm2.4$ & $3.2\pm2.3$ & $2.9\pm4.6$ & $2.8\pm3.7$ \\
 & \textcolor{red}{$17.2\pm1.9$} & \textcolor{red}{$28.9\pm3.2$} & \textcolor{red}{$20.4\pm6.8$} & \textcolor{red}{$25.3\pm0.0$} & \textcolor{red}{$29.2\pm11.4$} & \textcolor{red}{$28.8\pm8.7$} & \textcolor{red}{$30.4\pm20.2$} & \textcolor{red}{$31.8\pm20.6$} \\
\midrule
\multirow{2}{*}{CarButton1} & $6.4\pm0.4$ & $0.1\pm0.3$ & $-0.3\pm0.8$ & $4.7\pm0.6$ & $1.8\pm1.2$ & $1.3\pm1.1$ & $1.4\pm0.6$ & $0.7\pm0.9$ \\
 & \textcolor{red}{$15.4\pm0.6$} & \textcolor{red}{$41.3\pm12.5$} & \textcolor{red}{$31.6\pm8.5$} & \textcolor{red}{$25.9\pm0.1$} & \textcolor{red}{$28.9\pm13.2$} & \textcolor{red}{$20.0\pm4.1$} & \textcolor{red}{$31.6\pm8.0$} & \textcolor{red}{$28.2\pm5.9$} \\
\midrule
\multirow{2}{*}{CarButton2} & $4.7\pm0.3$ & $0.4\pm0.5$ & $0.0\pm0.3$ & $4.2\pm0.2$ & $1.6\pm0.6$ & $0.6\pm0.6$ & $1.4\pm0.6$ & $0.5\pm1.1$ \\
 & \textcolor{red}{$16.0\pm1.5$} & \textcolor{red}{$52.7\pm15.2$} & \textcolor{red}{$40.0\pm10.6$} & \textcolor{red}{$25.9\pm0.1$} & \textcolor{red}{$32.4\pm8.0$} & \textcolor{red}{$20.4\pm12.2$} & \textcolor{red}{$33.6\pm4.5$} & \textcolor{red}{$24.6\pm13.0$} \\
\midrule
\multirow{2}{*}{CarCircle1} & $22.6\pm0.8$ & $18.4\pm0.1$ & $13.5\pm1.5$ & $14.4\pm1.2$ & $18.8\pm2.0$ & $17.7\pm5.2$ & $18.7\pm1.5$ & $18.1\pm3.7$ \\
 & \textcolor{red}{$8.1\pm4.9$} & \textcolor{red}{$22.8\pm5.4$} & \textcolor{red}{$13.2\pm4.8$} & \textcolor{red}{$8.6\pm3.9$} & \textcolor{red}{$30.1\pm14.4$} & \textcolor{red}{$86.7\pm158.6$} & \textcolor{red}{$29.1\pm3.5$} & \textcolor{red}{$21.8\pm12.0$} \\
\midrule
\multirow{2}{*}{CarCircle2} & $18.2\pm0.5$ & $16.2\pm0.3$ & $13.0\pm0.6$ & $11.9\pm0.4$ & $15.2\pm4.1$ & $16.1\pm1.8$ & $17.5\pm0.9$ & $16.9\pm1.0$ \\
 & \textcolor{red}{$9.3\pm5.7$} & \textcolor{red}{$38.2\pm14.4$} & \textcolor{red}{$16.8\pm5.0$} & \textcolor{red}{$12.1\pm2.3$} & \textcolor{red}{$26.6\pm7.9$} & \textcolor{red}{$25.0\pm3.0$} & \textcolor{red}{$58.0\pm42.6$} & \textcolor{red}{$29.0\pm10.2$} \\
\midrule
\multirow{2}{*}{CarPush1} & $11.3\pm0.1$ & $0.4\pm0.2$ & $0.4\pm0.2$ & $0.5\pm0.1$ & $1.0\pm0.2$ & $1.0\pm0.4$ & $0.8\pm0.3$ & $0.9\pm0.1$ \\
 & \textcolor{red}{$9.4\pm0.8$} & \textcolor{red}{$24.5\pm7.1$} & \textcolor{red}{$16.1\pm6.2$} & \textcolor{red}{$6.3\pm0.7$} & \textcolor{red}{$20.0\pm17.3$} & \textcolor{red}{$16.4\pm2.4$} & \textcolor{red}{$18.2\pm16.4$} & \textcolor{red}{$22.3\pm7.4$} \\
\midrule
\multirow{2}{*}{CarPush2} & $5.3\pm1.0$ & $0.5\pm0.1$ & $0.3\pm0.1$ & $0.7\pm0.1$ & $0.6\pm0.5$ & $0.2\pm0.3$ & $0.4\pm0.4$ & $0.3\pm0.2$ \\
 & \textcolor{red}{$12.4\pm0.9$} & \textcolor{red}{$29.1\pm5.3$} & \textcolor{red}{$26.8\pm4.6$} & \textcolor{red}{$15.8\pm0.8$} & \textcolor{red}{$33.6\pm8.0$} & \textcolor{red}{$21.1\pm11.6$} & \textcolor{red}{$45.9\pm15.9$} & \textcolor{red}{$24.8\pm7.4$} \\
\bottomrule
\end{tabular}
\end{adjustbox}
\end{table}


\begin{table}[htbp!]
\centering
\caption{Results of UCP, evaluated at different initial stocks $z_0$.}
\label{tab:ctx_orig_combined}
\begin{adjustbox}{max width=1.2\linewidth,center}
\begin{tabular}{lcccc}
\toprule
Environment & $z_0=0$ & $z_0=-5$ & $z_0=-15$ & $z_0=-25$ \\
\midrule
\multicolumn{5}{l}{\textit{Velocity Environments}} \\
\midrule
\multirow{2}{*}{Ant} & $3289.6\pm104.5$ & $3285.4\pm99.3$ & $3239.3\pm127.4$ & $3088.7\pm247.1$ \\
 & \textcolor{red}{$0.9\pm1.1$} & \textcolor{red}{$5.7\pm1.3$} & \textcolor{red}{$15.7\pm1.2$} & \textcolor{red}{$24.8\pm1.9$} \\
\midrule
\multirow{2}{*}{HalfCheetah} & $3027.1\pm19.3$ & $3035.7\pm16.0$ & $3044.9\pm15.6$ & $3053.8\pm10.8$ \\
 & \textcolor{red}{$0.0\pm0.0$} & \textcolor{red}{$4.9\pm0.5$} & \textcolor{red}{$14.9\pm0.5$} & \textcolor{red}{$25.3\pm0.5$} \\
\midrule
\multirow{2}{*}{Hopper} & $1506.8\pm136.3$ & $1593.3\pm65.9$ & $1487.4\pm398.0$ & $1419.1\pm354.4$ \\
 & \textcolor{red}{$0.7\pm2.0$} & \textcolor{red}{$3.6\pm4.5$} & \textcolor{red}{$13.5\pm7.7$} & \textcolor{red}{$21.7\pm7.2$} \\
\midrule
\multirow{2}{*}{Humanoid} & $6256.0\pm198.1$ & $6334.0\pm214.3$ & $6349.6\pm208.1$ & $6361.7\pm205.5$ \\
 & \textcolor{red}{$0.0\pm0.0$} & \textcolor{red}{$0.4\pm1.1$} & \textcolor{red}{$2.5\pm6.2$} & \textcolor{red}{$4.5\pm11.1$} \\
\midrule
\multirow{2}{*}{Swimmer} & $157.5\pm11.1$ & $162.6\pm4.1$ & $165.4\pm3.6$ & $169.4\pm4.1$ \\
 & \textcolor{red}{$0.0\pm0.0$} & \textcolor{red}{$3.9\pm0.5$} & \textcolor{red}{$14.1\pm0.3$} & \textcolor{red}{$24.1\pm0.2$} \\
\midrule
\multirow{2}{*}{Walker2d} & $3056.5\pm45.2$ & $3036.6\pm107.1$ & $2974.9\pm217.3$ & $2994.9\pm120.9$ \\
 & \textcolor{red}{$0.3\pm0.3$} & \textcolor{red}{$5.2\pm1.3$} & \textcolor{red}{$15.4\pm1.4$} & \textcolor{red}{$25.6\pm2.2$} \\
\midrule
\multicolumn{5}{l}{\textit{Navigation Environments}} \\
\midrule
\multirow{2}{*}{PointGoal1} & $20.6\pm1.5$ & $22.7\pm1.1$ & $24.4\pm1.0$ & $25.4\pm0.7$ \\
 & \textcolor{red}{$1.9\pm0.4$} & \textcolor{red}{$2.4\pm0.3$} & \textcolor{red}{$7.3\pm0.9$} & \textcolor{red}{$12.0\pm0.6$} \\
\midrule
\multirow{2}{*}{PointGoal2} & $1.0\pm0.6$ & $3.9\pm1.3$ & $7.6\pm1.0$ & $10.2\pm0.7$ \\
 & \textcolor{red}{$2.8\pm1.0$} & \textcolor{red}{$4.4\pm0.9$} & \textcolor{red}{$7.5\pm1.4$} & \textcolor{red}{$15.4\pm1.3$} \\
\midrule
\multirow{2}{*}{PointButton1} & $1.4\pm2.2$ & $6.3\pm0.6$ & $10.5\pm0.5$ & $13.4\pm0.6$ \\
 & \textcolor{red}{$2.4\pm0.8$} & \textcolor{red}{$3.3\pm0.3$} & \textcolor{red}{$8.2\pm0.9$} & \textcolor{red}{$15.7\pm1.9$} \\
\midrule
\multirow{2}{*}{PointButton2} & $-0.6\pm0.7$ & $3.4\pm0.7$ & $6.8\pm0.7$ & $9.2\pm0.9$ \\
 & \textcolor{red}{$3.4\pm0.5$} & \textcolor{red}{$3.7\pm0.9$} & \textcolor{red}{$7.9\pm1.5$} & \textcolor{red}{$15.4\pm1.0$} \\
\midrule
\multirow{2}{*}{PointCircle1} & $50.8\pm0.5$ & $51.0\pm0.7$ & $51.2\pm0.6$ & $51.4\pm0.6$ \\
 & \textcolor{red}{$0.2\pm0.1$} & \textcolor{red}{$2.3\pm4.2$} & \textcolor{red}{$7.2\pm8.2$} & \textcolor{red}{$14.6\pm10.8$} \\
\midrule
\multirow{2}{*}{PointCircle2} & $42.4\pm0.3$ & $42.9\pm0.4$ & $43.0\pm0.6$ & $43.1\pm0.5$ \\
 & \textcolor{red}{$0.3\pm0.4$} & \textcolor{red}{$0.6\pm0.8$} & \textcolor{red}{$3.4\pm3.0$} & \textcolor{red}{$5.0\pm5.9$} \\
\midrule
\multirow{2}{*}{PointPush1} & $13.5\pm4.3$ & $14.2\pm4.2$ & $15.1\pm4.3$ & $15.6\pm4.4$ \\
 & \textcolor{red}{$1.9\pm0.8$} & \textcolor{red}{$2.7\pm0.5$} & \textcolor{red}{$5.0\pm1.3$} & \textcolor{red}{$10.4\pm1.7$} \\
\midrule
\multirow{2}{*}{PointPush2} & $2.5\pm1.6$ & $4.0\pm2.6$ & $5.5\pm3.1$ & $6.3\pm3.4$ \\
 & \textcolor{red}{$0.9\pm0.3$} & \textcolor{red}{$2.2\pm0.7$} & \textcolor{red}{$5.5\pm0.8$} & \textcolor{red}{$10.6\pm0.5$} \\
\midrule
\multirow{2}{*}{CarGoal1} & $26.2\pm1.7$ & $30.1\pm0.8$ & $32.2\pm1.0$ & $33.3\pm0.9$ \\
 & \textcolor{red}{$1.5\pm0.4$} & \textcolor{red}{$2.2\pm0.3$} & \textcolor{red}{$6.4\pm1.2$} & \textcolor{red}{$11.0\pm1.9$} \\
\midrule
\multirow{2}{*}{CarGoal2} & $3.5\pm0.4$ & $5.7\pm1.0$ & $9.8\pm0.9$ & $12.5\pm0.6$ \\
 & \textcolor{red}{$3.1\pm1.1$} & \textcolor{red}{$3.6\pm0.4$} & \textcolor{red}{$7.0\pm1.2$} & \textcolor{red}{$17.2\pm1.9$} \\
\midrule
\multirow{2}{*}{CarButton1} & $0.1\pm1.0$ & $3.3\pm0.2$ & $5.3\pm0.2$ & $6.4\pm0.4$ \\
 & \textcolor{red}{$2.7\pm1.7$} & \textcolor{red}{$3.1\pm0.6$} & \textcolor{red}{$6.7\pm0.5$} & \textcolor{red}{$15.4\pm0.6$} \\
\midrule
\multirow{2}{*}{CarButton2} & $-0.1\pm1.8$ & $2.0\pm0.5$ & $3.7\pm0.3$ & $4.7\pm0.3$ \\
 & \textcolor{red}{$2.8\pm1.7$} & \textcolor{red}{$2.7\pm0.8$} & \textcolor{red}{$6.7\pm1.0$} & \textcolor{red}{$16.0\pm1.5$} \\
\midrule
\multirow{2}{*}{CarCircle1} & $22.2\pm0.8$ & $22.3\pm0.7$ & $22.5\pm0.7$ & $22.6\pm0.8$ \\
 & \textcolor{red}{$0.3\pm0.2$} & \textcolor{red}{$0.6\pm0.6$} & \textcolor{red}{$2.8\pm2.6$} & \textcolor{red}{$8.1\pm4.9$} \\
\midrule
\multirow{2}{*}{CarCircle2} & $17.3\pm0.3$ & $17.6\pm0.3$ & $18.0\pm0.4$ & $18.2\pm0.5$ \\
 & \textcolor{red}{$0.1\pm0.0$} & \textcolor{red}{$0.7\pm0.8$} & \textcolor{red}{$3.3\pm3.2$} & \textcolor{red}{$9.3\pm5.7$} \\
\midrule
\multirow{2}{*}{CarPush1} & $10.1\pm0.4$ & $10.7\pm0.3$ & $11.1\pm0.2$ & $11.3\pm0.1$ \\
 & \textcolor{red}{$1.7\pm0.4$} & \textcolor{red}{$2.4\pm0.5$} & \textcolor{red}{$5.2\pm0.4$} & \textcolor{red}{$9.4\pm0.8$} \\
\midrule
\multirow{2}{*}{CarPush2} & $1.6\pm0.2$ & $2.8\pm0.6$ & $4.5\pm1.0$ & $5.3\pm1.0$ \\
 & \textcolor{red}{$1.7\pm0.6$} & \textcolor{red}{$2.6\pm0.5$} & \textcolor{red}{$6.2\pm0.2$} & \textcolor{red}{$12.4\pm0.9$} \\
\bottomrule
\end{tabular}
\end{adjustbox}
\end{table}


\begin{table}[htbp!]
\centering
\caption{Results of UCP-NR (No Randomization), trained separately for each $z_0$.}
\label{tab:single_orig_combined}
\begin{adjustbox}{max width=1.2\linewidth,center}
\begin{tabular}{lcccc}
\toprule
Environment & $z_0=0$ & $z_0=-5$ & $z_0=-15$ & $z_0=-25$ \\
\midrule
\multicolumn{5}{l}{\textit{Velocity Environments}} \\
\midrule
\multirow{2}{*}{Ant} & $3328.1\pm27.3$ & $3207.5\pm69.1$ & $3051.7\pm288.1$ & $3122.8\pm105.8$ \\
 & \textcolor{red}{$0.8\pm1.8$} & \textcolor{red}{$6.0\pm1.1$} & \textcolor{red}{$17.3\pm6.2$} & \textcolor{red}{$24.7\pm0.7$} \\
\midrule
\multirow{2}{*}{HalfCheetah} & $3027.2\pm6.4$ & $3059.2\pm15.2$ & $2994.6\pm69.1$ & $3025.8\pm54.4$ \\
 & \textcolor{red}{$0.1\pm0.2$} & \textcolor{red}{$3.6\pm0.5$} & \textcolor{red}{$14.2\pm0.2$} & \textcolor{red}{$25.3\pm1.9$} \\
\midrule
\multirow{2}{*}{Hopper} & $1609.2\pm71.1$ & $1478.3\pm432.7$ & $1602.2\pm111.7$ & $1486.5\pm223.1$ \\
 & \textcolor{red}{$0.2\pm0.4$} & \textcolor{red}{$2.8\pm2.4$} & \textcolor{red}{$5.3\pm7.0$} & \textcolor{red}{$12.4\pm7.5$} \\
\midrule
\multirow{2}{*}{Humanoid} & $6410.4\pm60.7$ & $6376.5\pm49.2$ & $6284.2\pm122.0$ & $6284.1\pm169.0$ \\
 & \textcolor{red}{$0.2\pm0.3$} & \textcolor{red}{$1.4\pm2.1$} & \textcolor{red}{$0.1\pm0.1$} & \textcolor{red}{$0.0\pm0.1$} \\
\midrule
\multirow{2}{*}{Swimmer} & $161.8\pm1.3$ & $165.8\pm1.1$ & $157.4\pm22.7$ & $167.5\pm5.6$ \\
 & \textcolor{red}{$0.0\pm0.0$} & \textcolor{red}{$3.1\pm0.7$} & \textcolor{red}{$11.9\pm3.4$} & \textcolor{red}{$20.8\pm2.3$} \\
\midrule
\multirow{2}{*}{Walker2d} & $2852.3\pm564.8$ & $2943.3\pm223.1$ & $2955.2\pm114.3$ & $2926.2\pm87.0$ \\
 & \textcolor{red}{$0.5\pm0.8$} & \textcolor{red}{$2.0\pm2.6$} & \textcolor{red}{$6.4\pm3.7$} & \textcolor{red}{$6.0\pm3.4$} \\
\midrule
\multicolumn{5}{l}{\textit{Navigation Environments}} \\
\midrule
\multirow{2}{*}{PointGoal1} & $1.6\pm2.1$ & $13.6\pm3.5$ & $20.5\pm4.7$ & $23.3\pm0.7$ \\
 & \textcolor{red}{$0.7\pm0.6$} & \textcolor{red}{$1.6\pm0.9$} & \textcolor{red}{$4.0\pm4.0$} & \textcolor{red}{$1.9\pm0.5$} \\
\midrule
\multirow{2}{*}{PointGoal2} & $-0.0\pm0.2$ & $0.9\pm0.9$ & $4.5\pm0.8$ & $8.4\pm1.2$ \\
 & \textcolor{red}{$1.0\pm1.5$} & \textcolor{red}{$1.0\pm0.7$} & \textcolor{red}{$3.2\pm1.5$} & \textcolor{red}{$8.9\pm2.5$} \\
\midrule
\multirow{2}{*}{PointButton1} & $-1.0\pm0.2$ & $1.8\pm0.9$ & $5.7\pm0.9$ & $8.9\pm0.9$ \\
 & \textcolor{red}{$1.7\pm0.5$} & \textcolor{red}{$1.5\pm0.4$} & \textcolor{red}{$3.6\pm1.4$} & \textcolor{red}{$6.2\pm0.4$} \\
\midrule
\multirow{2}{*}{PointButton2} & $-1.1\pm0.6$ & $0.5\pm0.3$ & $3.0\pm0.3$ & $4.9\pm1.0$ \\
 & \textcolor{red}{$1.8\pm0.5$} & \textcolor{red}{$1.9\pm0.9$} & \textcolor{red}{$3.4\pm1.0$} & \textcolor{red}{$5.1\pm1.0$} \\
\midrule
\multirow{2}{*}{PointCircle1} & $16.0\pm12.7$ & $46.3\pm2.1$ & $49.6\pm0.8$ & $49.7\pm1.8$ \\
 & \textcolor{red}{$0.1\pm0.1$} & \textcolor{red}{$0.3\pm0.2$} & \textcolor{red}{$0.2\pm0.4$} & \textcolor{red}{$5.0\pm11.9$} \\
\midrule
\multirow{2}{*}{PointCircle2} & $39.1\pm1.3$ & $36.2\pm14.4$ & $42.1\pm0.8$ & $42.5\pm0.5$ \\
 & \textcolor{red}{$0.1\pm0.1$} & \textcolor{red}{$4.3\pm6.8$} & \textcolor{red}{$1.3\pm3.3$} & \textcolor{red}{$4.1\pm8.9$} \\
\midrule
\multirow{2}{*}{PointPush1} & $8.9\pm5.1$ & $12.5\pm3.8$ & $12.4\pm5.1$ & $10.8\pm2.8$ \\
 & \textcolor{red}{$0.5\pm0.2$} & \textcolor{red}{$1.0\pm0.3$} & \textcolor{red}{$1.3\pm0.8$} & \textcolor{red}{$1.9\pm0.3$} \\
\midrule
\multirow{2}{*}{PointPush2} & $0.0\pm0.0$ & $1.7\pm0.3$ & $1.9\pm0.4$ & $3.3\pm2.4$ \\
 & \textcolor{red}{$0.2\pm0.2$} & \textcolor{red}{$1.0\pm1.2$} & \textcolor{red}{$0.9\pm0.4$} & \textcolor{red}{$2.5\pm1.2$} \\
\midrule
\multirow{2}{*}{CarGoal1} & $9.6\pm3.8$ & $27.6\pm1.1$ & $28.8\pm1.1$ & $30.1\pm2.6$ \\
 & \textcolor{red}{$0.7\pm0.5$} & \textcolor{red}{$1.1\pm0.4$} & \textcolor{red}{$1.5\pm0.2$} & \textcolor{red}{$2.9\pm0.9$} \\
\midrule
\multirow{2}{*}{CarGoal2} & $0.3\pm0.1$ & $2.1\pm0.8$ & $8.2\pm0.5$ & $10.4\pm1.3$ \\
 & \textcolor{red}{$1.0\pm0.4$} & \textcolor{red}{$1.7\pm1.3$} & \textcolor{red}{$5.7\pm0.8$} & \textcolor{red}{$6.7\pm2.5$} \\
\midrule
\multirow{2}{*}{CarButton1} & $-0.1\pm0.2$ & $0.7\pm0.6$ & $3.3\pm0.7$ & $3.9\pm0.8$ \\
 & \textcolor{red}{$0.7\pm1.1$} & \textcolor{red}{$1.4\pm1.3$} & \textcolor{red}{$3.6\pm0.8$} & \textcolor{red}{$7.2\pm3.5$} \\
\midrule
\multirow{2}{*}{CarButton2} & $0.0\pm0.0$ & $0.1\pm0.2$ & $1.7\pm0.3$ & $2.0\pm1.3$ \\
 & \textcolor{red}{$1.1\pm2.3$} & \textcolor{red}{$1.2\pm1.5$} & \textcolor{red}{$3.0\pm1.1$} & \textcolor{red}{$5.7\pm1.9$} \\
\midrule
\multirow{2}{*}{CarCircle1} & $11.8\pm7.1$ & $20.2\pm2.3$ & $21.5\pm0.5$ & $19.4\pm6.3$ \\
 & \textcolor{red}{$0.2\pm0.2$} & \textcolor{red}{$0.9\pm1.4$} & \textcolor{red}{$0.6\pm0.5$} & \textcolor{red}{$3.1\pm5.8$} \\
\midrule
\multirow{2}{*}{CarCircle2} & $15.7\pm1.7$ & $15.4\pm2.9$ & $16.1\pm2.1$ & $17.5\pm0.6$ \\
 & \textcolor{red}{$0.1\pm0.2$} & \textcolor{red}{$2.2\pm3.6$} & \textcolor{red}{$2.3\pm2.6$} & \textcolor{red}{$2.3\pm2.3$} \\
\midrule
\multirow{2}{*}{CarPush1} & $7.4\pm1.2$ & $8.6\pm0.7$ & $10.2\pm0.5$ & $10.6\pm0.6$ \\
 & \textcolor{red}{$0.3\pm0.1$} & \textcolor{red}{$0.5\pm0.2$} & \textcolor{red}{$1.1\pm0.2$} & \textcolor{red}{$2.0\pm0.8$} \\
\midrule
\multirow{2}{*}{CarPush2} & $0.7\pm0.6$ & $1.9\pm0.6$ & $3.8\pm0.4$ & $4.4\pm1.1$ \\
 & \textcolor{red}{$0.8\pm0.8$} & \textcolor{red}{$1.1\pm0.3$} & \textcolor{red}{$2.9\pm0.6$} & \textcolor{red}{$4.6\pm0.7$} \\
\bottomrule
\end{tabular}
\end{adjustbox}
\end{table}


\begin{table}[htbp!]
\centering
\caption{Results of UCP-Mean, evaluated at different initial stocks $z_0$.}
\label{tab:ctx_noabs_combined}
\begin{adjustbox}{max width=1.2\linewidth,center}
\begin{tabular}{lcccc}
\toprule
Environment & $z_0=0$ & $z_0=-5$ & $z_0=-15$ & $z_0=-25$ \\
\midrule
\multicolumn{5}{l}{\textit{Velocity Environments}} \\
\midrule
\multirow{2}{*}{Ant} & $3268.3\pm131.9$ & $3285.3\pm119.1$ & $3290.4\pm125.8$ & $3281.2\pm108.6$ \\
 & \textcolor{red}{$6.3\pm12.0$} & \textcolor{red}{$7.8\pm12.0$} & \textcolor{red}{$12.5\pm12.6$} & \textcolor{red}{$20.2\pm12.8$} \\
\midrule
\multirow{2}{*}{HalfCheetah} & $3051.5\pm10.9$ & $3054.9\pm12.3$ & $3050.9\pm12.4$ & $3055.7\pm9.2$ \\
 & \textcolor{red}{$0.0\pm0.0$} & \textcolor{red}{$0.5\pm1.0$} & \textcolor{red}{$2.1\pm2.3$} & \textcolor{red}{$12.3\pm2.5$} \\
\midrule
\multirow{2}{*}{Hopper} & $841.7\pm614.2$ & $1273.7\pm430.2$ & $1264.8\pm632.6$ & $1281.7\pm581.0$ \\
 & \textcolor{red}{$0.5\pm0.8$} & \textcolor{red}{$8.9\pm11.4$} & \textcolor{red}{$13.7\pm15.7$} & \textcolor{red}{$28.5\pm12.9$} \\
\midrule
\multirow{2}{*}{Humanoid} & $6409.0\pm57.3$ & $6427.6\pm64.8$ & $6433.5\pm109.0$ & $6407.5\pm193.2$ \\
 & \textcolor{red}{$0.9\pm2.2$} & \textcolor{red}{$3.4\pm4.9$} & \textcolor{red}{$8.9\pm9.4$} & \textcolor{red}{$14.2\pm14.4$} \\
\midrule
\multirow{2}{*}{Swimmer} & $-11.2\pm9.1$ & $3.5\pm15.3$ & $1.6\pm3.0$ & $9.2\pm16.5$ \\
 & \textcolor{red}{$0.5\pm0.9$} & \textcolor{red}{$1.3\pm2.1$} & \textcolor{red}{$2.8\pm6.8$} & \textcolor{red}{$6.5\pm10.2$} \\
\midrule
\multirow{2}{*}{Walker2d} & $632.6\pm622.6$ & $2578.3\pm460.6$ & $2728.0\pm290.5$ & $2775.0\pm333.0$ \\
 & \textcolor{red}{$3.4\pm7.1$} & \textcolor{red}{$10.2\pm10.6$} & \textcolor{red}{$15.6\pm9.9$} & \textcolor{red}{$23.1\pm8.4$} \\
\midrule
\multicolumn{5}{l}{\textit{Navigation Environments}} \\
\midrule
\multirow{2}{*}{PointGoal1} & $1.9\pm0.9$ & $25.4\pm0.6$ & $25.2\pm0.2$ & $25.9\pm0.9$ \\
 & \textcolor{red}{$3.3\pm1.4$} & \textcolor{red}{$12.6\pm3.8$} & \textcolor{red}{$19.5\pm3.1$} & \textcolor{red}{$23.7\pm2.6$} \\
\midrule
\multirow{2}{*}{PointGoal2} & $0.3\pm0.4$ & $7.8\pm2.2$ & $10.1\pm3.5$ & $11.7\pm4.2$ \\
 & \textcolor{red}{$4.7\pm6.1$} & \textcolor{red}{$16.0\pm11.4$} & \textcolor{red}{$20.4\pm6.4$} & \textcolor{red}{$28.3\pm3.4$} \\
\midrule
\multirow{2}{*}{PointButton1} & $2.1\pm0.7$ & $10.8\pm1.0$ & $15.1\pm0.8$ & $17.8\pm0.7$ \\
 & \textcolor{red}{$6.3\pm1.5$} & \textcolor{red}{$10.1\pm1.0$} & \textcolor{red}{$18.6\pm1.2$} & \textcolor{red}{$27.7\pm1.4$} \\
\midrule
\multirow{2}{*}{PointButton2} & $1.4\pm0.3$ & $7.9\pm0.4$ & $11.4\pm0.4$ & $13.5\pm0.5$ \\
 & \textcolor{red}{$6.4\pm1.2$} & \textcolor{red}{$9.6\pm0.7$} & \textcolor{red}{$18.5\pm2.2$} & \textcolor{red}{$27.5\pm1.1$} \\
\midrule
\multirow{2}{*}{PointCircle1} & $17.1\pm13.8$ & $50.8\pm1.0$ & $51.1\pm1.3$ & $51.2\pm1.1$ \\
 & \textcolor{red}{$0.8\pm1.5$} & \textcolor{red}{$19.0\pm19.0$} & \textcolor{red}{$32.7\pm30.7$} & \textcolor{red}{$38.4\pm32.0$} \\
\midrule
\multirow{2}{*}{PointCircle2} & $30.7\pm4.5$ & $43.0\pm1.2$ & $43.4\pm0.7$ & $43.4\pm0.9$ \\
 & \textcolor{red}{$3.5\pm8.9$} & \textcolor{red}{$11.5\pm16.8$} & \textcolor{red}{$22.5\pm19.0$} & \textcolor{red}{$25.4\pm32.0$} \\
\midrule
\multirow{2}{*}{PointPush1} & $0.5\pm0.5$ & $17.2\pm4.2$ & $17.8\pm4.2$ & $18.0\pm4.3$ \\
 & \textcolor{red}{$1.4\pm0.5$} & \textcolor{red}{$12.1\pm1.2$} & \textcolor{red}{$17.2\pm0.8$} & \textcolor{red}{$19.3\pm0.6$} \\
\midrule
\multirow{2}{*}{PointPush2} & $-0.2\pm0.6$ & $5.8\pm2.3$ & $7.8\pm3.6$ & $8.5\pm3.7$ \\
 & \textcolor{red}{$2.4\pm0.7$} & \textcolor{red}{$15.0\pm9.1$} & \textcolor{red}{$20.9\pm8.0$} & \textcolor{red}{$25.7\pm6.5$} \\
\midrule
\multirow{2}{*}{CarGoal1} & $4.5\pm1.7$ & $32.5\pm1.2$ & $33.3\pm0.9$ & $34.0\pm0.9$ \\
 & \textcolor{red}{$6.5\pm2.6$} & \textcolor{red}{$12.2\pm1.4$} & \textcolor{red}{$20.5\pm1.9$} & \textcolor{red}{$24.4\pm0.7$} \\
\midrule
\multirow{2}{*}{CarGoal2} & $1.6\pm0.6$ & $9.3\pm1.1$ & $12.3\pm0.4$ & $14.6\pm0.4$ \\
 & \textcolor{red}{$10.1\pm1.4$} & \textcolor{red}{$12.4\pm1.0$} & \textcolor{red}{$19.5\pm1.7$} & \textcolor{red}{$30.2\pm1.2$} \\
\midrule
\multirow{2}{*}{CarButton1} & $1.1\pm0.5$ & $5.9\pm0.6$ & $7.7\pm0.8$ & $8.7\pm0.9$ \\
 & \textcolor{red}{$8.6\pm3.2$} & \textcolor{red}{$12.2\pm1.7$} & \textcolor{red}{$20.8\pm2.0$} & \textcolor{red}{$27.7\pm2.4$} \\
\midrule
\multirow{2}{*}{CarButton2} & $0.2\pm0.8$ & $4.3\pm0.2$ & $5.7\pm1.2$ & $7.5\pm0.6$ \\
 & \textcolor{red}{$7.4\pm2.8$} & \textcolor{red}{$10.9\pm1.1$} & \textcolor{red}{$19.4\pm1.1$} & \textcolor{red}{$30.6\pm2.8$} \\
\midrule
\multirow{2}{*}{CarCircle1} & $0.5\pm2.2$ & $21.9\pm0.8$ & $22.1\pm1.1$ & $22.2\pm1.1$ \\
 & \textcolor{red}{$1.7\pm2.2$} & \textcolor{red}{$24.5\pm24.9$} & \textcolor{red}{$24.8\pm14.9$} & \textcolor{red}{$33.5\pm17.3$} \\
\midrule
\multirow{2}{*}{CarCircle2} & $8.4\pm8.5$ & $17.9\pm1.2$ & $18.1\pm1.1$ & $18.4\pm1.1$ \\
 & \textcolor{red}{$6.5\pm11.9$} & \textcolor{red}{$51.0\pm27.5$} & \textcolor{red}{$64.8\pm29.8$} & \textcolor{red}{$83.9\pm42.2$} \\
\midrule
\multirow{2}{*}{CarPush1} & $-0.1\pm0.5$ & $11.4\pm0.2$ & $11.7\pm0.3$ & $12.0\pm0.2$ \\
 & \textcolor{red}{$1.7\pm0.2$} & \textcolor{red}{$12.3\pm1.8$} & \textcolor{red}{$18.3\pm2.3$} & \textcolor{red}{$20.5\pm2.1$} \\
\midrule
\multirow{2}{*}{CarPush2} & $-0.8\pm0.6$ & $5.1\pm1.6$ & $6.5\pm1.8$ & $7.0\pm1.7$ \\
 & \textcolor{red}{$5.1\pm2.2$} & \textcolor{red}{$14.2\pm1.0$} & \textcolor{red}{$20.9\pm0.7$} & \textcolor{red}{$28.5\pm2.4$} \\
\bottomrule
\end{tabular}
\end{adjustbox}
\end{table}


\begin{table}[htbp!]
\centering
\caption{Results of UCP-NR-Mean (No Randomization), trained separately for each $z_0$.}
\label{tab:single_noabs_combined}
\begin{adjustbox}{max width=1.2\linewidth,center}
\begin{tabular}{lcccc}
\toprule
Environment & $z_0=0$ & $z_0=-5$ & $z_0=-15$ & $z_0=-25$ \\
\midrule
\multicolumn{5}{l}{\textit{Velocity Environments}} \\
\midrule
\multirow{2}{*}{Ant} & $3288.1\pm68.0$ & $3230.5\pm215.1$ & $3081.7\pm192.6$ & $2940.7\pm366.9$ \\
 & \textcolor{red}{$1.4\pm1.8$} & \textcolor{red}{$2.2\pm3.6$} & \textcolor{red}{$8.9\pm5.7$} & \textcolor{red}{$13.0\pm12.7$} \\
\midrule
\multirow{2}{*}{HalfCheetah} & $3030.1\pm10.4$ & $3036.6\pm18.2$ & $3037.4\pm25.9$ & $3035.8\pm31.0$ \\
 & \textcolor{red}{$0.0\pm0.0$} & \textcolor{red}{$0.0\pm0.0$} & \textcolor{red}{$0.1\pm0.2$} & \textcolor{red}{$0.3\pm0.4$} \\
\midrule
\multirow{2}{*}{Hopper} & $721.2\pm877.3$ & $508.8\pm889.1$ & $547.5\pm905.4$ & $616.9\pm795.7$ \\
 & \textcolor{red}{$5.6\pm8.3$} & \textcolor{red}{$3.8\pm1.1$} & \textcolor{red}{$12.5\pm11.7$} & \textcolor{red}{$17.3\pm19.9$} \\
\midrule
\multirow{2}{*}{Humanoid} & $6019.4\pm728.9$ & $5463.1\pm2315.1$ & $6296.7\pm228.7$ & $6281.7\pm184.6$ \\
 & \textcolor{red}{$1.9\pm4.7$} & \textcolor{red}{$3.1\pm4.9$} & \textcolor{red}{$0.8\pm1.4$} & \textcolor{red}{$3.5\pm8.8$} \\
\midrule
\multirow{2}{*}{Swimmer} & $-0.5\pm3.4$ & $-1.7\pm4.9$ & $72.0\pm90.4$ & $21.1\pm21.5$ \\
 & \textcolor{red}{$0.2\pm0.5$} & \textcolor{red}{$0.0\pm0.0$} & \textcolor{red}{$4.9\pm7.0$} & \textcolor{red}{$10.7\pm7.9$} \\
\midrule
\multirow{2}{*}{Walker2d} & $2700.7\pm901.3$ & $2333.1\pm1892.1$ & $2795.8\pm572.8$ & $2948.7\pm84.9$ \\
 & \textcolor{red}{$0.4\pm0.7$} & \textcolor{red}{$2.5\pm2.5$} & \textcolor{red}{$9.3\pm11.6$} & \textcolor{red}{$8.9\pm7.7$} \\
\midrule
\multicolumn{5}{l}{\textit{Navigation Environments}} \\
\midrule
\multirow{2}{*}{PointGoal1} & $1.7\pm1.2$ & $25.1\pm0.7$ & $27.1\pm0.7$ & $27.4\pm1.3$ \\
 & \textcolor{red}{$0.8\pm0.2$} & \textcolor{red}{$5.9\pm1.7$} & \textcolor{red}{$12.0\pm6.6$} & \textcolor{red}{$24.9\pm16.4$} \\
\midrule
\multirow{2}{*}{PointGoal2} & $-0.1\pm0.1$ & $6.0\pm1.3$ & $11.3\pm2.6$ & $15.9\pm3.5$ \\
 & \textcolor{red}{$0.9\pm0.6$} & \textcolor{red}{$5.4\pm1.7$} & \textcolor{red}{$15.0\pm7.2$} & \textcolor{red}{$24.2\pm3.0$} \\
\midrule
\multirow{2}{*}{PointButton1} & $-1.3\pm0.3$ & $7.7\pm1.5$ & $16.1\pm0.4$ & $20.9\pm2.6$ \\
 & \textcolor{red}{$1.4\pm0.5$} & \textcolor{red}{$5.4\pm1.0$} & \textcolor{red}{$13.0\pm1.9$} & \textcolor{red}{$26.8\pm10.6$} \\
\midrule
\multirow{2}{*}{PointButton2} & $-1.3\pm0.8$ & $5.2\pm1.0$ & $10.7\pm4.2$ & $15.9\pm1.4$ \\
 & \textcolor{red}{$2.1\pm0.8$} & \textcolor{red}{$5.4\pm1.0$} & \textcolor{red}{$16.8\pm4.9$} & \textcolor{red}{$26.4\pm4.4$} \\
\midrule
\multirow{2}{*}{PointCircle1} & $36.0\pm7.2$ & $49.6\pm2.1$ & $50.9\pm1.0$ & $51.3\pm2.4$ \\
 & \textcolor{red}{$1.9\pm3.8$} & \textcolor{red}{$6.6\pm9.1$} & \textcolor{red}{$10.5\pm9.7$} & \textcolor{red}{$26.5\pm41.7$} \\
\midrule
\multirow{2}{*}{PointCircle2} & $36.2\pm5.9$ & $42.6\pm0.4$ & $42.5\pm0.6$ & $43.0\pm0.3$ \\
 & \textcolor{red}{$0.3\pm0.2$} & \textcolor{red}{$3.4\pm8.0$} & \textcolor{red}{$1.8\pm1.6$} & \textcolor{red}{$4.5\pm9.8$} \\
\midrule
\multirow{2}{*}{PointPush1} & $5.1\pm3.0$ & $15.0\pm5.1$ & $16.8\pm4.2$ & $18.4\pm5.2$ \\
 & \textcolor{red}{$0.5\pm0.4$} & \textcolor{red}{$4.2\pm1.7$} & \textcolor{red}{$11.7\pm6.0$} & \textcolor{red}{$21.0\pm11.6$} \\
\midrule
\multirow{2}{*}{PointPush2} & $0.0\pm0.0$ & $8.1\pm3.3$ & $14.3\pm0.9$ & $11.5\pm3.7$ \\
 & \textcolor{red}{$0.3\pm0.3$} & \textcolor{red}{$6.3\pm1.7$} & \textcolor{red}{$12.8\pm4.1$} & \textcolor{red}{$19.1\pm12.8$} \\
\midrule
\multirow{2}{*}{CarGoal1} & $7.5\pm3.8$ & $34.0\pm1.0$ & $36.3\pm0.6$ & $36.8\pm0.4$ \\
 & \textcolor{red}{$0.5\pm0.2$} & \textcolor{red}{$6.1\pm2.7$} & \textcolor{red}{$13.8\pm10.8$} & \textcolor{red}{$21.1\pm6.5$} \\
\midrule
\multirow{2}{*}{CarGoal2} & $0.1\pm0.1$ & $9.3\pm0.5$ & $18.1\pm2.0$ & $22.5\pm0.9$ \\
 & \textcolor{red}{$0.4\pm0.4$} & \textcolor{red}{$6.3\pm1.2$} & \textcolor{red}{$22.1\pm5.8$} & \textcolor{red}{$28.2\pm5.7$} \\
\midrule
\multirow{2}{*}{CarButton1} & $-0.4\pm1.1$ & $5.3\pm0.6$ & $7.6\pm1.0$ & $10.4\pm2.9$ \\
 & \textcolor{red}{$0.5\pm0.8$} & \textcolor{red}{$6.6\pm2.5$} & \textcolor{red}{$13.6\pm7.0$} & \textcolor{red}{$29.1\pm25.5$} \\
\midrule
\multirow{2}{*}{CarButton2} & $-0.0\pm0.1$ & $3.6\pm0.4$ & $6.8\pm1.0$ & $7.9\pm1.5$ \\
 & \textcolor{red}{$0.5\pm1.1$} & \textcolor{red}{$5.8\pm1.3$} & \textcolor{red}{$18.5\pm5.5$} & \textcolor{red}{$24.1\pm9.9$} \\
\midrule
\multirow{2}{*}{CarCircle1} & $10.5\pm5.8$ & $21.7\pm0.5$ & $22.1\pm0.7$ & $21.7\pm0.7$ \\
 & \textcolor{red}{$1.8\pm2.4$} & \textcolor{red}{$10.3\pm12.8$} & \textcolor{red}{$37.5\pm53.9$} & \textcolor{red}{$3.9\pm3.1$} \\
\midrule
\multirow{2}{*}{CarCircle2} & $11.7\pm5.2$ & $17.3\pm0.8$ & $17.5\pm0.7$ & $17.6\pm1.3$ \\
 & \textcolor{red}{$2.8\pm4.2$} & \textcolor{red}{$3.2\pm1.5$} & \textcolor{red}{$12.7\pm20.0$} & \textcolor{red}{$7.9\pm8.3$} \\
\midrule
\multirow{2}{*}{CarPush1} & $7.2\pm1.6$ & $11.9\pm0.5$ & $12.7\pm0.6$ & $13.5\pm0.3$ \\
 & \textcolor{red}{$0.5\pm0.3$} & \textcolor{red}{$5.0\pm3.6$} & \textcolor{red}{$10.5\pm9.0$} & \textcolor{red}{$26.6\pm0.9$} \\
\midrule
\multirow{2}{*}{CarPush2} & $1.0\pm0.3$ & $5.5\pm0.7$ & $9.4\pm2.0$ & $11.9\pm0.5$ \\
 & \textcolor{red}{$0.5\pm0.5$} & \textcolor{red}{$7.6\pm2.5$} & \textcolor{red}{$19.9\pm12.0$} & \textcolor{red}{$31.4\pm10.0$} \\
\bottomrule
\end{tabular}
\end{adjustbox}
\end{table}


\begin{table}[htbp!]
\centering
\caption{Results of UCP-NA (No Stock-Augmentation), trained separately for each $z_0$.}
\label{tab:single_nocs_combined}
\begin{adjustbox}{max width=1.2\linewidth,center}
\begin{tabular}{lcccc}
\toprule
Environment & $z_0=0$ & $z_0=-5$ & $z_0=-15$ & $z_0=-25$ \\
\midrule
\multicolumn{5}{l}{\textit{Velocity Environments}} \\
\midrule
\multirow{2}{*}{Ant} & $3336.2\pm31.7$ & $3322.9\pm78.4$ & $2873.2\pm770.7$ & $2903.1\pm1053.9$ \\
 & \textcolor{red}{$0.4\pm0.3$} & \textcolor{red}{$0.6\pm0.3$} & \textcolor{red}{$22.4\pm30.9$} & \textcolor{red}{$16.6\pm22.7$} \\
\midrule
\multirow{2}{*}{HalfCheetah} & $3037.9\pm3.2$ & $3044.2\pm8.6$ & $3061.6\pm10.1$ & $3074.2\pm7.2$ \\
 & \textcolor{red}{$0.0\pm0.0$} & \textcolor{red}{$0.0\pm0.0$} & \textcolor{red}{$0.1\pm0.0$} & \textcolor{red}{$0.0\pm0.0$} \\
\midrule
\multirow{2}{*}{Hopper} & $998.6\pm955.9$ & $409.4\pm638.4$ & $1183.1\pm396.8$ & $1423.9\pm305.2$ \\
 & \textcolor{red}{$1.9\pm5.1$} & \textcolor{red}{$7.8\pm21.3$} & \textcolor{red}{$23.2\pm23.5$} & \textcolor{red}{$44.6\pm57.0$} \\
\midrule
\multirow{2}{*}{Humanoid} & $5952.3\pm962.3$ & $5745.1\pm1604.9$ & $6307.2\pm144.4$ & $6410.7\pm178.7$ \\
 & \textcolor{red}{$2.4\pm6.0$} & \textcolor{red}{$1.9\pm4.4$} & \textcolor{red}{$0.1\pm0.2$} & \textcolor{red}{$16.0\pm44.0$} \\
\midrule
\multirow{2}{*}{Swimmer} & $-0.2\pm2.5$ & $170.4\pm1.1$ & $165.8\pm16.4$ & $167.9\pm14.8$ \\
 & \textcolor{red}{$0.1\pm0.2$} & \textcolor{red}{$0.0\pm0.0$} & \textcolor{red}{$0.6\pm1.2$} & \textcolor{red}{$6.0\pm9.9$} \\
\midrule
\multirow{2}{*}{Walker2d} & $2780.7\pm559.1$ & $2964.3\pm518.7$ & $3028.4\pm351.5$ & $3189.3\pm26.6$ \\
 & \textcolor{red}{$1.8\pm2.6$} & \textcolor{red}{$26.9\pm39.5$} & \textcolor{red}{$32.0\pm68.6$} & \textcolor{red}{$24.4\pm37.9$} \\
\midrule
\multicolumn{5}{l}{\textit{Navigation Environments}} \\
\midrule
\multirow{2}{*}{PointGoal1} & $-0.6\pm0.6$ & $26.8\pm0.3$ & $27.9\pm0.8$ & $28.5\pm0.5$ \\
 & \textcolor{red}{$8.2\pm4.1$} & \textcolor{red}{$4.2\pm1.1$} & \textcolor{red}{$17.2\pm17.0$} & \textcolor{red}{$26.2\pm20.7$} \\
\midrule
\multirow{2}{*}{PointGoal2} & $-0.5\pm0.7$ & $-0.4\pm0.2$ & $2.0\pm2.6$ & $14.4\pm2.9$ \\
 & \textcolor{red}{$14.9\pm3.5$} & \textcolor{red}{$15.4\pm6.9$} & \textcolor{red}{$14.0\pm3.4$} & \textcolor{red}{$21.0\pm6.1$} \\
\midrule
\multirow{2}{*}{PointButton1} & $-0.9\pm0.5$ & $7.6\pm0.5$ & $16.0\pm3.8$ & $21.4\pm3.9$ \\
 & \textcolor{red}{$3.1\pm1.6$} & \textcolor{red}{$5.2\pm1.1$} & \textcolor{red}{$18.1\pm8.6$} & \textcolor{red}{$29.8\pm11.0$} \\
\midrule
\multirow{2}{*}{PointButton2} & $-1.2\pm1.0$ & $5.4\pm0.7$ & $13.4\pm1.5$ & $15.2\pm5.8$ \\
 & \textcolor{red}{$2.5\pm1.8$} & \textcolor{red}{$8.3\pm1.8$} & \textcolor{red}{$21.5\pm6.1$} & \textcolor{red}{$27.2\pm18.6$} \\
\midrule
\multirow{2}{*}{PointCircle1} & $47.1\pm1.5$ & $51.3\pm0.7$ & $51.3\pm0.4$ & $52.2\pm2.6$ \\
 & \textcolor{red}{$0.1\pm0.2$} & \textcolor{red}{$7.4\pm7.1$} & \textcolor{red}{$6.5\pm14.6$} & \textcolor{red}{$54.9\pm79.9$} \\
\midrule
\multirow{2}{*}{PointCircle2} & $41.7\pm0.8$ & $43.1\pm0.4$ & $43.0\pm0.3$ & $42.8\pm0.5$ \\
 & \textcolor{red}{$0.5\pm0.5$} & \textcolor{red}{$2.5\pm4.7$} & \textcolor{red}{$3.4\pm6.5$} & \textcolor{red}{$5.1\pm11.4$} \\
\midrule
\multirow{2}{*}{PointPush1} & $0.8\pm0.9$ & $18.5\pm3.5$ & $21.3\pm0.5$ & $20.2\pm3.4$ \\
 & \textcolor{red}{$1.3\pm1.3$} & \textcolor{red}{$8.2\pm4.1$} & \textcolor{red}{$18.2\pm7.5$} & \textcolor{red}{$26.3\pm9.6$} \\
\midrule
\multirow{2}{*}{PointPush2} & $-0.2\pm0.1$ & $6.0\pm3.7$ & $13.4\pm4.4$ & $14.9\pm3.3$ \\
 & \textcolor{red}{$3.6\pm2.1$} & \textcolor{red}{$9.2\pm5.0$} & \textcolor{red}{$15.9\pm6.8$} & \textcolor{red}{$23.7\pm12.1$} \\
\midrule
\multirow{2}{*}{CarGoal1} & $-0.1\pm0.8$ & $35.5\pm0.5$ & $36.7\pm0.5$ & $36.7\pm0.3$ \\
 & \textcolor{red}{$2.1\pm2.7$} & \textcolor{red}{$5.6\pm2.1$} & \textcolor{red}{$21.1\pm10.5$} & \textcolor{red}{$19.2\pm5.8$} \\
\midrule
\multirow{2}{*}{CarGoal2} & $-0.1\pm0.2$ & $0.7\pm1.2$ & $9.8\pm7.1$ & $21.1\pm2.3$ \\
 & \textcolor{red}{$0.6\pm0.4$} & \textcolor{red}{$9.9\pm5.0$} & \textcolor{red}{$13.0\pm5.8$} & \textcolor{red}{$29.7\pm6.6$} \\
\midrule
\multirow{2}{*}{CarButton1} & $-1.6\pm1.0$ & $4.9\pm1.0$ & $8.6\pm2.4$ & $9.6\pm2.5$ \\
 & \textcolor{red}{$2.1\pm2.0$} & \textcolor{red}{$6.8\pm2.4$} & \textcolor{red}{$19.6\pm13.0$} & \textcolor{red}{$25.9\pm14.2$} \\
\midrule
\multirow{2}{*}{CarButton2} & $-2.4\pm1.5$ & $3.2\pm0.4$ & $6.6\pm1.4$ & $7.6\pm2.7$ \\
 & \textcolor{red}{$3.0\pm0.7$} & \textcolor{red}{$6.6\pm1.8$} & \textcolor{red}{$16.9\pm6.8$} & \textcolor{red}{$25.8\pm14.8$} \\
\midrule
\multirow{2}{*}{CarCircle1} & $19.6\pm0.5$ & $21.9\pm0.7$ & $21.9\pm0.9$ & $21.9\pm1.0$ \\
 & \textcolor{red}{$1.4\pm2.1$} & \textcolor{red}{$2.3\pm2.5$} & \textcolor{red}{$1.0\pm1.4$} & \textcolor{red}{$28.3\pm53.5$} \\
\midrule
\multirow{2}{*}{CarCircle2} & $15.8\pm0.8$ & $17.2\pm1.1$ & $17.7\pm0.7$ & $17.7\pm0.8$ \\
 & \textcolor{red}{$0.9\pm1.4$} & \textcolor{red}{$3.7\pm6.5$} & \textcolor{red}{$3.6\pm2.7$} & \textcolor{red}{$4.8\pm4.6$} \\
\midrule
\multirow{2}{*}{CarPush1} & $1.5\pm2.2$ & $12.0\pm1.6$ & $13.1\pm0.9$ & $13.9\pm0.2$ \\
 & \textcolor{red}{$0.5\pm0.6$} & \textcolor{red}{$6.0\pm4.2$} & \textcolor{red}{$13.3\pm11.3$} & \textcolor{red}{$25.3\pm6.6$} \\
\midrule
\multirow{2}{*}{CarPush2} & $-1.2\pm0.7$ & $2.7\pm0.8$ & $9.2\pm1.5$ & $12.5\pm0.3$ \\
 & \textcolor{red}{$3.9\pm2.1$} & \textcolor{red}{$10.0\pm3.9$} & \textcolor{red}{$13.6\pm7.1$} & \textcolor{red}{$43.9\pm13.8$} \\
\bottomrule
\end{tabular}
\end{adjustbox}
\end{table}


\begin{table}[htbp!]
\centering
\caption{Results of UCP-DS (Discounted Stock), evaluated at different initial stocks $z_0$.}
\label{tab:ctx_cgamma_combined}
\begin{adjustbox}{max width=1.2\linewidth,center}
\begin{tabular}{lcccc}
\toprule
Environment & $z_0=0$ & $z_0=-5$ & $z_0=-15$ & $z_0=-25$ \\
\midrule
\multicolumn{5}{l}{\textit{Velocity Environments}} \\
\midrule
\multirow{2}{*}{Ant} & $2437.5\pm395.5$ & $2853.6\pm392.4$ & $2780.9\pm411.3$ & $2282.9\pm395.3$ \\
 & \textcolor{red}{$0.5\pm0.4$} & \textcolor{red}{$35.9\pm29.5$} & \textcolor{red}{$48.6\pm29.1$} & \textcolor{red}{$57.1\pm24.5$} \\
\midrule
\multirow{2}{*}{HalfCheetah} & $2705.4\pm400.8$ & $3027.6\pm53.0$ & $3043.6\pm66.7$ & $3052.3\pm72.6$ \\
 & \textcolor{red}{$2.7\pm5.7$} & \textcolor{red}{$92.2\pm53.3$} & \textcolor{red}{$110.8\pm62.4$} & \textcolor{red}{$129.7\pm63.6$} \\
\midrule
\multirow{2}{*}{Hopper} & $1051.6\pm546.3$ & $1155.6\pm708.2$ & $1123.6\pm493.8$ & $1016.5\pm319.5$ \\
 & \textcolor{red}{$0.0\pm0.0$} & \textcolor{red}{$261.5\pm247.7$} & \textcolor{red}{$295.2\pm171.2$} & \textcolor{red}{$253.2\pm92.4$} \\
\midrule
\multirow{2}{*}{Humanoid} & $6237.7\pm45.2$ & $6281.4\pm76.0$ & $6290.9\pm78.5$ & $6293.2\pm82.3$ \\
 & \textcolor{red}{$0.0\pm0.1$} & \textcolor{red}{$0.0\pm0.1$} & \textcolor{red}{$0.0\pm0.0$} & \textcolor{red}{$0.0\pm0.0$} \\
\midrule
\multirow{2}{*}{Swimmer} & $12.8\pm15.5$ & $44.8\pm3.7$ & $49.7\pm3.8$ & $45.7\pm4.6$ \\
 & \textcolor{red}{$0.0\pm0.1$} & \textcolor{red}{$71.3\pm49.0$} & \textcolor{red}{$88.7\pm56.8$} & \textcolor{red}{$107.7\pm68.6$} \\
\midrule
\multirow{2}{*}{Walker2d} & $2227.0\pm460.5$ & $2527.0\pm780.9$ & $2252.9\pm538.6$ & $1974.6\pm707.3$ \\
 & \textcolor{red}{$0.4\pm0.8$} & \textcolor{red}{$249.4\pm194.7$} & \textcolor{red}{$290.2\pm210.9$} & \textcolor{red}{$273.5\pm210.6$} \\
\midrule
\multicolumn{5}{l}{\textit{Navigation Environments}} \\
\midrule
\multirow{2}{*}{PointGoal1} & $15.2\pm1.7$ & $20.9\pm1.4$ & $24.9\pm0.8$ & $26.0\pm0.6$ \\
 & \textcolor{red}{$2.4\pm0.7$} & \textcolor{red}{$2.9\pm0.9$} & \textcolor{red}{$8.1\pm1.6$} & \textcolor{red}{$16.8\pm1.9$} \\
\midrule
\multirow{2}{*}{PointGoal2} & $1.1\pm1.4$ & $2.9\pm1.7$ & $5.9\pm3.1$ & $10.0\pm3.3$ \\
 & \textcolor{red}{$2.9\pm1.8$} & \textcolor{red}{$3.0\pm1.8$} & \textcolor{red}{$7.4\pm2.1$} & \textcolor{red}{$20.9\pm3.3$} \\
\midrule
\multirow{2}{*}{PointButton1} & $-0.5\pm0.4$ & $5.6\pm1.4$ & $10.3\pm1.3$ & $14.4\pm0.9$ \\
 & \textcolor{red}{$2.1\pm0.9$} & \textcolor{red}{$3.3\pm0.9$} & \textcolor{red}{$8.7\pm1.0$} & \textcolor{red}{$22.4\pm1.9$} \\
\midrule
\multirow{2}{*}{PointButton2} & $-1.4\pm0.6$ & $3.1\pm0.7$ & $6.8\pm0.7$ & $10.5\pm0.3$ \\
 & \textcolor{red}{$3.5\pm0.9$} & \textcolor{red}{$3.3\pm1.1$} & \textcolor{red}{$8.6\pm1.4$} & \textcolor{red}{$21.6\pm1.5$} \\
\midrule
\multirow{2}{*}{PointCircle1} & $45.5\pm3.6$ & $47.5\pm3.0$ & $50.0\pm1.5$ & $51.2\pm1.2$ \\
 & \textcolor{red}{$0.9\pm1.5$} & \textcolor{red}{$13.1\pm3.9$} & \textcolor{red}{$52.1\pm6.2$} & \textcolor{red}{$83.6\pm4.9$} \\
\midrule
\multirow{2}{*}{PointCircle2} & $41.0\pm0.2$ & $40.8\pm0.9$ & $42.7\pm0.6$ & $43.3\pm1.1$ \\
 & \textcolor{red}{$1.9\pm4.7$} & \textcolor{red}{$11.3\pm3.8$} & \textcolor{red}{$49.2\pm7.7$} & \textcolor{red}{$88.5\pm10.6$} \\
\midrule
\multirow{2}{*}{PointPush1} & $10.4\pm4.4$ & $12.5\pm4.4$ & $14.4\pm4.5$ & $15.3\pm4.7$ \\
 & \textcolor{red}{$1.6\pm0.6$} & \textcolor{red}{$2.2\pm0.5$} & \textcolor{red}{$6.4\pm0.7$} & \textcolor{red}{$13.6\pm1.2$} \\
\midrule
\multirow{2}{*}{PointPush2} & $1.2\pm0.4$ & $3.9\pm1.5$ & $6.3\pm2.4$ & $7.2\pm2.6$ \\
 & \textcolor{red}{$0.9\pm0.2$} & \textcolor{red}{$2.7\pm0.3$} & \textcolor{red}{$7.3\pm1.4$} & \textcolor{red}{$14.3\pm2.6$} \\
\midrule
\multirow{2}{*}{CarGoal1} & $1.0\pm4.2$ & $20.0\pm6.3$ & $30.5\pm1.9$ & $32.4\pm1.3$ \\
 & \textcolor{red}{$0.7\pm0.4$} & \textcolor{red}{$1.7\pm1.2$} & \textcolor{red}{$8.7\pm1.2$} & \textcolor{red}{$18.9\pm1.5$} \\
\midrule
\multirow{2}{*}{CarGoal2} & $-1.5\pm0.8$ & $0.0\pm0.8$ & $6.3\pm3.9$ & $10.1\pm4.8$ \\
 & \textcolor{red}{$3.7\pm1.4$} & \textcolor{red}{$2.5\pm1.8$} & \textcolor{red}{$8.3\pm1.3$} & \textcolor{red}{$24.4\pm1.9$} \\
\midrule
\multirow{2}{*}{CarButton1} & $-13.4\pm14.4$ & $-8.4\pm14.6$ & $-6.7\pm18.9$ & $-3.6\pm14.7$ \\
 & \textcolor{red}{$2.9\pm2.1$} & \textcolor{red}{$3.1\pm2.3$} & \textcolor{red}{$5.4\pm3.0$} & \textcolor{red}{$10.1\pm10.3$} \\
\midrule
\multirow{2}{*}{CarButton2} & $-8.3\pm11.0$ & $-1.6\pm1.6$ & $1.6\pm2.0$ & $2.9\pm2.7$ \\
 & \textcolor{red}{$3.3\pm3.9$} & \textcolor{red}{$2.3\pm2.7$} & \textcolor{red}{$4.8\pm3.9$} & \textcolor{red}{$16.0\pm5.0$} \\
\midrule
\multirow{2}{*}{CarCircle1} & $18.2\pm1.7$ & $19.2\pm1.2$ & $20.9\pm1.0$ & $21.8\pm1.0$ \\
 & \textcolor{red}{$4.6\pm4.9$} & \textcolor{red}{$18.6\pm9.3$} & \textcolor{red}{$57.0\pm2.3$} & \textcolor{red}{$90.4\pm5.0$} \\
\midrule
\multirow{2}{*}{CarCircle2} & $16.0\pm0.9$ & $16.1\pm1.1$ & $17.3\pm0.9$ & $17.7\pm0.9$ \\
 & \textcolor{red}{$1.6\pm3.0$} & \textcolor{red}{$8.3\pm6.1$} & \textcolor{red}{$57.2\pm5.8$} & \textcolor{red}{$88.2\pm7.3$} \\
\midrule
\multirow{2}{*}{CarPush1} & $6.9\pm4.2$ & $9.0\pm2.1$ & $10.5\pm1.0$ & $10.9\pm0.8$ \\
 & \textcolor{red}{$1.1\pm1.0$} & \textcolor{red}{$1.9\pm1.1$} & \textcolor{red}{$5.5\pm1.3$} & \textcolor{red}{$10.4\pm0.7$} \\
\midrule
\multirow{2}{*}{CarPush2} & $-0.6\pm4.8$ & $-4.3\pm19.9$ & $-0.9\pm17.9$ & $2.4\pm11.4$ \\
 & \textcolor{red}{$3.7\pm5.4$} & \textcolor{red}{$2.9\pm2.5$} & \textcolor{red}{$10.5\pm9.2$} & \textcolor{red}{$17.5\pm4.0$} \\
\bottomrule
\end{tabular}
\end{adjustbox}
\end{table}


\section{Effect of Lower Budgets on a Baseline}

In \cref{fig:comparison_navigation}, all 16 Safety Navigation results use the same budget of $25$ for UCP and the baselines, yet UCP achieves higher return at lower cost.
For Safety Velocity in \cref{fig:comparison_velocity}, however, we selected UCP's best feasible budget after training, although its performance remains strong at budget $25$.
This raises the question of whether training with a lower budget could help other methods too.
We therefore trained PPOSaute at budget $5$, and we compare the results with the original budget-$25$ results in \cref{tab:pposaute_budget_velocity,tab:pposaute_budget_navigation}.
The lower budget reduces mean cost in all 22 environments but generally yields the same or lower return; the only clear return improvements are in Swimmer and Walker2d.
Moreover, budget-$5$ PPOSaute satisfies its nominal budget in only five of the six Velocity tasks and in none of the 16 Navigation tasks.
Simply training a baseline with a smaller nominal budget therefore does not reproduce UCP's gains.
\begin{table}[htbp!]
\centering
\caption{PPOSaute trained with cost budgets $b=5$ and $b=25$, evaluated on Velocity environments.}
\label{tab:pposaute_budget_velocity}
\begin{adjustbox}{max width=1.2\linewidth,center}
\begin{tabular}{lcc}
\toprule
Environment & PPOSaute ($b=5$) & PPOSaute ($b=25$) \\
\midrule
\multirow{2}{*}{Ant} & $2704.4\pm45.2$ & $2711.2\pm84.6$ \\
 & \textcolor{red}{$3.9\pm0.2$} & \textcolor{red}{$17.5\pm0.4$} \\
\midrule
\multirow{2}{*}{HalfCheetah} & $2743.1\pm56.0$ & $2792.7\pm31.7$ \\
 & \textcolor{red}{$3.7\pm0.1$} & \textcolor{red}{$17.1\pm0.2$} \\
\midrule
\multirow{2}{*}{Hopper} & $1697.0\pm22.0$ & $1656.4\pm29.0$ \\
 & \textcolor{red}{$3.3\pm0.1$} & \textcolor{red}{$17.9\pm0.4$} \\
\midrule
\multirow{2}{*}{Humanoid} & $5896.3\pm84.0$ & $5986.0\pm42.5$ \\
 & \textcolor{red}{$2.7\pm0.1$} & \textcolor{red}{$20.4\pm0.8$} \\
\midrule
\multirow{2}{*}{Swimmer} & $141.1\pm9.9$ & $111.6\pm72.4$ \\
 & \textcolor{red}{$5.8\pm3.0$} & \textcolor{red}{$12.5\pm6.7$} \\
\midrule
\multirow{2}{*}{Walker2d} & $3023.3\pm88.7$ & $2064.9\pm1381.5$ \\
 & \textcolor{red}{$3.6\pm0.1$} & \textcolor{red}{$18.5\pm0.8$} \\
\bottomrule
\end{tabular}
\end{adjustbox}
\end{table}

\begin{table}[htbp!]
\centering
\caption{PPOSaute trained with cost budgets $b=5$ and $b=25$, evaluated on Navigation environments.}
\label{tab:pposaute_budget_navigation}
\begin{adjustbox}{max width=1.2\linewidth,center}
\begin{tabular}{lcc}
\toprule
Environment & PPOSaute ($b=5$) & PPOSaute ($b=25$) \\
\midrule
\multirow{2}{*}{PointGoal1} & $3.7\pm1.4$ & $6.2\pm3.3$ \\
 & \textcolor{red}{$9.6\pm2.1$} & \textcolor{red}{$21.1\pm3.3$} \\
\midrule
\multirow{2}{*}{PointGoal2} & $0.2\pm0.7$ & $0.6\pm1.0$ \\
 & \textcolor{red}{$17.4\pm2.9$} & \textcolor{red}{$28.1\pm3.6$} \\
\midrule
\multirow{2}{*}{PointButton1} & $-1.2\pm0.5$ & $0.8\pm1.3$ \\
 & \textcolor{red}{$10.9\pm1.4$} & \textcolor{red}{$26.4\pm10.8$} \\
\midrule
\multirow{2}{*}{PointButton2} & $-1.7\pm0.7$ & $-0.2\pm0.3$ \\
 & \textcolor{red}{$14.4\pm3.2$} & \textcolor{red}{$29.4\pm12.6$} \\
\midrule
\multirow{2}{*}{PointCircle1} & $41.5\pm2.1$ & $42.4\pm1.8$ \\
 & \textcolor{red}{$6.1\pm3.4$} & \textcolor{red}{$14.3\pm6.7$} \\
\midrule
\multirow{2}{*}{PointCircle2} & $38.9\pm0.5$ & $38.6\pm1.2$ \\
 & \textcolor{red}{$9.6\pm3.7$} & \textcolor{red}{$19.7\pm10.2$} \\
\midrule
\multirow{2}{*}{PointPush1} & $0.5\pm0.1$ & $0.3\pm0.2$ \\
 & \textcolor{red}{$7.4\pm0.8$} & \textcolor{red}{$9.9\pm7.2$} \\
\midrule
\multirow{2}{*}{PointPush2} & $0.3\pm0.1$ & $0.1\pm0.5$ \\
 & \textcolor{red}{$10.2\pm2.7$} & \textcolor{red}{$13.5\pm3.7$} \\
\midrule
\multirow{2}{*}{CarGoal1} & $3.7\pm1.6$ & $9.6\pm2.9$ \\
 & \textcolor{red}{$11.2\pm5.4$} & \textcolor{red}{$22.6\pm6.0$} \\
\midrule
\multirow{2}{*}{CarGoal2} & $0.1\pm0.2$ & $0.6\pm0.4$ \\
 & \textcolor{red}{$10.1\pm3.6$} & \textcolor{red}{$20.4\pm6.8$} \\
\midrule
\multirow{2}{*}{CarButton1} & $0.1\pm0.2$ & $-0.3\pm0.8$ \\
 & \textcolor{red}{$9.3\pm3.4$} & \textcolor{red}{$31.6\pm8.5$} \\
\midrule
\multirow{2}{*}{CarButton2} & $-0.1\pm0.3$ & $0.0\pm0.3$ \\
 & \textcolor{red}{$18.1\pm6.5$} & \textcolor{red}{$40.0\pm10.6$} \\
\midrule
\multirow{2}{*}{CarCircle1} & $13.4\pm1.1$ & $13.5\pm1.5$ \\
 & \textcolor{red}{$8.8\pm5.8$} & \textcolor{red}{$13.2\pm4.8$} \\
\midrule
\multirow{2}{*}{CarCircle2} & $12.7\pm0.9$ & $13.0\pm0.6$ \\
 & \textcolor{red}{$10.3\pm2.9$} & \textcolor{red}{$16.8\pm5.0$} \\
\midrule
\multirow{2}{*}{CarPush1} & $0.2\pm0.4$ & $0.4\pm0.2$ \\
 & \textcolor{red}{$7.6\pm2.7$} & \textcolor{red}{$16.1\pm6.2$} \\
\midrule
\multirow{2}{*}{CarPush2} & $0.2\pm0.1$ & $0.3\pm0.1$ \\
 & \textcolor{red}{$16.4\pm6.0$} & \textcolor{red}{$26.8\pm4.6$} \\
\bottomrule
\end{tabular}
\end{adjustbox}
\end{table}

\clearpage

\section{Multiple Constraints}
\label{app:multiconstraint}

The main results presented in \cref{sec:experiments} consider the one-cost setting from the Safety Gymnasium, where multiple cost signals are aggregated into one; however,
the UCMDP framework supports vector-valued costs, and UCP is algorithmically compatible with such a case.
In the CarButton2, CarPush2, and CarGoal2 tasks, the cost (as defined for the Safety Gymnasium benchmark) is a combination of multiple components.
Therefore, it is natural for us to consider evaluating UCP with multiple constraints in these tasks by constraining the cost components separately.

In this section, we present a study of UCP under multiple constraints, a variant we call MC-UCP (multi-constraint UCP).
We trained MC-UCP on CarButton2, CarPush2, and CarGoal2 for three million steps each, matching the experimental design in \cref{sec:experiments}. 
We used one nonnegative multiplier per constraint and one distributional cost critic per stock dimension for vector-valued stock.
Every cost dimension uses the same range and discretization as scalar-cost UCP ($K=31$ equally spaced values in $[-30, 0]$).
The initial stock of each cost dimension is sampled independently, so no component receives a privileged range or additional training.
At evaluation, an initial-stock vector specifies the limits independently. For example, $(-25, 0)$ relaxes only the first cost limit, keeping the second at zero.
We use five seeds and $1000$ evaluation episodes per seed and initial-stock vector.

We highlight three interesting findings from this section:
\begin{itemize}
    \item[-] MC-UCP is competitive with or better than UCP (which uses a single constraint on the aggregate cost) in the original benchmark evaluation, where its total cost must not exceed $25$.
    \item[-] MC-UCP controls cost components separately (when the task allows).
    \item[-] MC-UCP does not spend unusable budget (when the task allows).
\end{itemize}

The results for CarButton2 are presented in \cref{tab:multiconstraint_carbutton2}.
The three additional cost dimensions of the stock ($(z_{0, 2}, z_{0, 3}, z_{0, 4})$) correspond to the costs from buttons, gremlins and hazards, in this order.
The total cost (which is not directly constrained) is the scalar cost as measured in the benchmark version of the task.
\begin{table}[htbp!]
\centering
\caption{MC-UCP on CarButton2, evaluated at different initial cost stocks. Intervals are 95\% Student's $t$ intervals across five seeds.}
\label{tab:multiconstraint_carbutton2}
\begin{adjustbox}{max width=1.2\linewidth,center}
\begin{tabular}{lccccc}
\toprule
$(z_{0, 2},z_{0, 3},z_{0, 4})$ & Return & Total cost & Buttons & Gremlins & Hazards \\
\midrule
$(0, 0, 0)$ & $-0.23\pm1.71$ & \textcolor{red}{$5.07\pm1.23$} & \textcolor{red}{$1.80\pm0.68$} & \textcolor{red}{$1.73\pm0.27$} & \textcolor{red}{$1.63\pm0.63$} \\
$(-15, 0, 0)$ & $2.17\pm0.86$ & \textcolor{red}{$8.32\pm1.22$} & \textcolor{red}{$4.89\pm1.00$} & \textcolor{red}{$2.10\pm0.11$} & \textcolor{red}{$1.57\pm0.39$} \\
$(0, -15, 0)$ & $2.56\pm0.94$ & \textcolor{red}{$7.38\pm0.77$} & \textcolor{red}{$1.38\pm0.52$} & \textcolor{red}{$4.13\pm0.34$} & \textcolor{red}{$1.96\pm1.20$} \\
$(0, 0, -15)$ & $1.67\pm1.00$ & \textcolor{red}{$5.11\pm1.12$} & \textcolor{red}{$0.87\pm0.10$} & \textcolor{red}{$1.48\pm0.16$} & \textcolor{red}{$2.79\pm1.10$} \\
$(-25, 0, 0)$ & $2.25\pm0.52$ & \textcolor{red}{$9.38\pm1.25$} & \textcolor{red}{$5.55\pm0.77$} & \textcolor{red}{$2.18\pm0.10$} & \textcolor{red}{$1.94\pm0.71$} \\
$(0, -25, 0)$ & $2.76\pm0.96$ & \textcolor{red}{$10.51\pm0.72$} & \textcolor{red}{$1.43\pm0.47$} & \textcolor{red}{$7.08\pm0.80$} & \textcolor{red}{$2.06\pm1.22$} \\
$(0, 0, -25)$ & $1.57\pm0.96$ & \textcolor{red}{$6.78\pm1.28$} & \textcolor{red}{$0.94\pm0.17$} & \textcolor{red}{$1.65\pm0.31$} & \textcolor{red}{$4.24\pm1.17$} \\
$(-5, -5, -5)$ & $3.10\pm0.78$ & \textcolor{red}{$5.65\pm0.59$} & \textcolor{red}{$1.59\pm0.23$} & \textcolor{red}{$2.41\pm0.27$} & \textcolor{red}{$1.71\pm0.74$} \\
$(-10, -10, -10)$ & $4.52\pm0.53$ & \textcolor{red}{$10.06\pm0.96$} & \textcolor{red}{$3.60\pm0.83$} & \textcolor{red}{$3.83\pm0.19$} & \textcolor{red}{$2.86\pm0.36$} \\
$(-15, -15, -15)$ & $5.64\pm0.39$ & \textcolor{red}{$16.86\pm1.21$} & \textcolor{red}{$6.27\pm1.24$} & \textcolor{red}{$5.81\pm0.50$} & \textcolor{red}{$5.34\pm0.43$} \\
$(-25, -25, -25)$ & $7.34\pm0.47$ & \textcolor{red}{$35.93\pm3.08$} & \textcolor{red}{$13.56\pm1.29$} & \textcolor{red}{$12.06\pm0.98$} & \textcolor{red}{$11.33\pm1.17$} \\
\bottomrule
\end{tabular}
\end{adjustbox}
\end{table}
We see that MC-UCP attains higher return while still satisfying the benchmark limit of $25$ on the cost.
MC-UCP with $z_{0, (2, 3, 4)} = (-15, -15, -15)$ attains return $5.64$ with a total cost of $16.86$, whereas
UCP has a return of $4.7$ and a total cost of $16$ (see \cref{tab:comparison_combined}).
CarButton2 allows for behaviors that incur different types of costs (contact with more or fewer buttons, gremlins or hazards), and MC-UCP, accordingly, controls the cost components separately.
Relative to an all-zero initial stock, relaxing one coordinate to $-25$ primarily raises the matching cost, with buttons going from $1.80$ to $5.55$, gremlins from $1.73$ to $7.08$, and hazards from $1.63$ to $4.24$, while the other components remain comparatively low.

The results for CarGoal2 are presented in \cref{tab:multiconstraint_cargoal2}.
The three additional cost dimensions of the stock ($(z_{0, 2}, z_{0, 3}, z_{0, 4})$) correspond to the costs from hazards, contact with vases and the velocity of vases, in this order.
The total cost (which is not directly constrained) is the scalar cost as measured in the benchmark version of the task.
\begin{table}[htbp!]
\centering
\caption{MC-UCP on CarGoal2, evaluated at different initial cost stocks. Intervals are 95\% Student's $t$ intervals across five seeds.}
\label{tab:multiconstraint_cargoal2}
\begin{adjustbox}{max width=1.2\linewidth,center}
\begin{tabular}{lccccc}
\toprule
$(z_{0, 2},z_{0, 3},z_{0, 4})$ & Return & Total cost & Hazards & Vases Contact & Vases Velocity \\
\midrule
$(0, 0, 0)$ & $-0.33\pm0.57$ & \textcolor{red}{$4.30\pm2.18$} & \textcolor{red}{$2.66\pm1.74$} & \textcolor{red}{$0.34\pm0.16$} & \textcolor{red}{$1.67\pm0.47$} \\
$(-15, 0, 0)$ & $1.78\pm1.37$ & \textcolor{red}{$3.94\pm1.22$} & \textcolor{red}{$2.92\pm0.87$} & \textcolor{red}{$0.19\pm0.12$} & \textcolor{red}{$1.02\pm0.43$} \\
$(0, -15, 0)$ & $1.49\pm1.10$ & \textcolor{red}{$3.67\pm2.65$} & \textcolor{red}{$1.96\pm1.23$} & \textcolor{red}{$0.36\pm0.42$} & \textcolor{red}{$1.73\pm1.51$} \\
$(0, 0, -15)$ & $4.18\pm1.85$ & \textcolor{red}{$4.15\pm0.66$} & \textcolor{red}{$1.44\pm0.47$} & \textcolor{red}{$0.45\pm0.11$} & \textcolor{red}{$2.71\pm0.23$} \\
$(-25, 0, 0)$ & $2.36\pm1.30$ & \textcolor{red}{$7.40\pm0.99$} & \textcolor{red}{$5.89\pm0.82$} & \textcolor{red}{$0.29\pm0.13$} & \textcolor{red}{$1.52\pm0.23$} \\
$(0, -25, 0)$ & $2.17\pm1.65$ & \textcolor{red}{$3.38\pm1.27$} & \textcolor{red}{$1.38\pm0.39$} & \textcolor{red}{$0.38\pm0.22$} & \textcolor{red}{$1.99\pm0.89$} \\
$(0, 0, -25)$ & $6.58\pm2.05$ & \textcolor{red}{$7.53\pm0.88$} & \textcolor{red}{$2.36\pm0.69$} & \textcolor{red}{$0.87\pm0.17$} & \textcolor{red}{$5.16\pm0.28$} \\
$(-15, 0, -15)$ & $9.55\pm1.44$ & \textcolor{red}{$7.77\pm1.61$} & \textcolor{red}{$5.23\pm0.83$} & \textcolor{red}{$0.38\pm0.11$} & \textcolor{red}{$2.54\pm0.86$} \\
$(-25, 0, -25)$ & $13.30\pm1.60$ & \textcolor{red}{$19.25\pm2.01$} & \textcolor{red}{$12.57\pm0.60$} & \textcolor{red}{$0.98\pm0.39$} & \textcolor{red}{$6.76\pm1.67$} \\
$(-5, -5, -5)$ & $5.24\pm1.16$ & \textcolor{red}{$2.82\pm0.32$} & \textcolor{red}{$1.42\pm0.43$} & \textcolor{red}{$0.22\pm0.03$} & \textcolor{red}{$1.41\pm0.13$} \\
$(-10, -10, -10)$ & $10.55\pm0.32$ & \textcolor{red}{$6.26\pm1.34$} & \textcolor{red}{$3.41\pm0.49$} & \textcolor{red}{$0.40\pm0.16$} & \textcolor{red}{$2.84\pm0.94$} \\
$(-15, -15, -15)$ & $14.15\pm0.42$ & \textcolor{red}{$12.45\pm1.69$} & \textcolor{red}{$6.55\pm0.58$} & \textcolor{red}{$0.82\pm0.21$} & \textcolor{red}{$5.92\pm1.41$} \\
$(-25, -25, -25)$ & $16.36\pm0.68$ & \textcolor{red}{$27.85\pm2.39$} & \textcolor{red}{$13.72\pm0.83$} & \textcolor{red}{$2.19\pm0.46$} & \textcolor{red}{$14.25\pm2.45$} \\
\bottomrule
\end{tabular}
\end{adjustbox}
\end{table}
MC-UCP outperforms UCP at lower total cost with the same number of training steps.
MC-UCP with $z_{0, (2, 3, 4)} = (-15, -15, -15)$ attains return $14.15$ with a total cost of $12.45$, whereas
UCP has a return of $12.5$ and a total cost of $17.2$ (see \cref{tab:comparison_combined}).
CarGoal2 also separates hazard and vase-velocity behavior. Relaxing only hazards raises hazard cost from $2.66$ to $5.89$, and relaxing only vase velocity raises that cost from $1.67$ to $5.16$.
In CarGoal2, vase contacts are relatively rare at lower constraint budgets, so the policy does not spend the budget when only that constraint is relaxed. With $z_{0, (2, 3, 4)} = (0, -25, 0)$, vase-contact cost barely changes (from $0.34$ to $0.38$), while return increases from $-0.33$ to $2.17$. By contrast, jointly relaxing hazards and vase velocity produces a large return improvement (consider $z_{0, (2, 3, 4)} = (-25, 0, -25)$, with return $13.30$, versus $2.36$ and $6.58$ when relaxing either coordinate alone).

The results for CarPush2 are presented in \cref{tab:multiconstraint_carpush2}.
The two additional cost dimensions of the stock ($(z_{0, 2}, z_{0, 3})$) correspond to the costs from hazards and contact with pillars, in this order.
The total cost (which is not directly constrained) is the scalar cost as measured in the benchmark version of the task.
\begin{table}[htbp!]
\centering
\caption{MC-UCP on CarPush2, evaluated at different initial cost stocks. Intervals are 95\% Student's $t$ intervals across five seeds.}
\label{tab:multiconstraint_carpush2}
\begin{adjustbox}{max width=1.2\linewidth,center}
\begin{tabular}{lcccc}
\toprule
$(z_{0, 2},z_{0, 3})$ & Return & Total cost & Hazards & Pillars \\
\midrule
$(0, 0)$ & $0.95\pm0.43$ & \textcolor{red}{$2.81\pm0.59$} & \textcolor{red}{$1.59\pm0.50$} & \textcolor{red}{$1.22\pm0.14$} \\
$(-15, 0)$ & $3.47\pm0.91$ & \textcolor{red}{$5.99\pm0.29$} & \textcolor{red}{$4.56\pm0.58$} & \textcolor{red}{$1.43\pm0.31$} \\
$(0, -15)$ & $2.44\pm0.81$ & \textcolor{red}{$4.69\pm0.49$} & \textcolor{red}{$1.16\pm0.17$} & \textcolor{red}{$3.54\pm0.38$} \\
$(-25, 0)$ & $3.72\pm0.77$ & \textcolor{red}{$8.89\pm1.21$} & \textcolor{red}{$7.86\pm1.00$} & \textcolor{red}{$1.03\pm0.22$} \\
$(0, -25)$ & $2.64\pm0.70$ & \textcolor{red}{$6.63\pm1.09$} & \textcolor{red}{$1.40\pm0.64$} & \textcolor{red}{$5.23\pm0.47$} \\
$(-5, -5)$ & $3.24\pm1.12$ & \textcolor{red}{$3.66\pm0.30$} & \textcolor{red}{$1.92\pm0.11$} & \textcolor{red}{$1.74\pm0.22$} \\
$(-10, -10)$ & $4.61\pm1.17$ & \textcolor{red}{$6.72\pm1.23$} & \textcolor{red}{$3.26\pm0.20$} & \textcolor{red}{$3.47\pm1.07$} \\
$(-15, -15)$ & $5.16\pm1.24$ & \textcolor{red}{$9.42\pm0.77$} & \textcolor{red}{$5.31\pm0.66$} & \textcolor{red}{$4.11\pm0.11$} \\
$(-25, -25)$ & $5.55\pm1.03$ & \textcolor{red}{$17.87\pm1.48$} & \textcolor{red}{$10.69\pm0.93$} & \textcolor{red}{$7.19\pm0.80$} \\
\bottomrule
\end{tabular}
\end{adjustbox}
\end{table}
In this task, MC-UCP with $z_{0, (2, 3)} = (-15, -15)$ attains a similar return at lower total cost, with return $5.16$ and a total cost of $9.42$, whereas UCP has a return of $5.3$ and a total cost of $12.4$ (see \cref{tab:comparison_combined}).
Similar to CarButton2, CarPush2 also allows for behaviors that incur different types of costs, and MC-UCP, accordingly, can control them separately. For example, $z_{0, (2, 3)} = (-25, 0)$ raises hazard cost to $7.86$ while pillar cost remains $1.03$, whereas $z_{0, (2, 3)} = (0, -25)$ raises pillar cost to $5.23$ while hazard cost remains $1.40$.

These asymmetric and joint-budget evaluations show more than aggregate cost control as one trained policy can reveal and independently tune distinct modes of unsafe behavior. This also makes the trade-off between cost and reward in each task clearly visible, which allows for better budget selection.

\newpage
\section*{NeurIPS Paper Checklist}

\begin{enumerate}

\item {\bf Claims}
    \item[] Question: Do the main claims made in the abstract and introduction accurately reflect the paper's contributions and scope?
    \item[] Answer: \answerYes{}
    \item[] Justification: Our framework is discussed in \cref{sec:methodology} and our experiments and ablation studies are presented in \cref{sec:experiments}.
    \item[] Guidelines:
    \begin{itemize}
        \item The answer \answerNA{} means that the abstract and introduction do not include the claims made in the paper.
        \item The abstract and/or introduction should clearly state the claims made, including the contributions made in the paper and important assumptions and limitations. A \answerNo{} or \answerNA{} answer to this question will not be perceived well by the reviewers. 
        \item The claims made should match theoretical and experimental results, and reflect how much the results can be expected to generalize to other settings. 
        \item It is fine to include aspirational goals as motivation as long as it is clear that these goals are not attained by the paper. 
    \end{itemize}

\item {\bf Limitations}
    \item[] Question: Does the paper discuss the limitations of the work performed by the authors?
    \item[] Answer: \answerYes{}
    \item[] Justification: We discuss our limitations in \cref{sec:conclusion}.
    \item[] Guidelines:
    \begin{itemize}
        \item The answer \answerNA{} means that the paper has no limitation while the answer \answerNo{} means that the paper has limitations, but those are not discussed in the paper. 
        \item The authors are encouraged to create a separate ``Limitations'' section in their paper.
        \item The paper should point out any strong assumptions and how robust the results are to violations of these assumptions (e.g., independence assumptions, noiseless settings, model well-specification, asymptotic approximations only holding locally). The authors should reflect on how these assumptions might be violated in practice and what the implications would be.
        \item The authors should reflect on the scope of the claims made, e.g., if the approach was only tested on a few datasets or with a few runs. In general, empirical results often depend on implicit assumptions, which should be articulated.
        \item The authors should reflect on the factors that influence the performance of the approach. For example, a facial recognition algorithm may perform poorly when image resolution is low or images are taken in low lighting. Or a speech-to-text system might not be used reliably to provide closed captions for online lectures because it fails to handle technical jargon.
        \item The authors should discuss the computational efficiency of the proposed algorithms and how they scale with dataset size.
        \item If applicable, the authors should discuss possible limitations of their approach to address problems of privacy and fairness.
        \item While the authors might fear that complete honesty about limitations might be used by reviewers as grounds for rejection, a worse outcome might be that reviewers discover limitations that aren't acknowledged in the paper. The authors should use their best judgment and recognize that individual actions in favor of transparency play an important role in developing norms that preserve the integrity of the community. Reviewers will be specifically instructed to not penalize honesty concerning limitations.
    \end{itemize}

\item {\bf Theory assumptions and proofs}
    \item[] Question: For each theoretical result, does the paper provide the full set of assumptions and a complete (and correct) proof?
    \item[] Answer: \answerNA{}
    \item[] Justification: The paper does not include any theoretical results.
    \item[] Guidelines:
    \begin{itemize}
        \item The answer \answerNA{} means that the paper does not include theoretical results. 
        \item All the theorems, formulas, and proofs in the paper should be numbered and cross-referenced.
        \item All assumptions should be clearly stated or referenced in the statement of any theorems.
        \item The proofs can either appear in the main paper or the supplemental material, but if they appear in the supplemental material, the authors are encouraged to provide a short proof sketch to provide intuition. 
        \item Inversely, any informal proof provided in the core of the paper should be complemented by formal proofs provided in appendix or supplemental material.
        \item Theorems and Lemmas that the proof relies upon should be properly referenced. 
    \end{itemize}

    \item {\bf Experimental result reproducibility}
    \item[] Question: Does the paper fully disclose all the information needed to reproduce the main experimental results of the paper to the extent that it affects the main claims and/or conclusions of the paper (regardless of whether the code and data are provided or not)?
    \item[] Answer: \answerYes{}
    \item[] Justification: We have highlighted significant components of the method in \cref{sec:methodology} and provided implementation details in \cref{app:implementation}. We have used the OmniSafe~\citep{ji2024omnisafe} library for our experiments and baselines. We have not changed the tasks or task parameters. We have not changed the baselines, except for some hyperparameters, where they led to improved baseline performance, and we have provided details in \cref{app:hyperparameters}. We have provided tables with benchmark performance for our main method and all our ablations, so that reproduced results can be adequately compared.
    \item[] Guidelines:
    \begin{itemize}
        \item The answer \answerNA{} means that the paper does not include experiments.
        \item If the paper includes experiments, a \answerNo{} answer to this question will not be perceived well by the reviewers: Making the paper reproducible is important, regardless of whether the code and data are provided or not.
        \item If the contribution is a dataset and\slash or model, the authors should describe the steps taken to make their results reproducible or verifiable. 
        \item Depending on the contribution, reproducibility can be accomplished in various ways. For example, if the contribution is a novel architecture, describing the architecture fully might suffice, or if the contribution is a specific model and empirical evaluation, it may be necessary to either make it possible for others to replicate the model with the same dataset, or provide access to the model. In general. releasing code and data is often one good way to accomplish this, but reproducibility can also be provided via detailed instructions for how to replicate the results, access to a hosted model (e.g., in the case of a large language model), releasing of a model checkpoint, or other means that are appropriate to the research performed.
        \item While NeurIPS does not require releasing code, the conference does require all submissions to provide some reasonable avenue for reproducibility, which may depend on the nature of the contribution. For example
        \begin{enumerate}
            \item If the contribution is primarily a new algorithm, the paper should make it clear how to reproduce that algorithm.
            \item If the contribution is primarily a new model architecture, the paper should describe the architecture clearly and fully.
            \item If the contribution is a new model (e.g., a large language model), then there should either be a way to access this model for reproducing the results or a way to reproduce the model (e.g., with an open-source dataset or instructions for how to construct the dataset).
            \item We recognize that reproducibility may be tricky in some cases, in which case authors are welcome to describe the particular way they provide for reproducibility. In the case of closed-source models, it may be that access to the model is limited in some way (e.g., to registered users), but it should be possible for other researchers to have some path to reproducing or verifying the results.
        \end{enumerate}
    \end{itemize}

\item {\bf Open access to data and code}
    \item[] Question: Does the paper provide open access to the data and code, with sufficient instructions to faithfully reproduce the main experimental results, as described in supplemental material?
    \item[] Answer: \answerYes{}
    \item[] Justification: The code is available at \url{https://github.com/MehrdadMoghimi/UCP}.
    \item[] Guidelines:
    \begin{itemize}
        \item The answer \answerNA{} means that paper does not include experiments requiring code.
        \item Please see the NeurIPS code and data submission guidelines (\url{https://neurips.cc/public/guides/CodeSubmissionPolicy}) for more details.
        \item While we encourage the release of code and data, we understand that this might not be possible, so \answerNo{} is an acceptable answer. Papers cannot be rejected simply for not including code, unless this is central to the contribution (e.g., for a new open-source benchmark).
        \item The instructions should contain the exact command and environment needed to run to reproduce the results. See the NeurIPS code and data submission guidelines (\url{https://neurips.cc/public/guides/CodeSubmissionPolicy}) for more details.
        \item The authors should provide instructions on data access and preparation, including how to access the raw data, preprocessed data, intermediate data, and generated data, etc.
        \item The authors should provide scripts to reproduce all experimental results for the new proposed method and baselines. If only a subset of experiments are reproducible, they should state which ones are omitted from the script and why.
        \item At submission time, to preserve anonymity, the authors should release anonymized versions (if applicable).
        \item Providing as much information as possible in supplemental material (appended to the paper) is recommended, but including URLs to data and code is permitted.
    \end{itemize}

\item {\bf Experimental setting/details}
    \item[] Question: Does the paper specify all the training and test details (e.g., data splits, hyperparameters, how they were chosen, type of optimizer) necessary to understand the results?
    \item[] Answer: \answerYes{}
    \item[] Justification: We have presented the methodology and experimental setup in \cref{sec:methodology,sec:experiments} to enough detail that the reader will be able to appreciate the experimental results, though we assume some basic knowledge of the Safety Gymnasium benchmark~\citep{ji2023safety} tasks. We have provided implementation details in \cref{app:implementation}. The baselines are established methods, and have been described in detail by \citet{ji2024omnisafe}.
    \item[] Guidelines:
    \begin{itemize}
        \item The answer \answerNA{} means that the paper does not include experiments.
        \item The experimental setting should be presented in the core of the paper to a level of detail that is necessary to appreciate the results and make sense of them.
        \item The full details can be provided either with the code, in appendix, or as supplemental material.
    \end{itemize}

\item {\bf Experiment statistical significance}
    \item[] Question: Does the paper report error bars suitably and correctly defined or other appropriate information about the statistical significance of the experiments?
    \item[] Answer: \answerYes{}
    \item[] Justification: We have reported results with 95\% confidence intervals (Student's $t$) over five independent runs for each experiment, and we believe the numbers we have presented to be representative of what will be observed in independent runs.
    \item[] Guidelines:
    \begin{itemize}
        \item The answer \answerNA{} means that the paper does not include experiments.
        \item The authors should answer \answerYes{} if the results are accompanied by error bars, confidence intervals, or statistical significance tests, at least for the experiments that support the main claims of the paper.
        \item The factors of variability that the error bars are capturing should be clearly stated (for example, train/test split, initialization, random drawing of some parameter, or overall run with given experimental conditions).
        \item The method for calculating the error bars should be explained (closed form formula, call to a library function, bootstrap, etc.)
        \item The assumptions made should be given (e.g., Normally distributed errors).
        \item It should be clear whether the error bar is the standard deviation or the standard error of the mean.
        \item It is OK to report 1-sigma error bars, but one should state it. The authors should preferably report a 2-sigma error bar than state that they have a 96\% CI, if the hypothesis of Normality of errors is not verified.
        \item For asymmetric distributions, the authors should be careful not to show in tables or figures symmetric error bars that would yield results that are out of range (e.g., negative error rates).
        \item If error bars are reported in tables or plots, the authors should explain in the text how they were calculated and reference the corresponding figures or tables in the text.
    \end{itemize}

\item {\bf Experiments compute resources}
    \item[] Question: For each experiment, does the paper provide sufficient information on the computer resources (type of compute workers, memory, time of execution) needed to reproduce the experiments?
    \item[] Answer: \answerYes{}
    \item[] Justification: We discuss this in \cref{app:compute}.
    \item[] Guidelines:
    \begin{itemize}
        \item The answer \answerNA{} means that the paper does not include experiments.
        \item The paper should indicate the type of compute workers CPU or GPU, internal cluster, or cloud provider, including relevant memory and storage.
        \item The paper should provide the amount of compute required for each of the individual experimental runs as well as estimate the total compute. 
        \item The paper should disclose whether the full research project required more compute than the experiments reported in the paper (e.g., preliminary or failed experiments that didn't make it into the paper). 
    \end{itemize}
    
\item {\bf Code of ethics}
    \item[] Question: Does the research conducted in the paper conform, in every respect, with the NeurIPS Code of Ethics \url{https://neurips.cc/public/EthicsGuidelines}?
    \item[] Answer: \answerYes{}
    \item[] Justification: The research adheres to the NeurIPS Code of Ethics, as it involves a framework and experimental results in Safe RL, which do not pose ethical risks or misuse concerns.
    \item[] Guidelines:
    \begin{itemize}
        \item The answer \answerNA{} means that the authors have not reviewed the NeurIPS Code of Ethics.
        \item If the authors answer \answerNo, they should explain the special circumstances that require a deviation from the Code of Ethics.
        \item The authors should make sure to preserve anonymity (e.g., if there is a special consideration due to laws or regulations in their jurisdiction).
    \end{itemize}

\item {\bf Broader impacts}
    \item[] Question: Does the paper discuss both potential positive societal impacts and negative societal impacts of the work performed?
    \item[] Answer: \answerNA{}
    \item[] Justification: This is foundational algorithmic research. This work introduces an RL methodology evaluated on a standard simulated benchmark without ties to specific applications or deployments, so we do not expect direct societal impact.
    \item[] Guidelines:
    \begin{itemize}
        \item The answer \answerNA{} means that there is no societal impact of the work performed.
        \item If the authors answer \answerNA{} or \answerNo, they should explain why their work has no societal impact or why the paper does not address societal impact.
        \item Examples of negative societal impacts include potential malicious or unintended uses (e.g., disinformation, generating fake profiles, surveillance), fairness considerations (e.g., deployment of technologies that could make decisions that unfairly impact specific groups), privacy considerations, and security considerations.
        \item The conference expects that many papers will be foundational research and not tied to particular applications, let alone deployments. However, if there is a direct path to any negative applications, the authors should point it out. For example, it is legitimate to point out that an improvement in the quality of generative models could be used to generate Deepfakes for disinformation. On the other hand, it is not needed to point out that a generic algorithm for optimizing neural networks could enable people to train models that generate Deepfakes faster.
        \item The authors should consider possible harms that could arise when the technology is being used as intended and functioning correctly, harms that could arise when the technology is being used as intended but gives incorrect results, and harms following from (intentional or unintentional) misuse of the technology.
        \item If there are negative societal impacts, the authors could also discuss possible mitigation strategies (e.g., gated release of models, providing defenses in addition to attacks, mechanisms for monitoring misuse, mechanisms to monitor how a system learns from feedback over time, improving the efficiency and accessibility of ML).
    \end{itemize}
    
\item {\bf Safeguards}
    \item[] Question: Does the paper describe safeguards that have been put in place for responsible release of data or models that have a high risk for misuse (e.g., pre-trained language models, image generators, or scraped datasets)?
    \item[] Answer: \answerNA{}
    \item[] Justification: The paper poses no such risks.
    \item[] Guidelines:
    \begin{itemize}
        \item The answer \answerNA{} means that the paper poses no such risks.
        \item Released models that have a high risk for misuse or dual-use should be released with necessary safeguards to allow for controlled use of the model, for example by requiring that users adhere to usage guidelines or restrictions to access the model or implementing safety filters. 
        \item Datasets that have been scraped from the Internet could pose safety risks. The authors should describe how they avoided releasing unsafe images.
        \item We recognize that providing effective safeguards is challenging, and many papers do not require this, but we encourage authors to take this into account and make a best faith effort.
    \end{itemize}

\item {\bf Licenses for existing assets}
    \item[] Question: Are the creators or original owners of assets (e.g., code, data, models), used in the paper, properly credited and are the license and terms of use explicitly mentioned and properly respected?
    \item[] Answer: \answerYes{}
    \item[] Justification: The environments used for our evaluation and the code used in our benchmarks are properly credited and cited.
    \item[] Guidelines:
    \begin{itemize}
        \item The answer \answerNA{} means that the paper does not use existing assets.
        \item The authors should cite the original paper that produced the code package or dataset.
        \item The authors should state which version of the asset is used and, if possible, include a URL.
        \item The name of the license (e.g., CC-BY 4.0) should be included for each asset.
        \item For scraped data from a particular source (e.g., website), the copyright and terms of service of that source should be provided.
        \item If assets are released, the license, copyright information, and terms of use in the package should be provided. For popular datasets, \url{paperswithcode.com/datasets} has curated licenses for some datasets. Their licensing guide can help determine the license of a dataset.
        \item For existing datasets that are re-packaged, both the original license and the license of the derived asset (if it has changed) should be provided.
        \item If this information is not available online, the authors are encouraged to reach out to the asset's creators.
    \end{itemize}

\item {\bf New assets}
    \item[] Question: Are new assets introduced in the paper well documented and is the documentation provided alongside the assets?
    \item[] Answer: \answerYes{}
    \item[] Justification: We release the code for UCP, including its multi-constraint extension, with documentation and an Apache 2.0 license at \url{https://github.com/MehrdadMoghimi/UCP}.
    \item[] Guidelines:
    \begin{itemize}
        \item The answer \answerNA{} means that the paper does not release new assets.
        \item Researchers should communicate the details of the dataset\slash code\slash model as part of their submissions via structured templates. This includes details about training, license, limitations, etc. 
        \item The paper should discuss whether and how consent was obtained from people whose asset is used.
        \item At submission time, remember to anonymize your assets (if applicable). You can either create an anonymized URL or include an anonymized zip file.
    \end{itemize}

\item {\bf Crowdsourcing and research with human subjects}
    \item[] Question: For crowdsourcing experiments and research with human subjects, does the paper include the full text of instructions given to participants and screenshots, if applicable, as well as details about compensation (if any)? 
    \item[] Answer: \answerNA{}
    \item[] Justification: The paper does not involve crowdsourcing nor research with human subjects.
    \item[] Guidelines:
    \begin{itemize}
        \item The answer \answerNA{} means that the paper does not involve crowdsourcing nor research with human subjects.
        \item Including this information in the supplemental material is fine, but if the main contribution of the paper involves human subjects, then as much detail as possible should be included in the main paper. 
        \item According to the NeurIPS Code of Ethics, workers involved in data collection, curation, or other labor should be paid at least the minimum wage in the country of the data collector. 
    \end{itemize}

\item {\bf Institutional review board (IRB) approvals or equivalent for research with human subjects}
    \item[] Question: Does the paper describe potential risks incurred by study participants, whether such risks were disclosed to the subjects, and whether Institutional Review Board (IRB) approvals (or an equivalent approval/review based on the requirements of your country or institution) were obtained?
    \item[] Answer: \answerNA{}
    \item[] Justification: The paper does not involve crowdsourcing nor research with human subjects.
    \item[] Guidelines:
    \begin{itemize}
        \item The answer \answerNA{} means that the paper does not involve crowdsourcing nor research with human subjects.
        \item Depending on the country in which research is conducted, IRB approval (or equivalent) may be required for any human subjects research. If you obtained IRB approval, you should clearly state this in the paper. 
        \item We recognize that the procedures for this may vary significantly between institutions and locations, and we expect authors to adhere to the NeurIPS Code of Ethics and the guidelines for their institution. 
        \item For initial submissions, do not include any information that would break anonymity (if applicable), such as the institution conducting the review.
    \end{itemize}

\item {\bf Declaration of LLM usage}
    \item[] Question: Does the paper describe the usage of LLMs if it is an important, original, or non-standard component of the core methods in this research? Note that if the LLM is used only for writing, editing, or formatting purposes and does \emph{not} impact the core methodology, scientific rigor, or originality of the research, declaration is not required.

    \item[] Answer: \answerNA{}
    \item[] Justification: The core method development in this research does not involve LLMs.
    \item[] Guidelines:
    \begin{itemize}
        \item The answer \answerNA{} means that the core method development in this research does not involve LLMs as any important, original, or non-standard components.
        \item Please refer to our LLM policy in the NeurIPS handbook for what should or should not be described.
    \end{itemize}

\end{enumerate}

\end{document}